%% file: mainarxiv.tex
\documentclass{article}
\usepackage{iclr2027_conference,times}
\usepackage[letterpaper,margin=1.05in,includeheadfoot,headheight=12pt,headsep=18pt,footskip=24pt]{geometry}
\usepackage{amsmath,amssymb,amsfonts,amsthm,mathtools}
\usepackage{bm}

\usepackage[utf8]{inputenc}
\usepackage[T1]{fontenc}
\DeclareUnicodeCharacter{2212}{\ensuremath{-}}  
\usepackage{url}
\usepackage{booktabs}
\usepackage[table]{xcolor}
\usepackage{colortbl}
\usepackage{microtype}
\usepackage{graphicx}
\IfFileExists{latexml.sty}{\usepackage{latexml}}{\newif\iflatexml}
\newcommand{\galhtmlname}[1]{\iflatexml\mbox{#1}\else#1\fi}

\usepackage{wrapfig}
\usepackage{algorithm}
\usepackage{algpseudocode}
\usepackage{hyperref}


\usepackage[capitalize,noabbrev]{cleveref}

\newtheoremstyle{notefirst}{\topsep}{\topsep}{\itshape}{}{\bfseries}{.}{ }{\thmnote{#3. }\thmname{#1}~\thmnumber{#2}}
\theoremstyle{notefirst}

\theoremstyle{plain}

\theoremstyle{definition}

\theoremstyle{remark}

\crefname{assumption}{Assumption}{Assumptions}
\Crefname{assumption}{Assumption}{Assumptions}
\crefname{proposition}{Proposition}{Propositions}
\Crefname{proposition}{Proposition}{Propositions}
\crefname{algorithm}{Algorithm}{Algorithms}
\Crefname{algorithm}{Algorithm}{Algorithms}
\crefalias{assumption}{assumption}
\crefalias{proposition}{proposition}
\crefname{equation}{Eq.}{Eqs.}
\Crefname{equation}{Eq.}{Eqs.}

\input{math_commands}

\iflatexml\else\iclrfinalcopy\fi

\title{Golden Anchor for Video Generation}
\iflatexml
\author{%
Jinyuan Xu$^{1}$ \quad Shitong Shao$^{1}$ \quad Lichen Bai$^{1}$ \quad Zikai Zhou$^{1}$ \\
{\bfseries Guoxi Huang$^{2}$} \quad {\bfseries Nantheera Anantrasirichai$^{2}$} \quad {\bfseries Zeke Xie$^{1}$} \\
{\normalfont\small $^{1}$The Hong Kong University of Science and Technology (Guangzhou)} \\
{\normalfont\small $^{2}$University of Bristol}
}
\else
\author{%
Jinyuan Xu$^{1}$ \quad Shitong Shao$^{1}$ \quad Lichen Bai$^{1}$ \quad Zikai Zhou$^{1}$ \\
{\bfseries Guoxi Huang$^{2}$ \quad Nantheera Anantrasirichai$^{2}$ \quad Zeke Xie$^{1}$} \\
{\normalfont\small $^{1}$The Hong Kong University of Science and Technology (Guangzhou)} \\
{\normalfont\small $^{2}$University of Bristol}
}
\fi
\hypersetup{
  pdftitle={Golden Anchor for Video Generation},
  pdfauthor={Jinyuan Xu, Shitong Shao, Lichen Bai, Zikai Zhou, Guoxi Huang, Nantheera Anantrasirichai, Zeke Xie}
}

\begin{document}

\maketitle
\lhead{Preprint}

\begin{abstract}
Video generation supports digital content creation, with applications in virtual avatars, AI-driven filmmaking, interactive game engines, and world models. State-of-the-art systems achieve controllability primarily through image-to-video (I2V) generation, in which an injected \emph{anchor} image determines the identity, composition, and style of the resulting video. In modern pipelines, this anchor is synthesized by a text-to-image (T2I) generator. This raises a largely overlooked question: \emph{Are images obtained directly from the T2I model truly suitable for current video models?} Our analysis shows that anchor aesthetic and quality scores are weakly correlated with downstream video quality. The structure-and-texture score is negatively correlated with I2V video quality in both the visual and motion dimensions. Fine-tuning the T2I model on an image reward accordingly yields anchors with visibly denser structural and character detail: they score higher on image metrics, yet lower the video reward on every I2V model we test and change downstream motion unpredictably. We therefore introduce \textbf{Golden Anchor Learning} (\textbf{GAL}), to our knowledge the first framework for learning golden anchors by optimizing a T2I generator with downstream video feedback. GAL evaluates candidate anchors through the videos they induce under a frozen I2V model and converts video rewards into image-level preferences for DPO fine-tuning of the T2I generator. On all four I2V models, GAL raises the VBench average by $2.97\%$--$8.46\%$ over the untuned generator. The rendered clips also score higher under VideoReward-Overall, and their extracted frames under HPSv3 and UniPercept. GAL yields consistent motion gains across all four I2V models, as measured by a $5.62\%$--$38.24\%$ increase in VBench dynamic degree. Although GAL is trained only on HunyuanVideo-1.5 renders, these gains extend to the other three I2V models.
\end{abstract}

\ifdefined\fh\else\newlength{\fh}\fi
\providecommand{\qgrot}[1]{\raisebox{0.5\fh}{\rotatebox[origin=c]{-90}{\makebox[\fh][l]{\fontsize{8}{9}\selectfont #1}}}}
\begin{figure}[!t]
  \centering
  \iflatexml
  \includegraphics[width=1.5\textwidth]{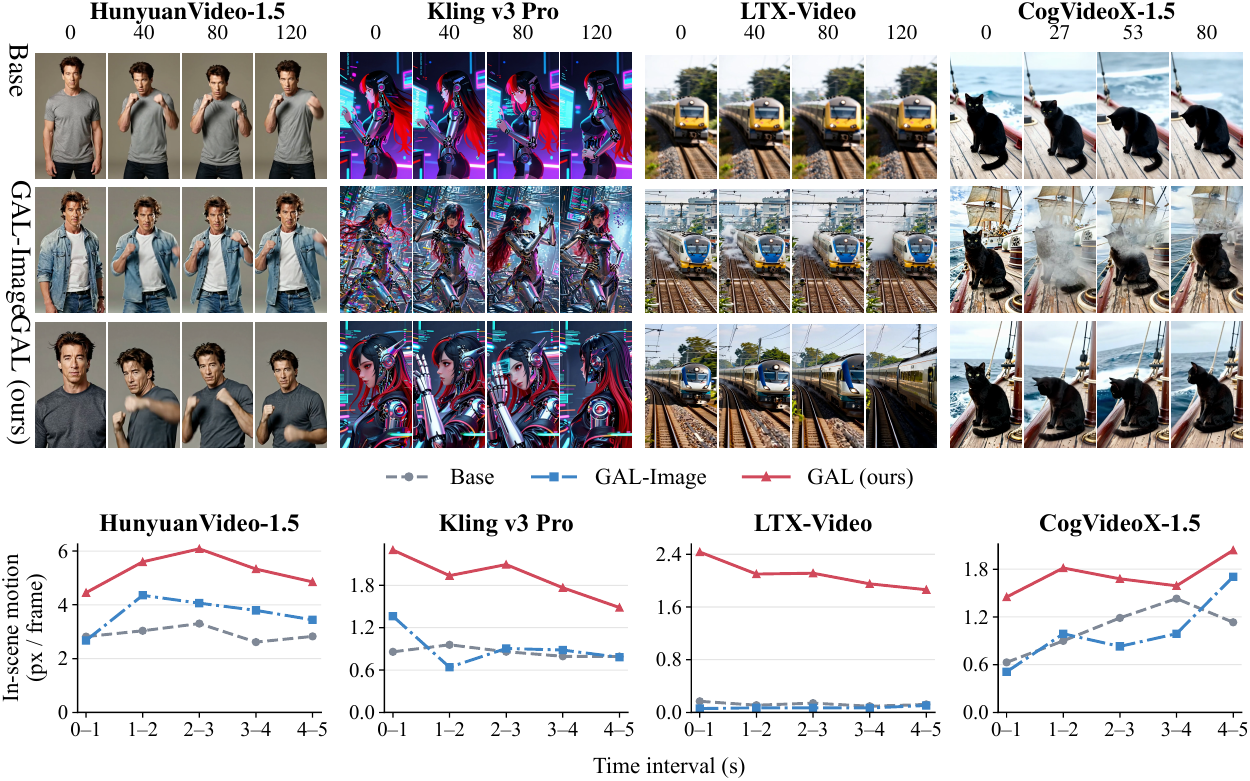}
  \else
  \setlength{\tabcolsep}{0.4pt}%
  \renewcommand{\arraystretch}{0.3}%
  \def\fw{0.0440\linewidth}
  \setlength{\fh}{0.078\linewidth}
  {\scriptsize
  \begin{tabular}{@{}c@{\,}cccc@{\hspace{4pt}}cccc@{\hspace{4pt}}cccc@{\hspace{4pt}}cccc@{}}
    & \multicolumn{4}{c}{\textbf{HunyuanVideo-1.5}}
    & \multicolumn{4}{c}{\textbf{Kling v3 Pro}}
    & \multicolumn{4}{c}{\textbf{LTX-Video}}
    & \multicolumn{4}{c}{\textbf{CogVideoX-1.5}} \\[1pt]
    & {\tiny 0} & {\tiny 40} & {\tiny 80} & {\tiny 120}
    & {\tiny 0} & {\tiny 40} & {\tiny 80} & {\tiny 120}
    & {\tiny 0} & {\tiny 40} & {\tiny 80} & {\tiny 120}
    & {\tiny 0} & {\tiny 27} & {\tiny 53} & {\tiny 80} \\[0.5pt]
    \qgrot{Base}
      & \includegraphics[width=\fw]{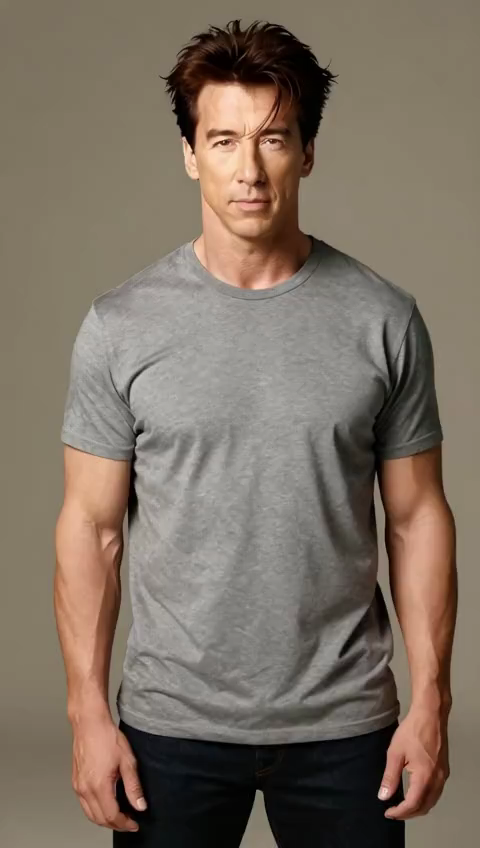}    & \includegraphics[width=\fw]{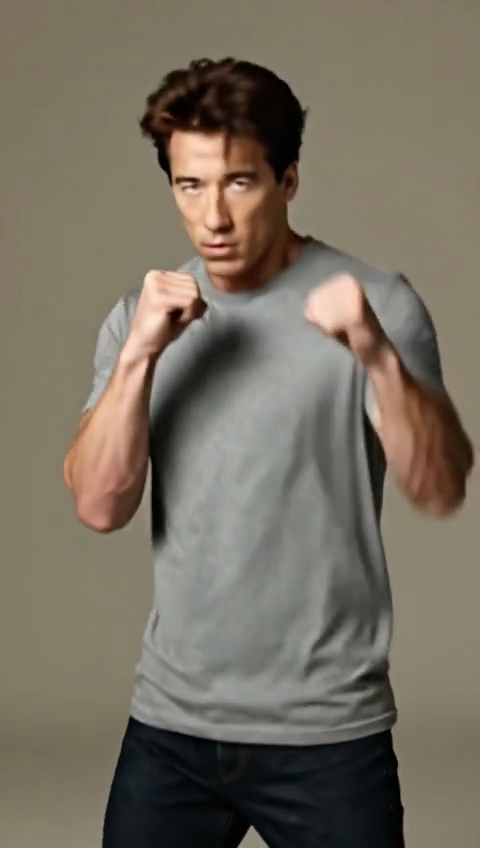}    & \includegraphics[width=\fw]{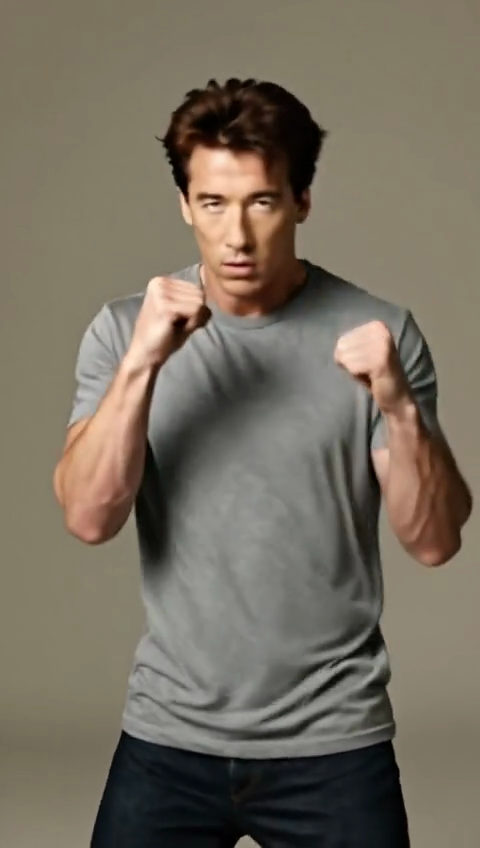}    & \includegraphics[width=\fw]{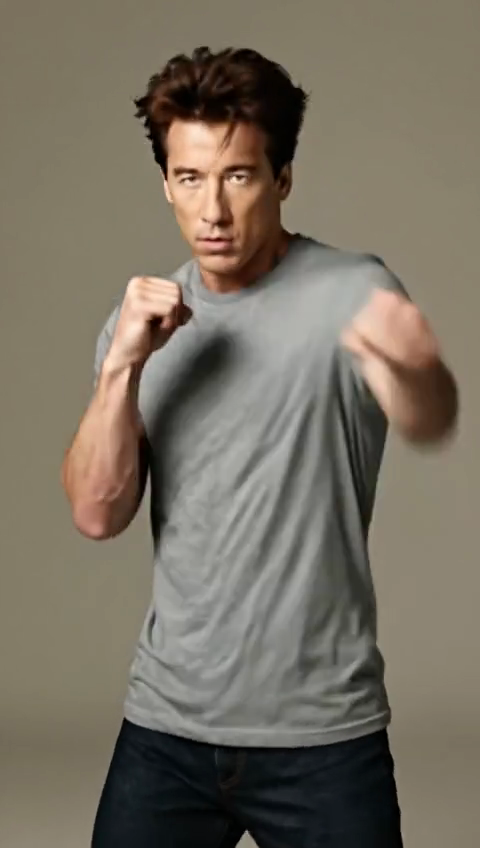}
      & \includegraphics[width=\fw]{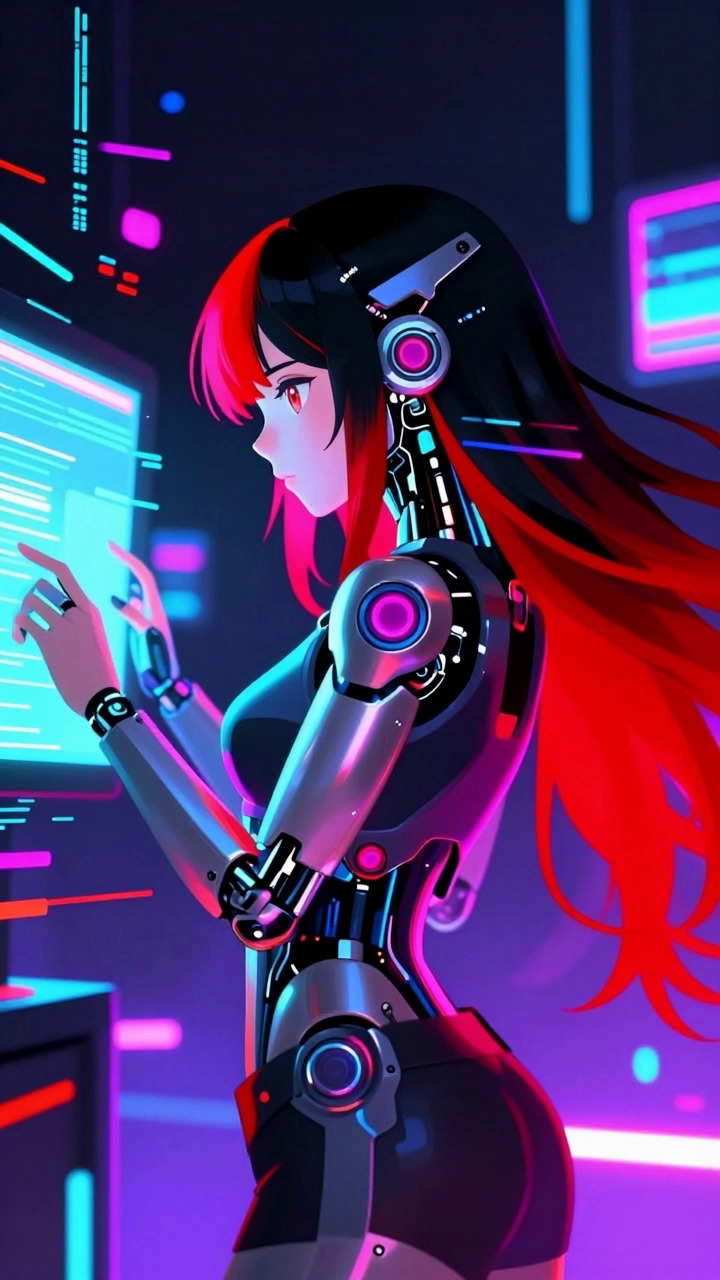} & \includegraphics[width=\fw]{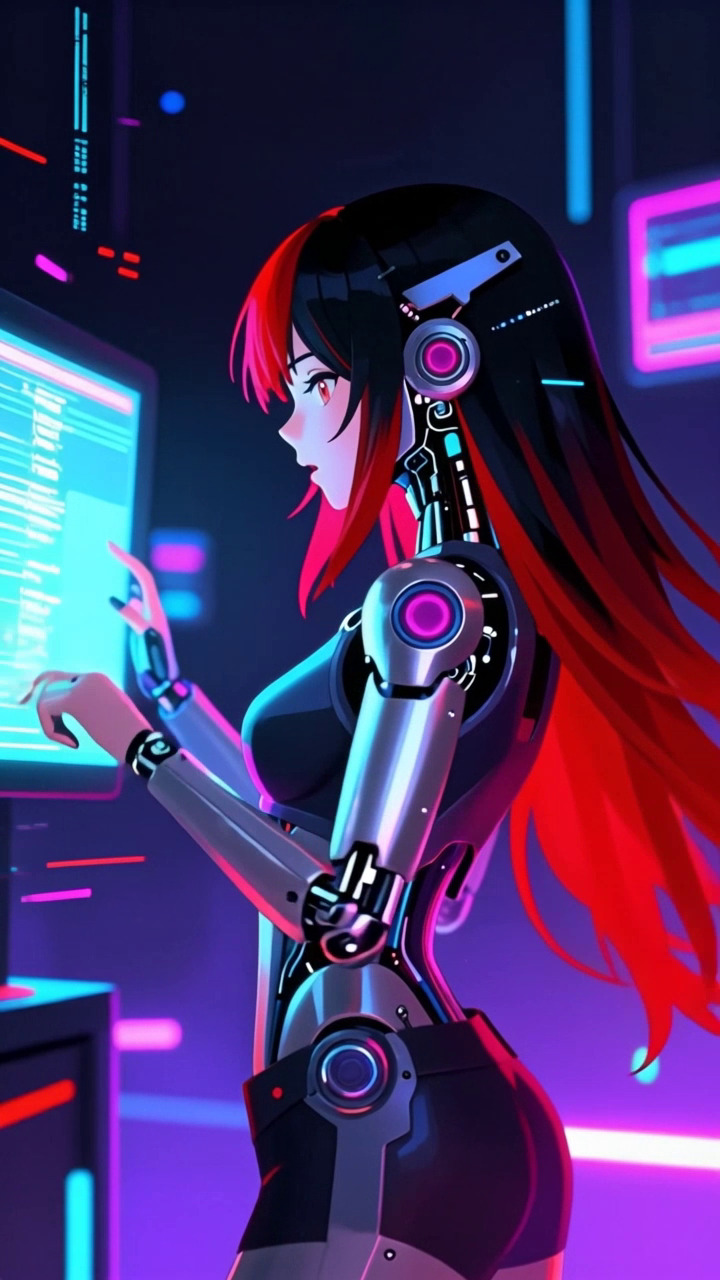} & \includegraphics[width=\fw]{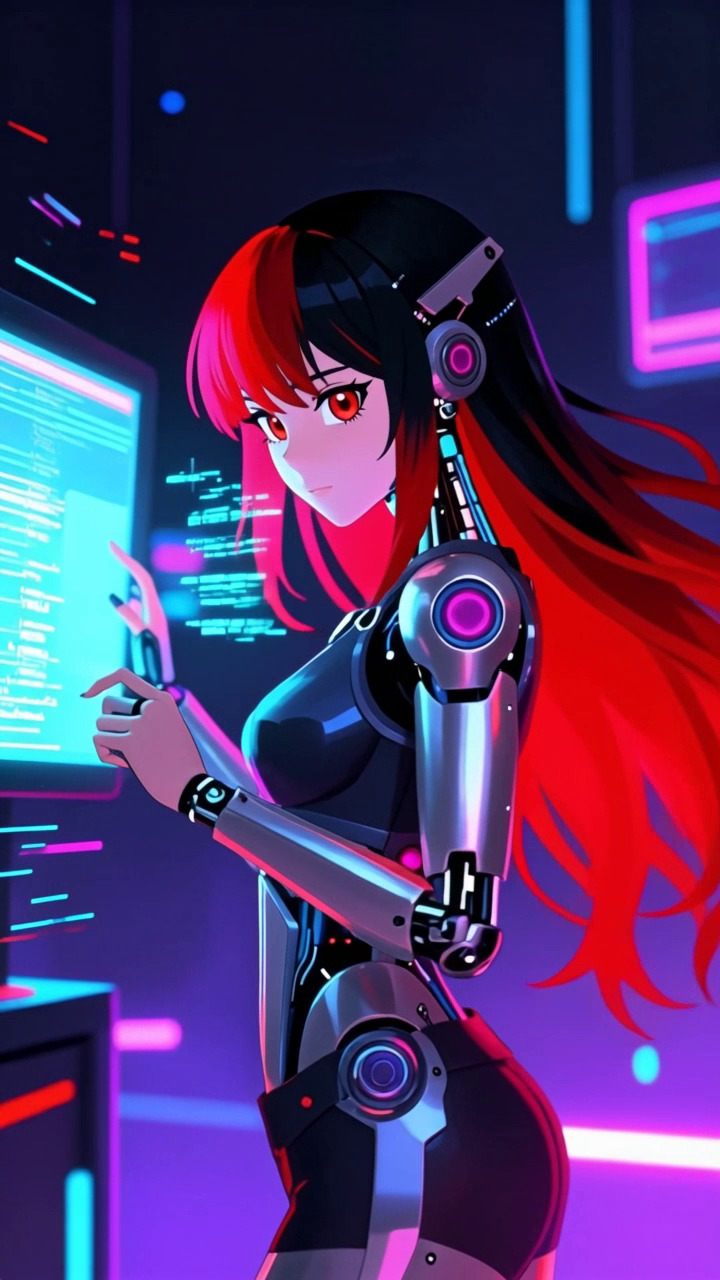} & \includegraphics[width=\fw]{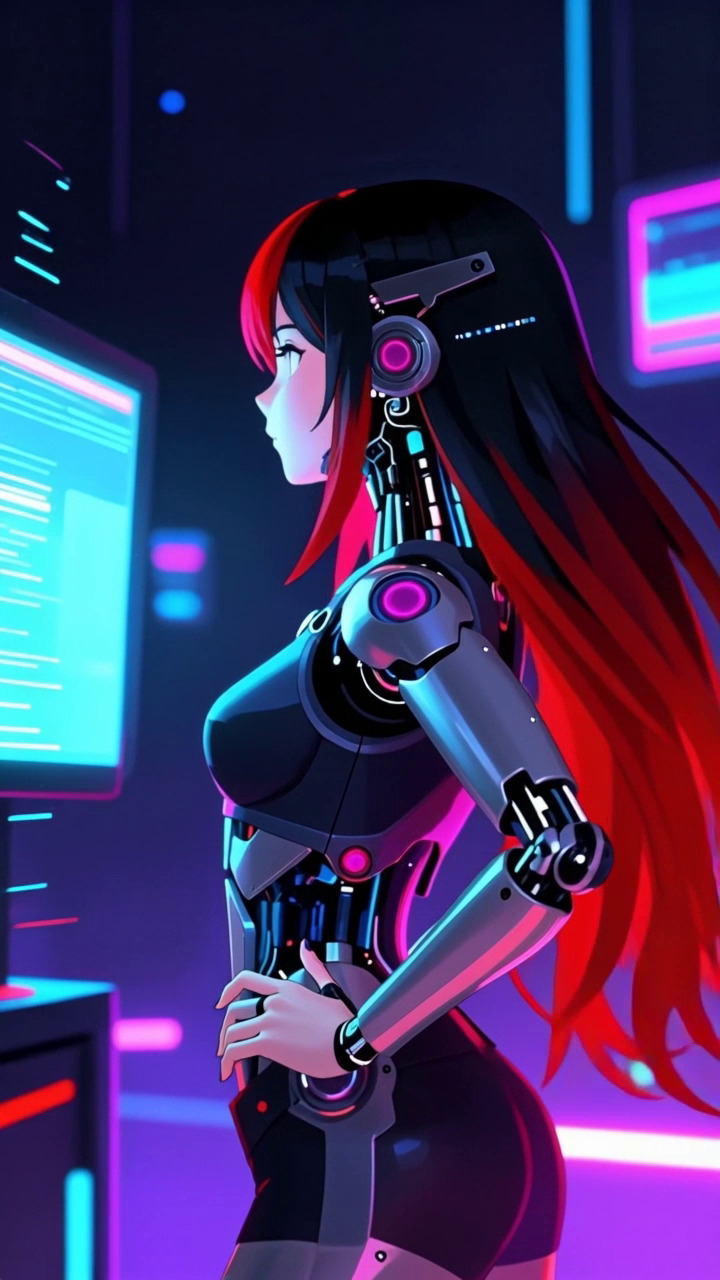}
      & \includegraphics[width=\fw]{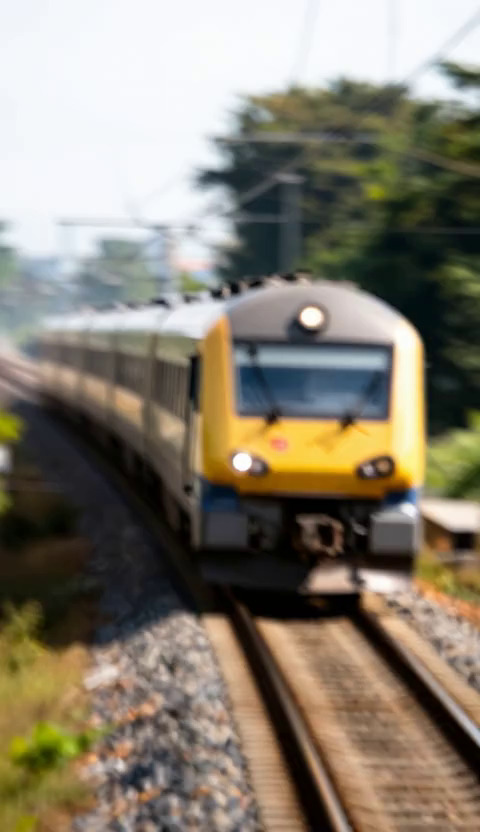}   & \includegraphics[width=\fw]{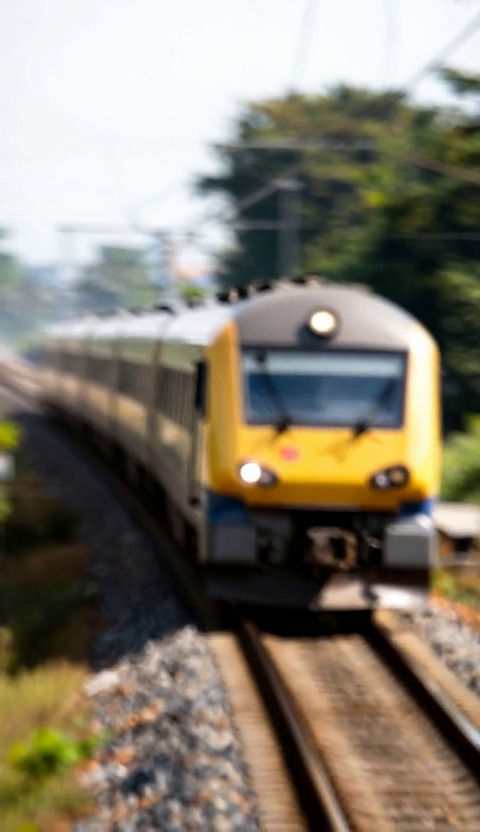}   & \includegraphics[width=\fw]{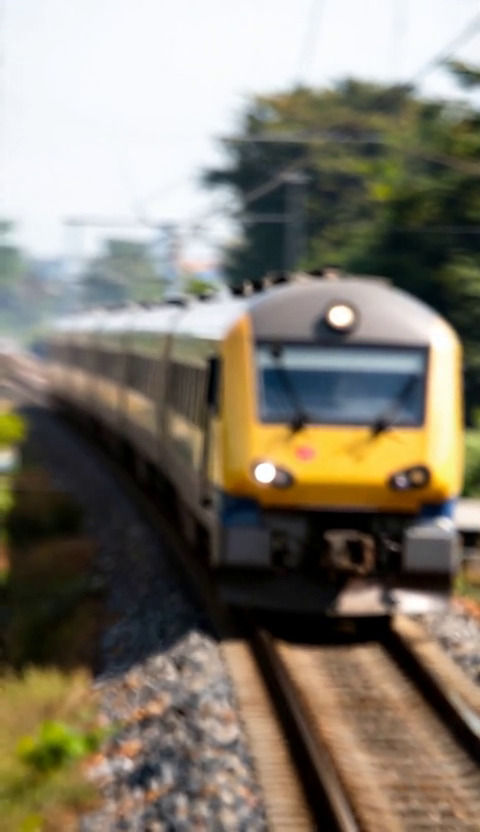}   & \includegraphics[width=\fw]{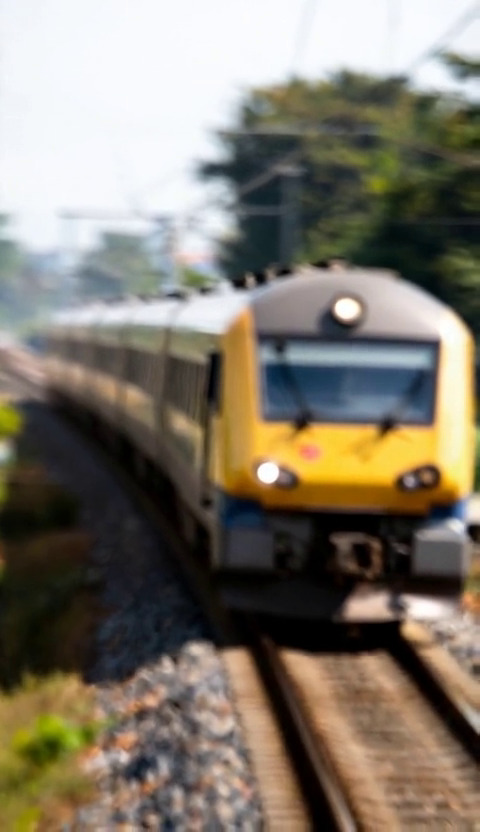}
      & \includegraphics[width=\fw]{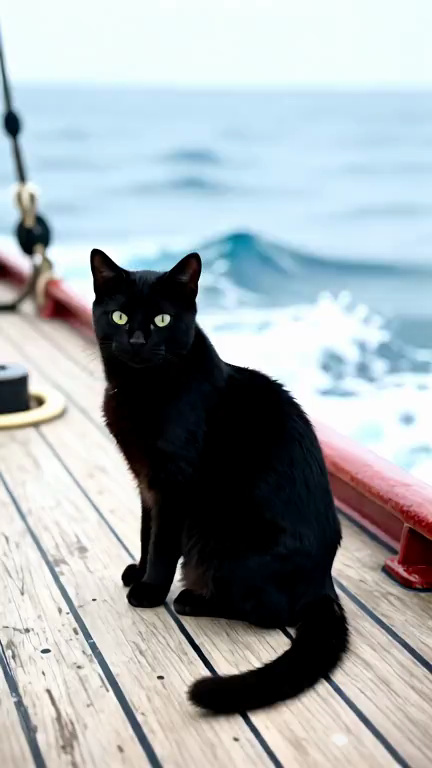}   & \includegraphics[width=\fw]{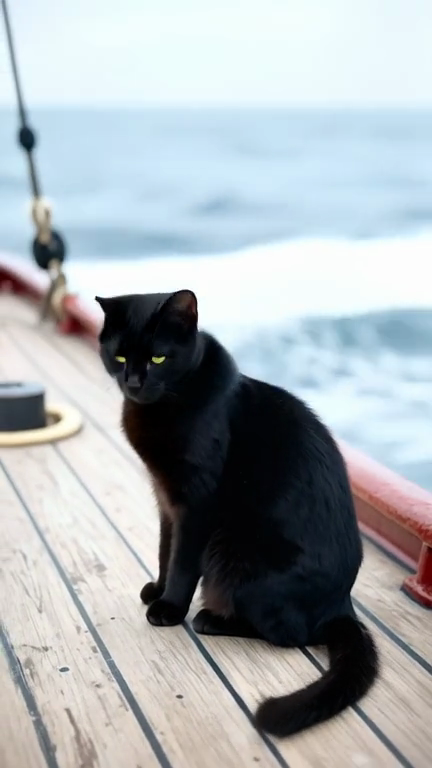}   & \includegraphics[width=\fw]{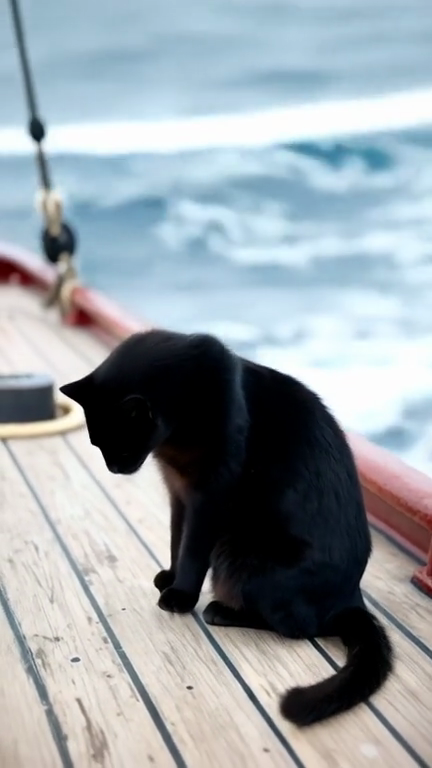}   & \includegraphics[width=\fw]{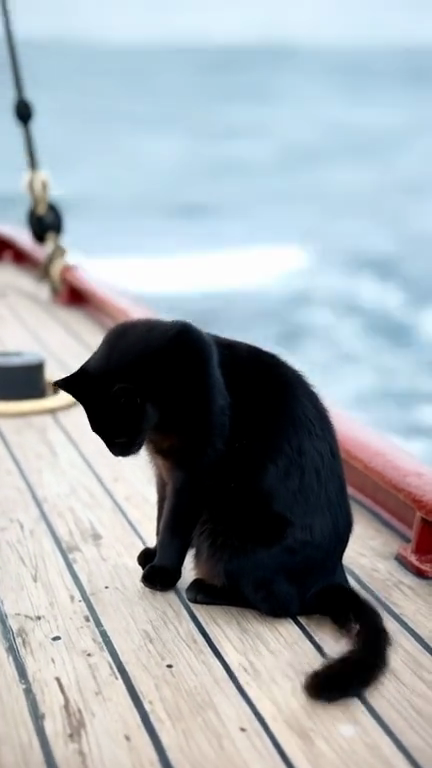} \\
    \qgrot{GAL-Image}
      & \includegraphics[width=\fw]{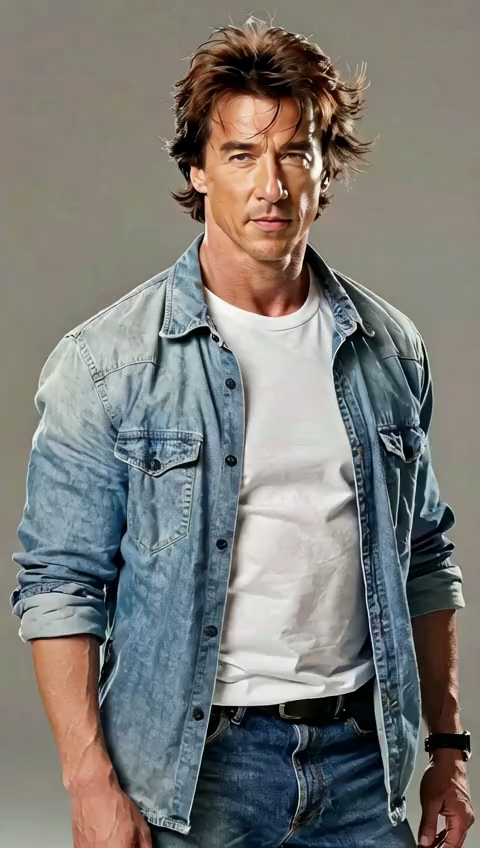}    & \includegraphics[width=\fw]{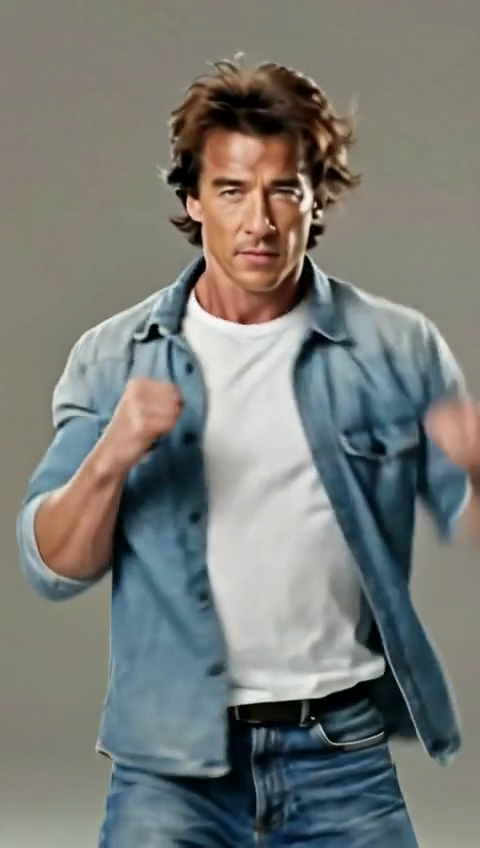}    & \includegraphics[width=\fw]{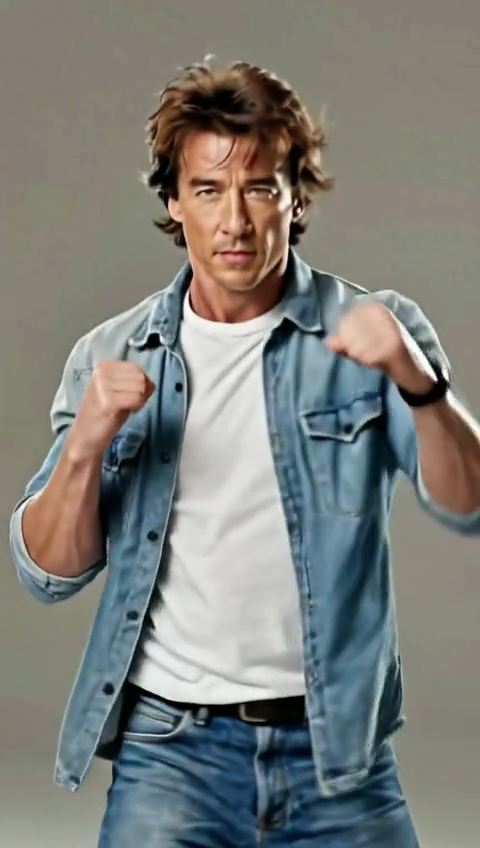}    & \includegraphics[width=\fw]{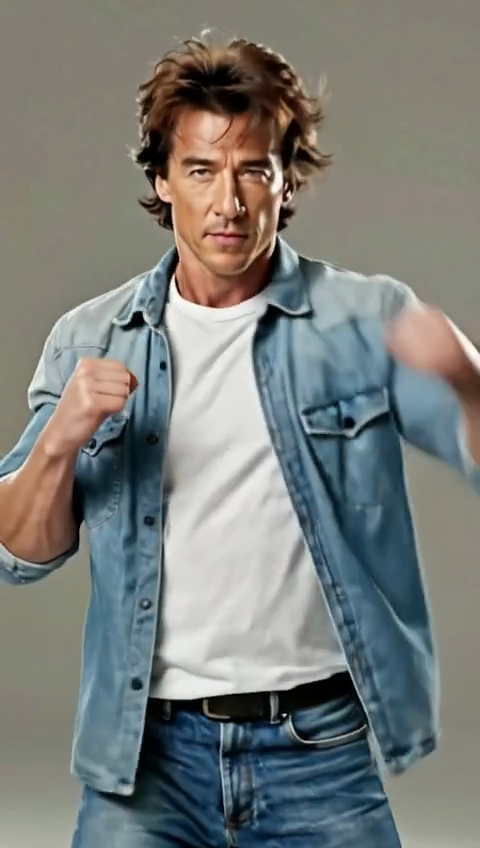}
      & \includegraphics[width=\fw]{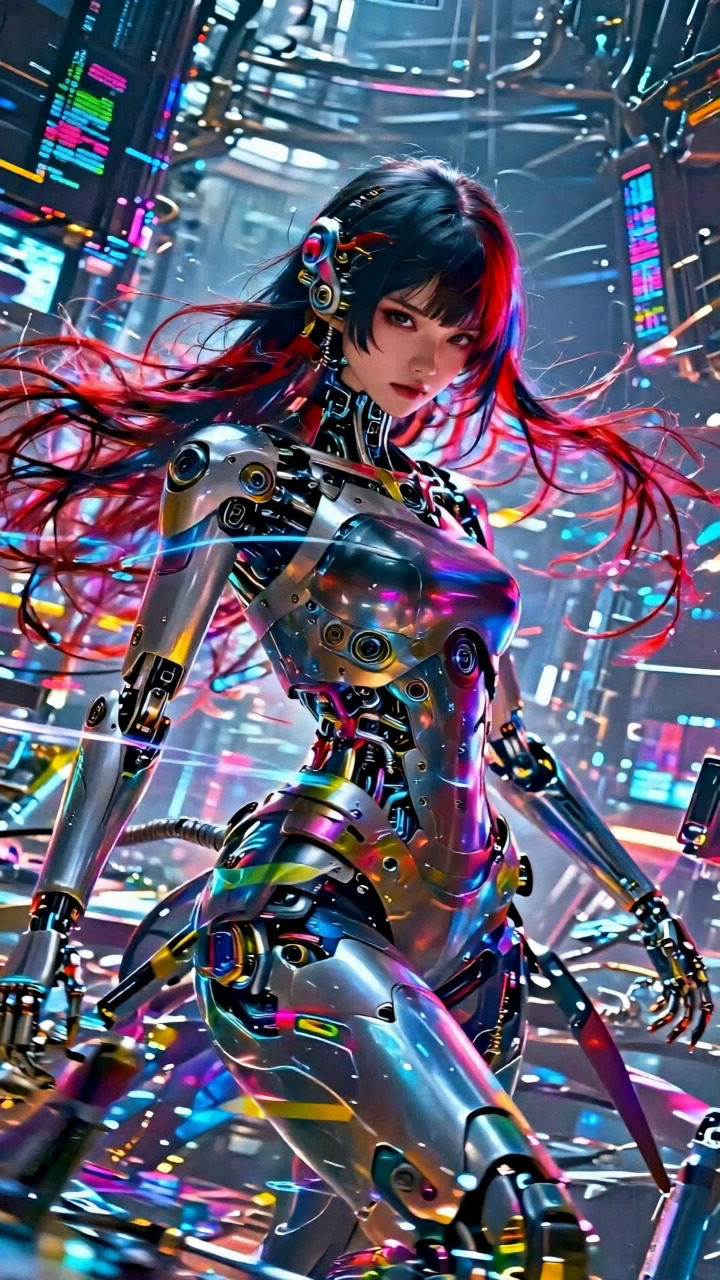} & \includegraphics[width=\fw]{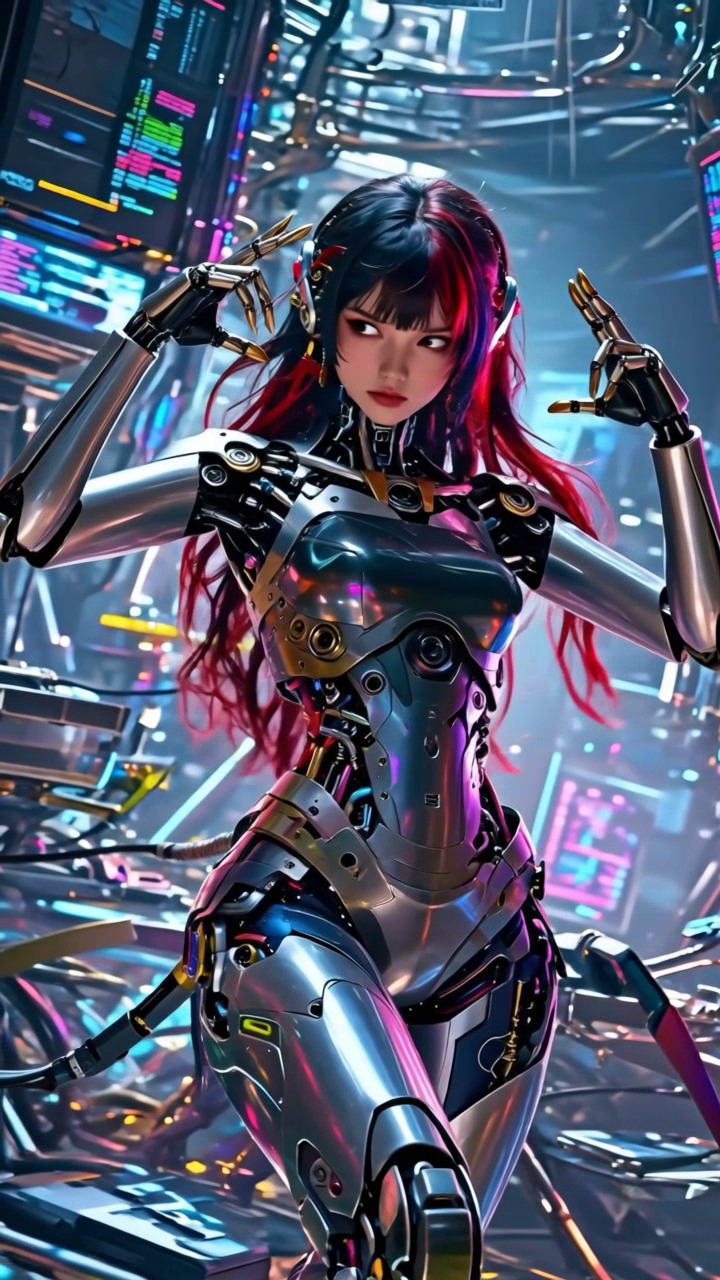} & \includegraphics[width=\fw]{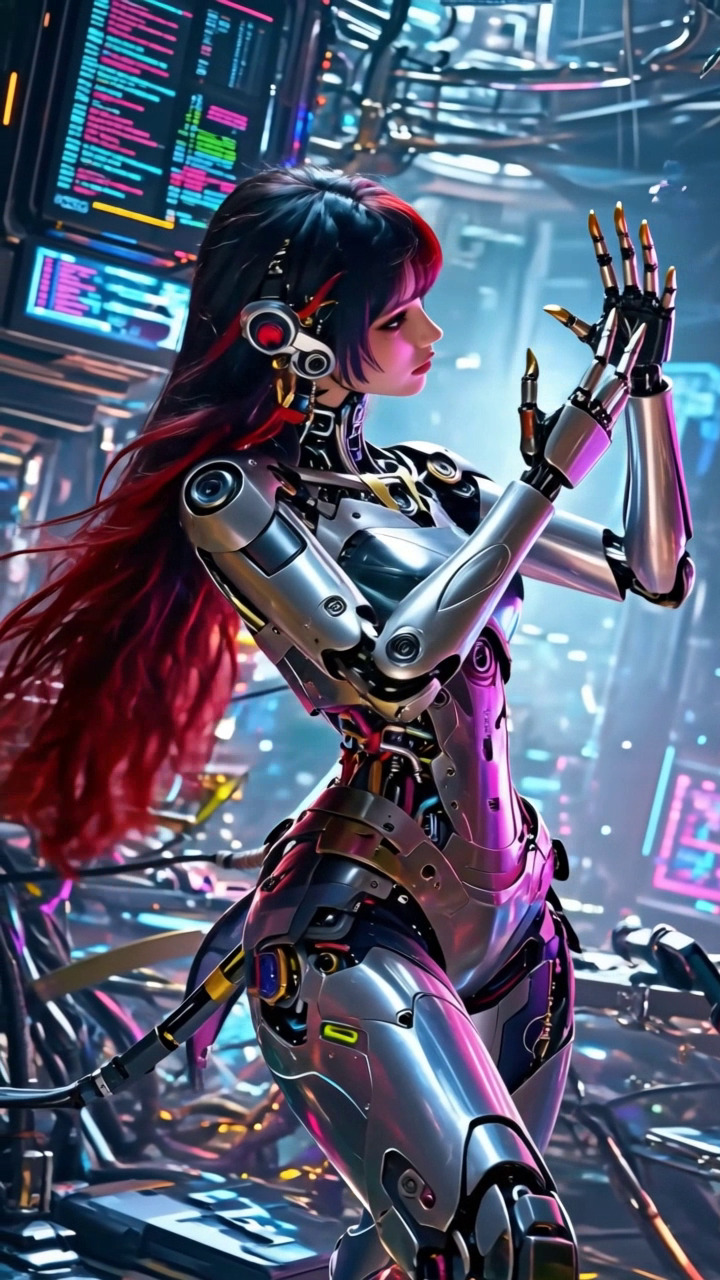} & \includegraphics[width=\fw]{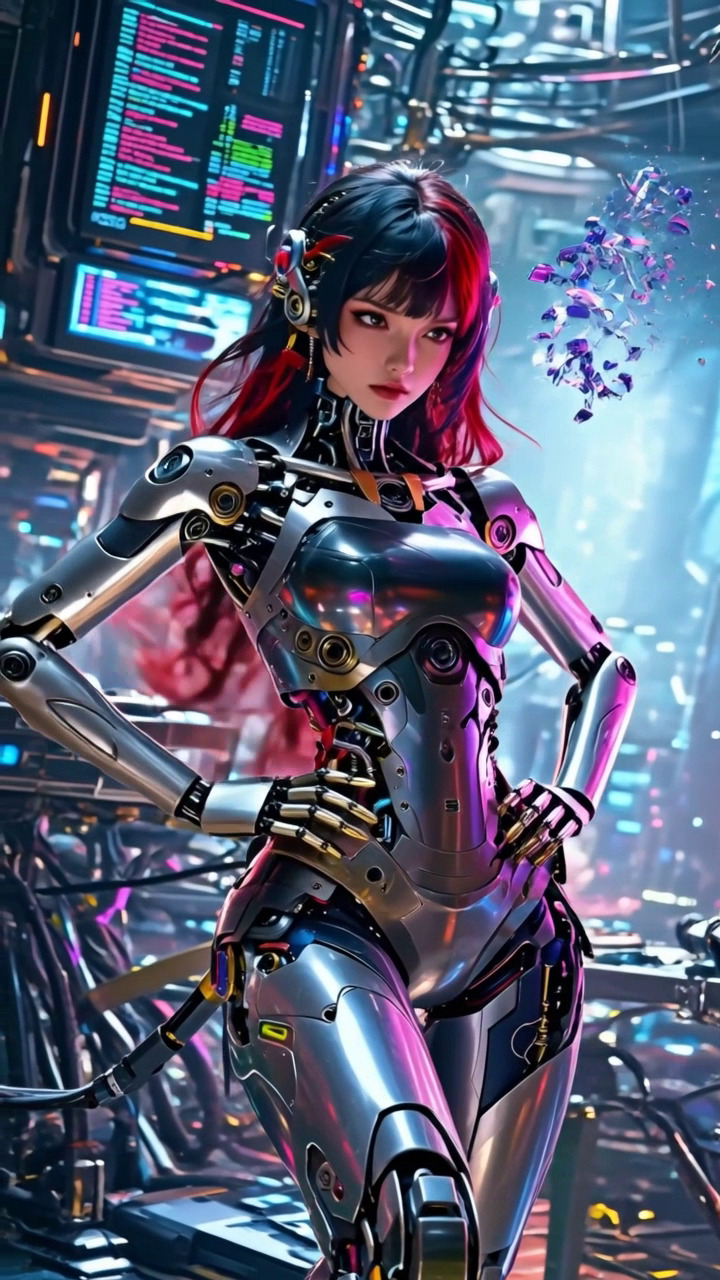}
      & \includegraphics[width=\fw]{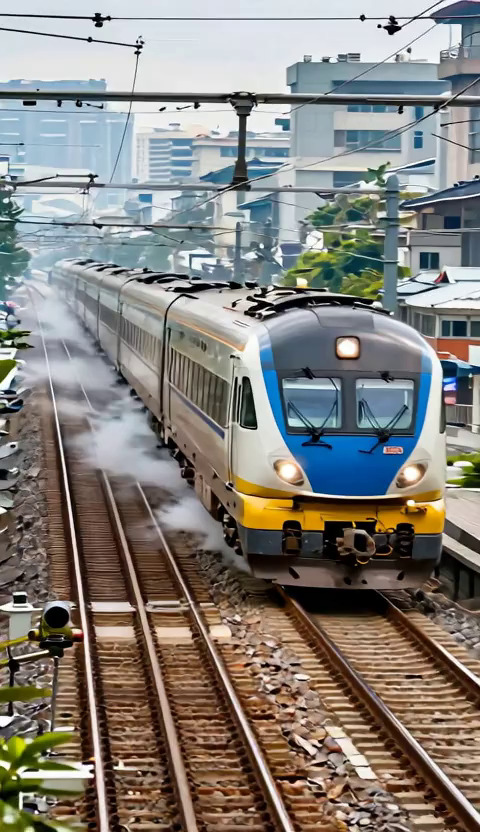}   & \includegraphics[width=\fw]{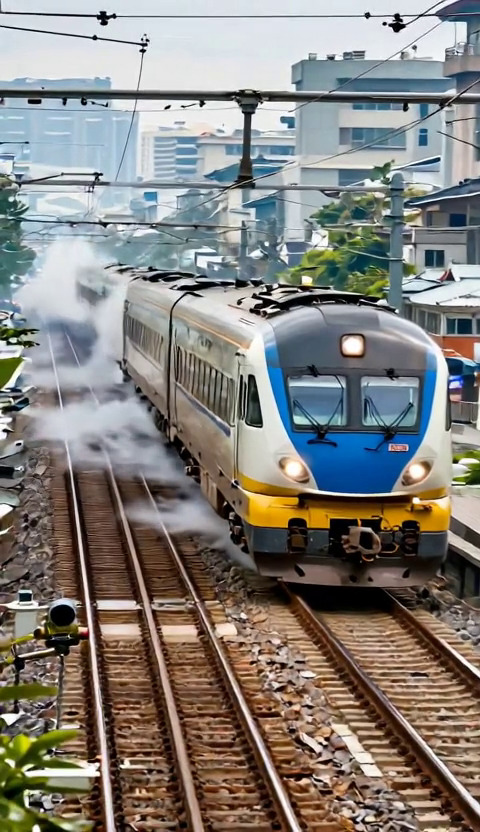}   & \includegraphics[width=\fw]{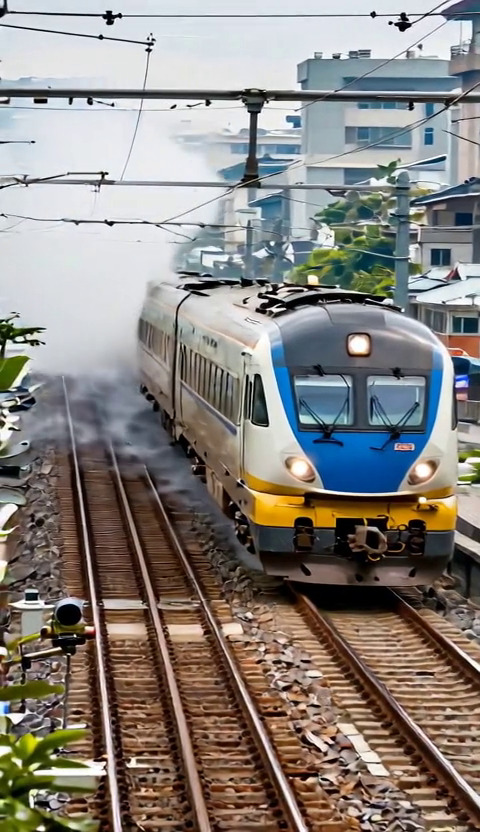}   & \includegraphics[width=\fw]{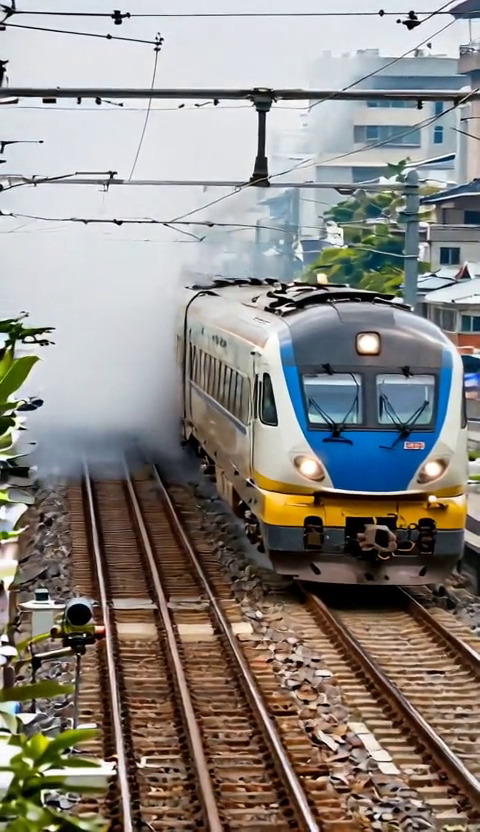}
      & \includegraphics[width=\fw]{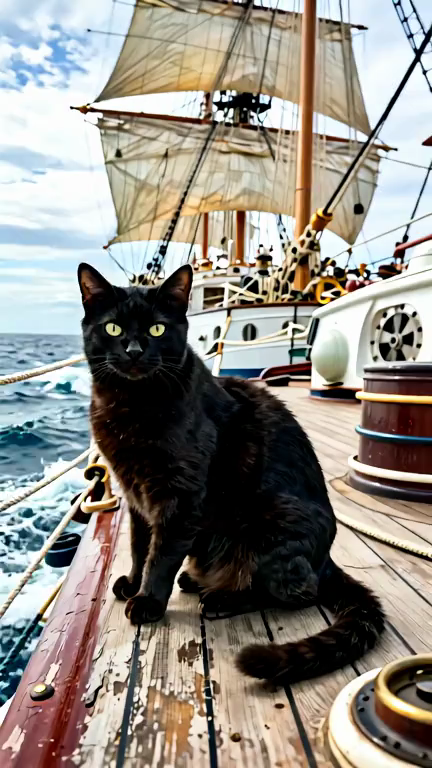}   & \includegraphics[width=\fw]{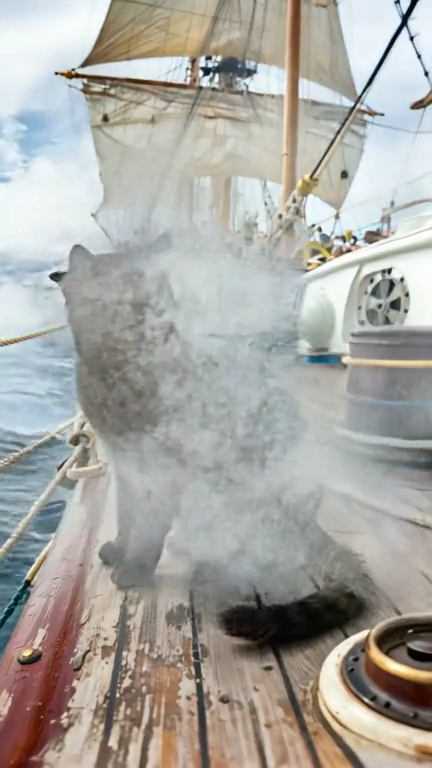}   & \includegraphics[width=\fw]{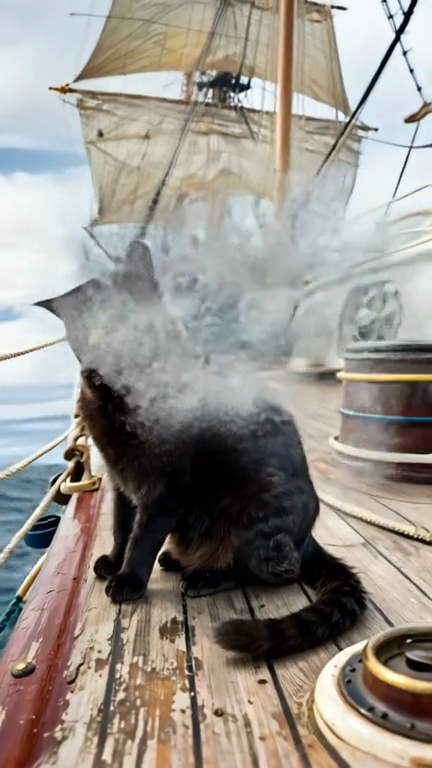}   & \includegraphics[width=\fw]{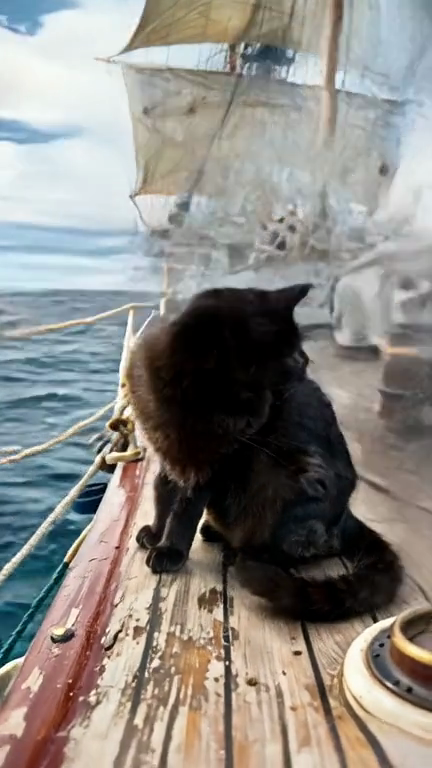} \\
    \qgrot{GAL (ours)}
      & \includegraphics[width=\fw]{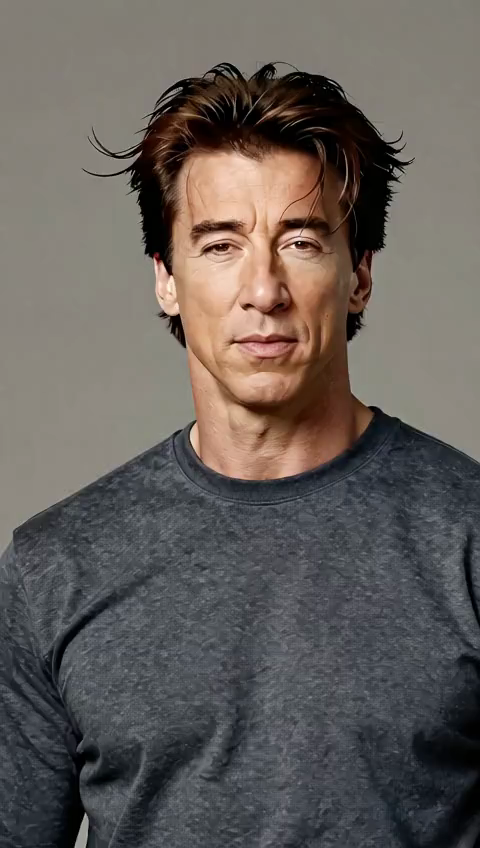}    & \includegraphics[width=\fw]{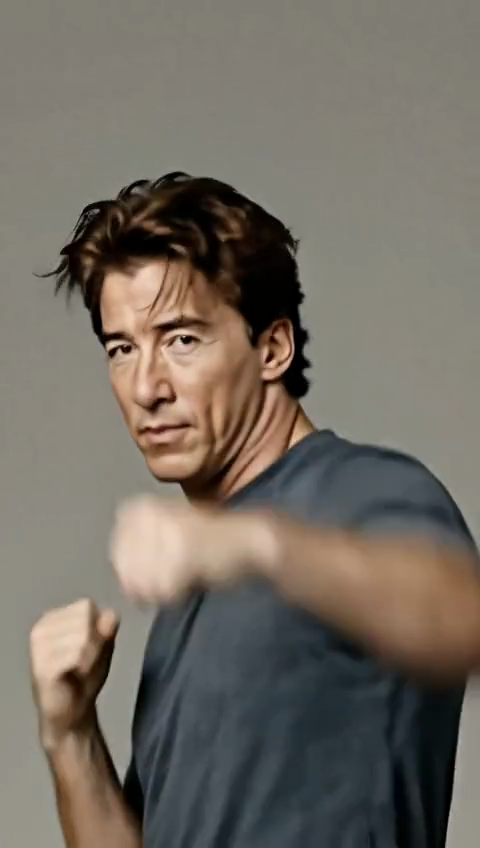}    & \includegraphics[width=\fw]{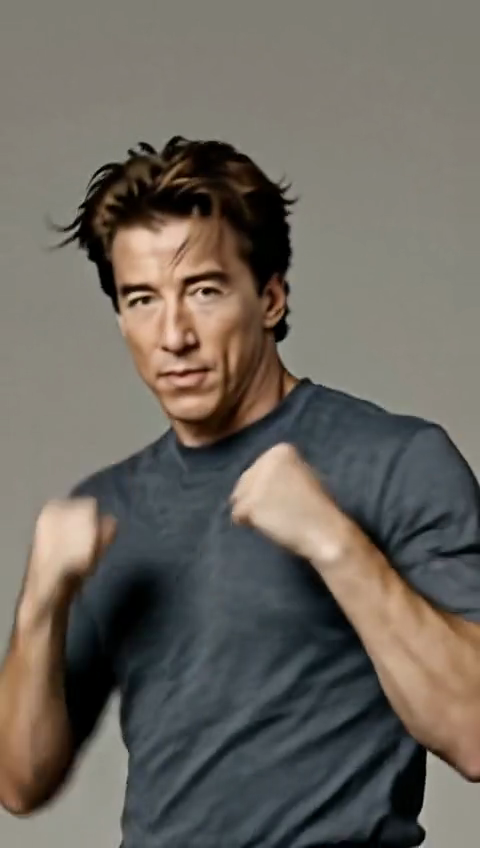}    & \includegraphics[width=\fw]{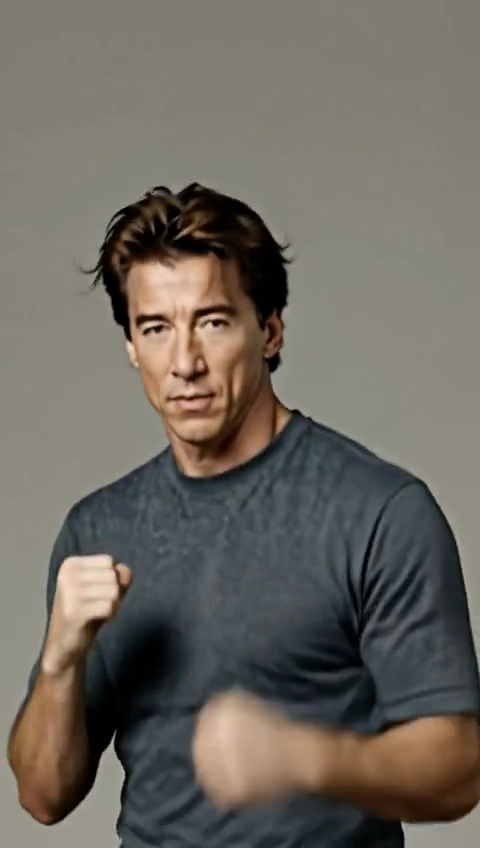}
      & \includegraphics[width=\fw]{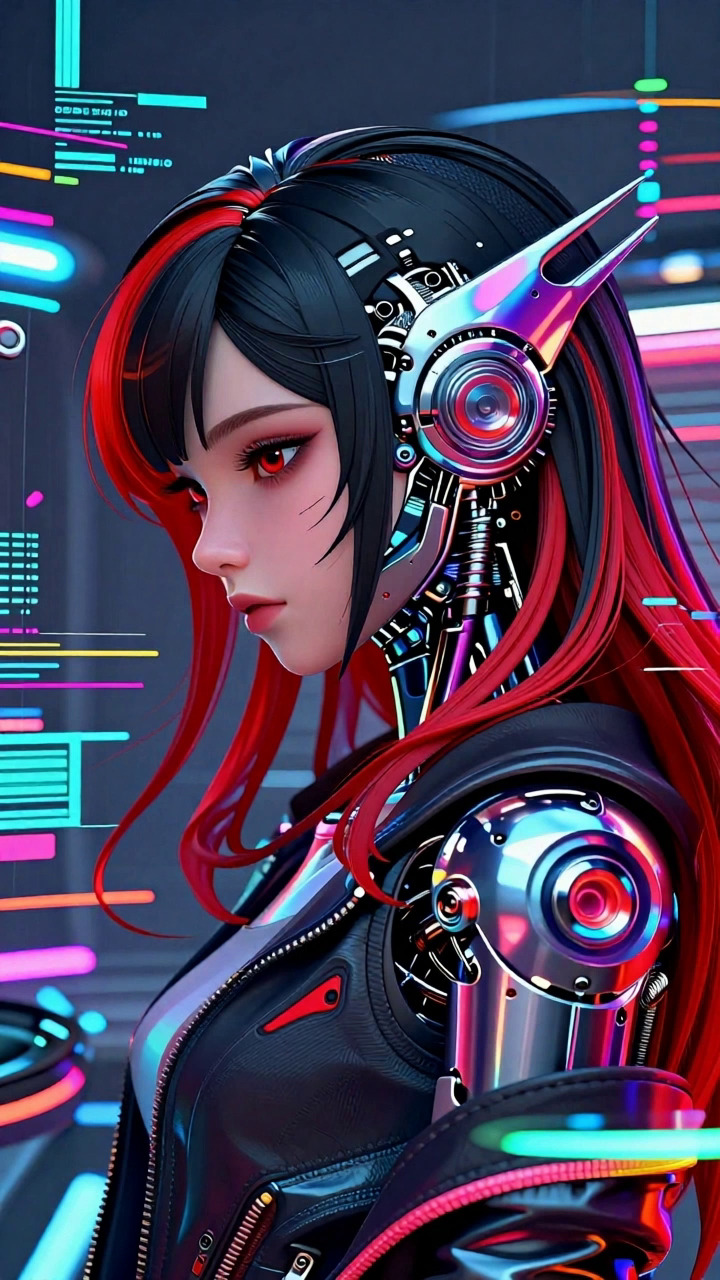} & \includegraphics[width=\fw]{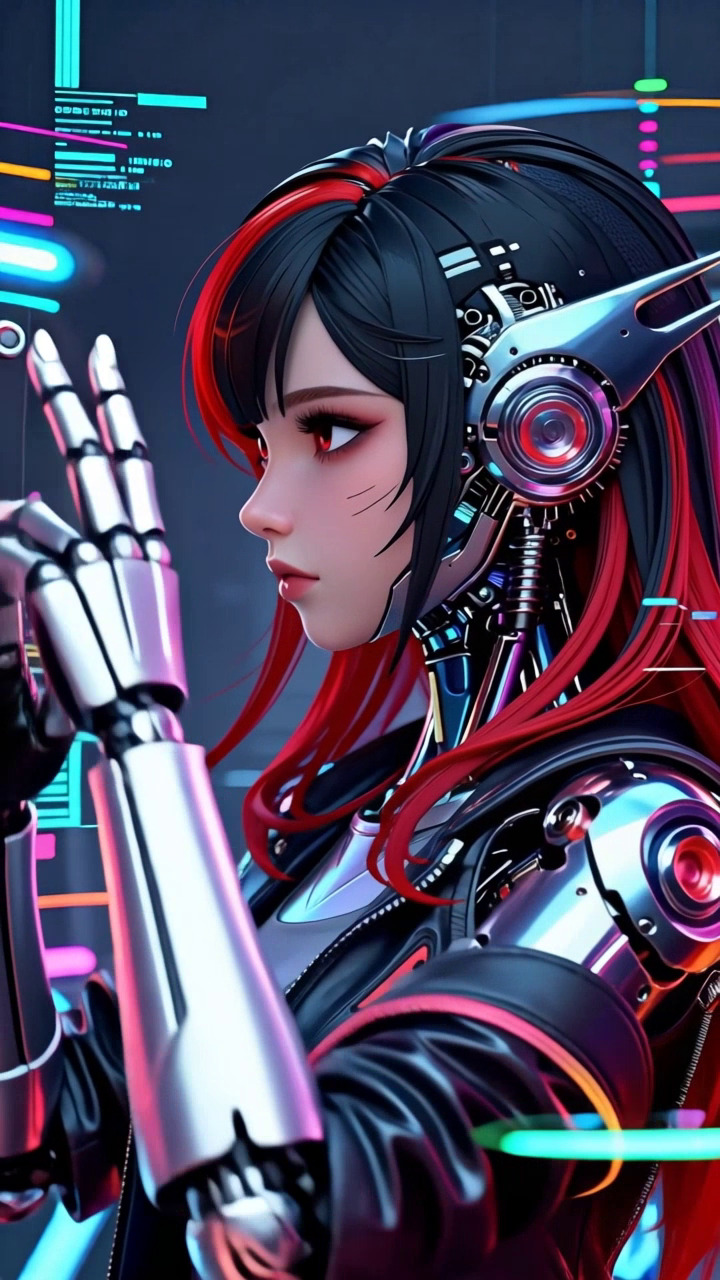} & \includegraphics[width=\fw]{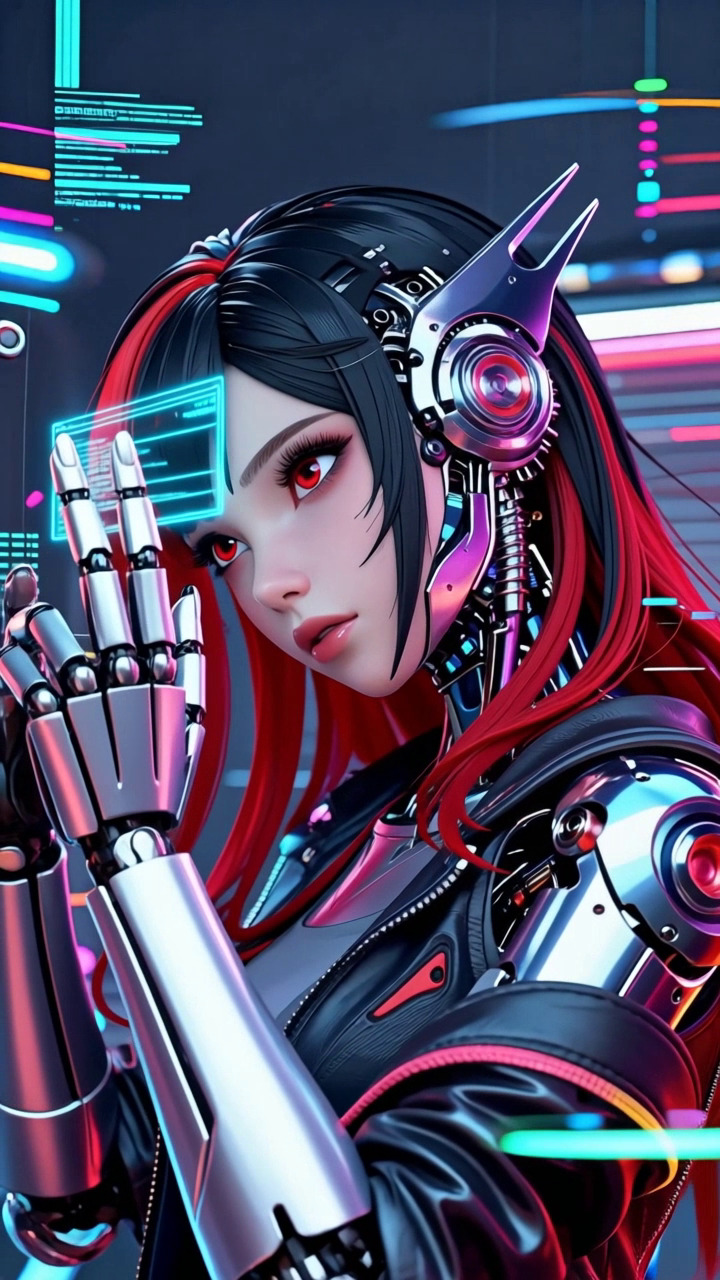} & \includegraphics[width=\fw]{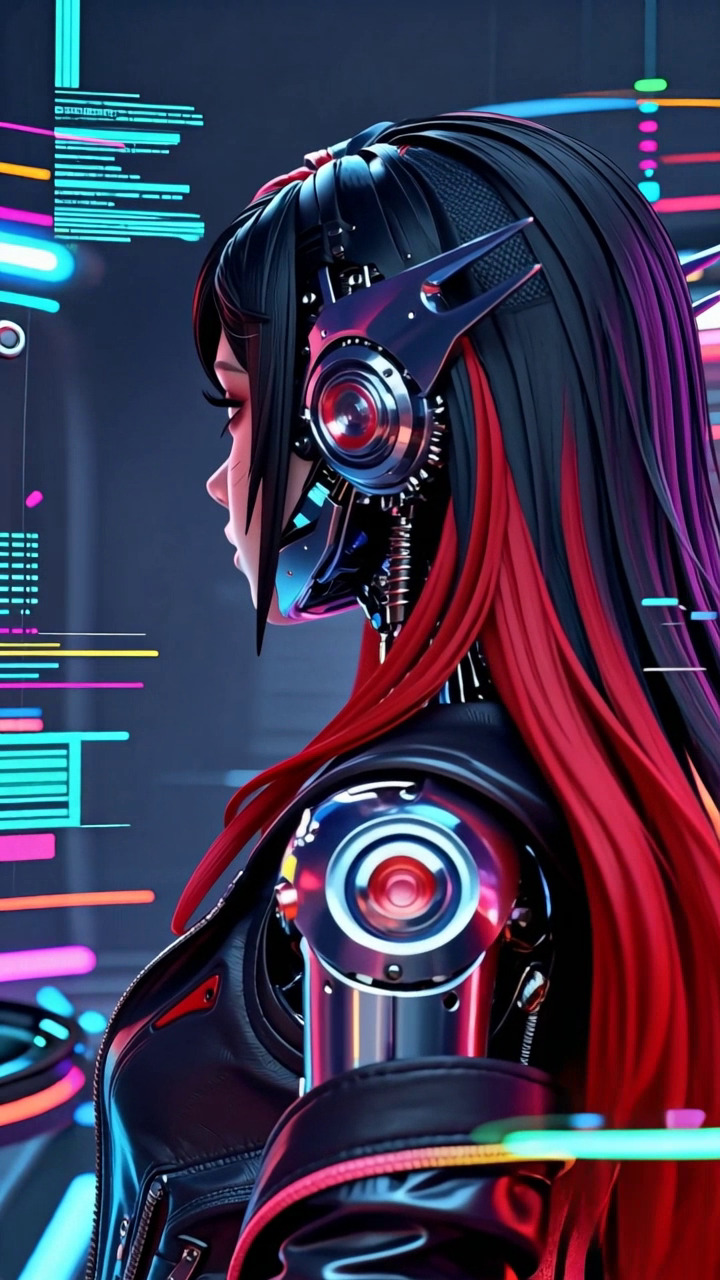}
      & \includegraphics[width=\fw]{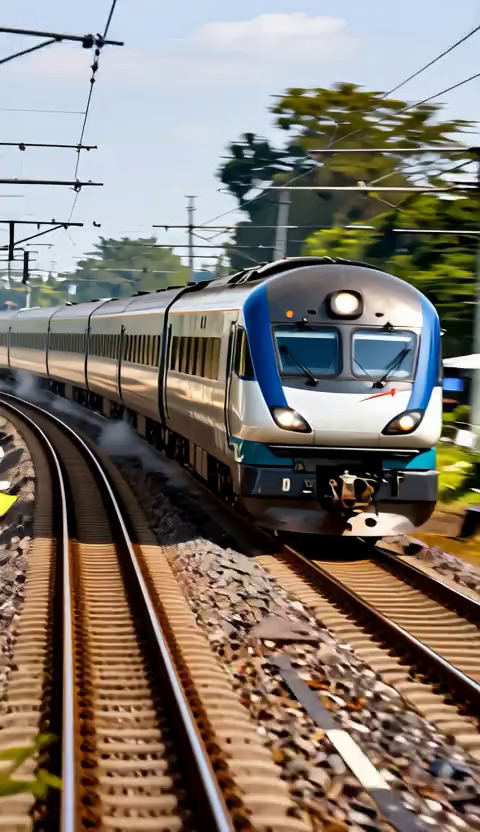}   & \includegraphics[width=\fw]{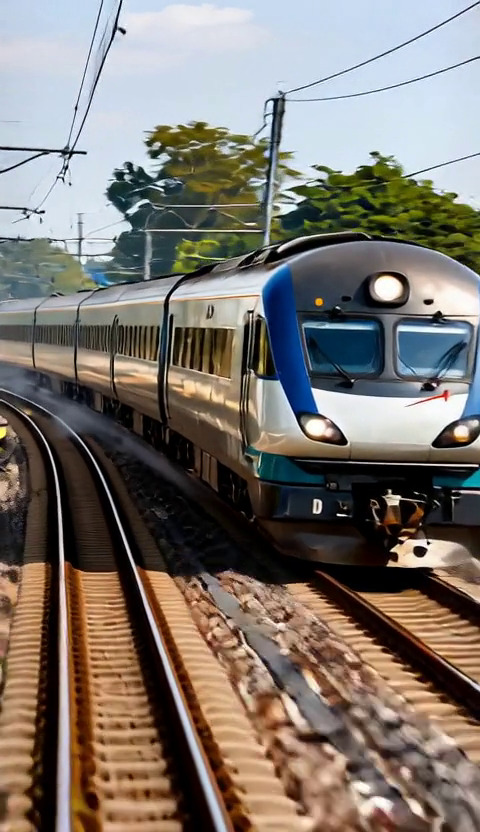}   & \includegraphics[width=\fw]{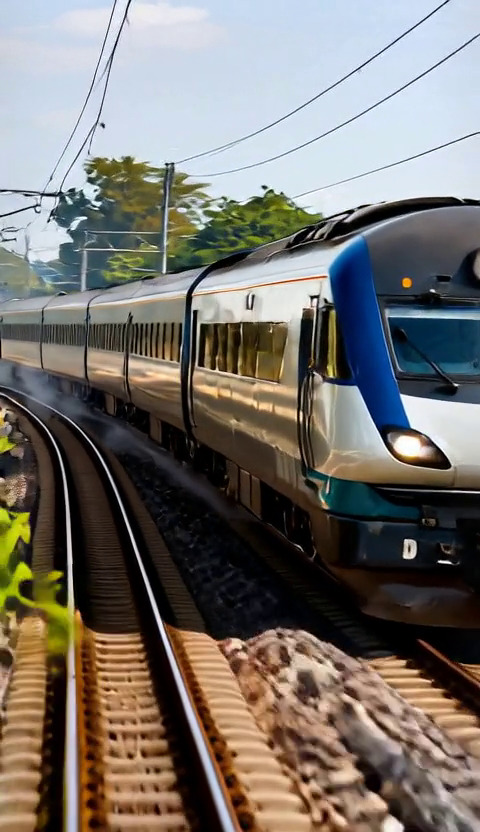}   & \includegraphics[width=\fw]{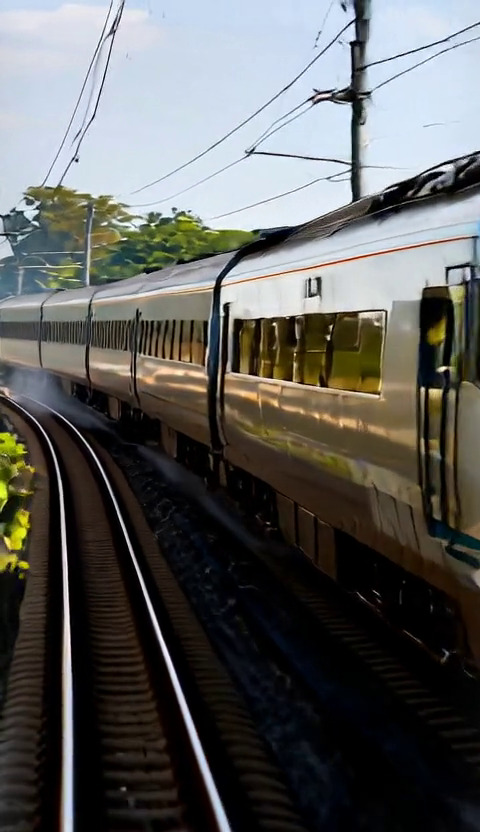}
      & \includegraphics[width=\fw]{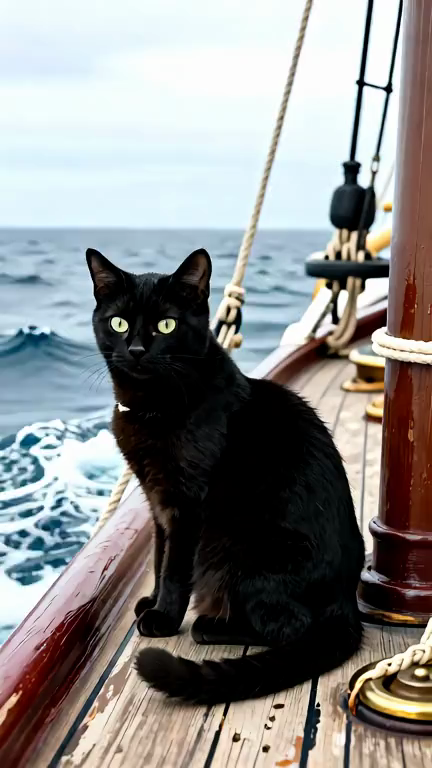}   & \includegraphics[width=\fw]{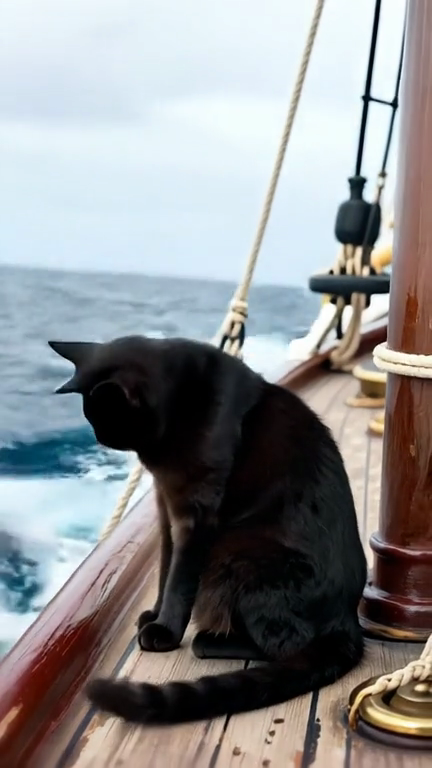}   & \includegraphics[width=\fw]{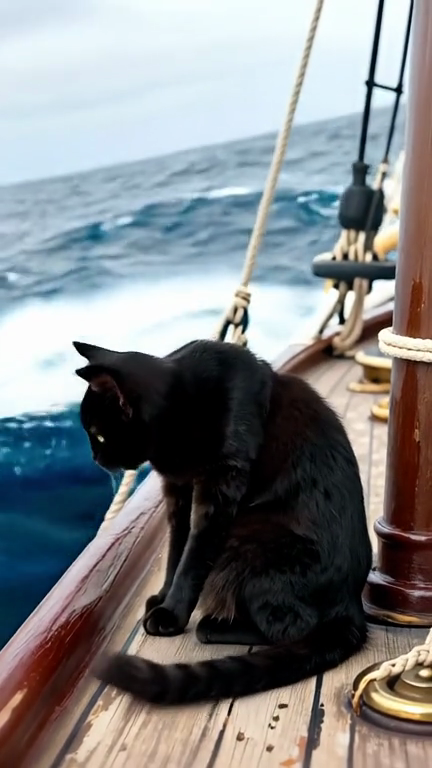}   & \includegraphics[width=\fw]{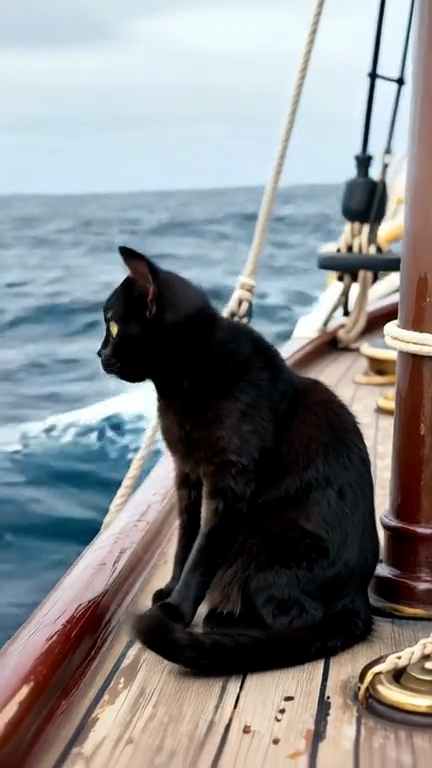} \\
  \end{tabular}}
  \vspace{0pt}
  \includegraphics[
    width=0.80\linewidth
  ]{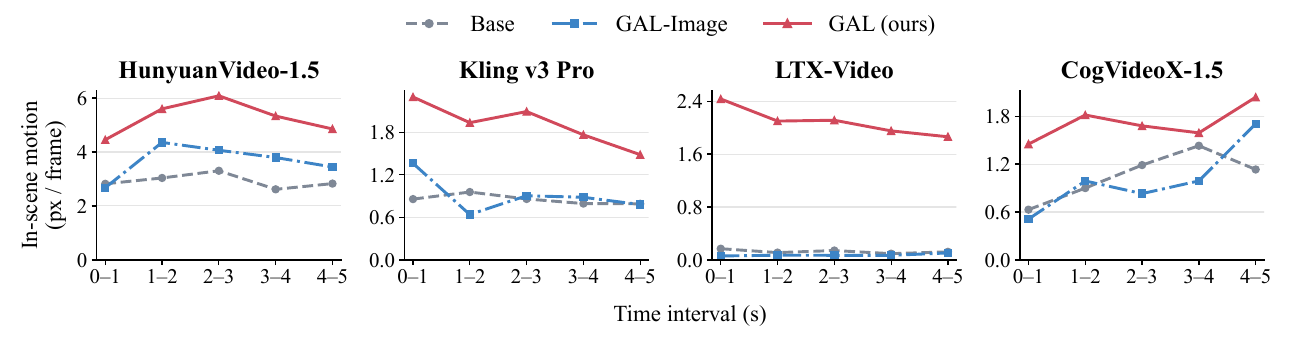}
  \fi
\caption{\textbf{In the shown examples, the GAL anchor induces more in-scene motion in every one-second interval across all four I2V models.} Top: within each I2V-model block, rows are clips rendered from Base (the anchor produced by the untuned image model), GAL-Image (trained based on image rewards), and our GAL anchor (trained based video rewards), under the same prompt and renderer; four frames span each clip. Bottom: line plots of per-interval motion for the same clips, measured as the mean residual optical-flow magnitude inside each one-second interval (\Cref{app:motion_trajectory_protocol}).}
\label{fig:qual_grid}
\end{figure}

\section{Introduction}
\label{sec:intro}

Video generation~\citep{kong2024hunyuanvideo,hacohen2024ltx,kling2025omni} now powers virtual avatars~\citep{Lin_2025_ICCV,tian2024emo,shao2025magicdistillationweaktostrongvideodistillation}, AI films~\citep{polyak2024movie,wu2025movieagent,wang-etal-2026-mavis}, and interactive world models~\citep{bruce2024genie,nvidia2025cosmos,valevski2025diffusion,DBLP:journals/corr/abs-2606-17800}. One practical approach is a text-to-image-to-video pipeline~\citep{10.1007/978-3-031-73033-7_12}: a text-to-image (T2I) model synthesizes an anchor image~\citep{podell2024sdxl,esser2024scaling,cai2025z}\iflatexml\stepcounter{footnote}\hypertarget{gal-anchor-note-ref}{}\hyperlink{gal-anchor-note}{\textsuperscript{\thefootnote}}\else\footnote{We use \emph{anchor} rather than \emph{first frame}: in many I2V systems, the conditioning image guides the generation without necessarily being the literal first frame of the output.}\fi, and an image-to-video (I2V) model animates it~\citep{blattmann2023stable,xing2024dynamicrafter,yang2025cogvideox}. The anchor strongly influences the visual content of the resulting clip~\citep{10.1007/978-3-031-73033-7_12,chen2026first}. Agentic pipelines inspect and refine keyframes~\citep{li2024anim,wang-etal-2026-mavis}, and evaluate and select candidate video clips~\citep{shi2025animaker}.
\iflatexml
\begin{flushleft}
\hypertarget{gal-anchor-note}{}\hyperlink{gal-anchor-note-ref}{\textsuperscript{1}}\,We use \emph{anchor} rather than \emph{first frame}: in many I2V systems, the conditioning image guides the generation without necessarily being the literal first frame of the output.
\end{flushleft}
\fi

Image-based anchor selection uses image quality as a proxy for downstream video quality. Candidate anchor images are ranked by aesthetics, prompt--image alignment~\citep{guo2024i4vgen}, or human judgment of the image. These criteria do not account for how the I2V model responds to the anchor. I2V models can over-rely on the conditioning image~\citep{zhao2024identifying,ge2025flashi2v}, and high-frequency detail exposed early in sampling can suppress motion~\citep{choi2025improving}.

\begin{wrapfigure}{r}{0.44\textwidth}
\centering
\includegraphics[width=\linewidth]{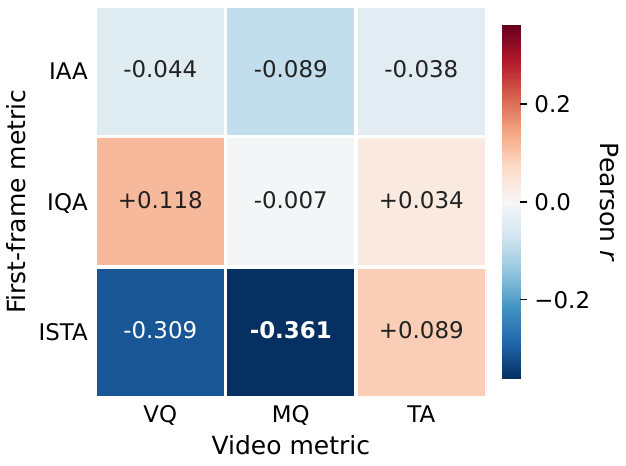}
\caption{\textbf{Mismatch between image scores and video rewards.} Pearson correlation between UniPercept scores of the anchor (IAA: aesthetics; IQA: quality; ISTA: structure and texture) and VideoReward scores of its HunyuanVideo-1.5 render (VQ: visual quality; MQ: motion quality; TA: text alignment) over $49{,}416$ (prompt--anchor--video) records. Aesthetics and quality are nearly uncorrelated with every video axis, and structure/texture is anti-correlated with visual and motion quality. }
\label{fig:reward_correlation}
\end{wrapfigure}

Our experiments show that image rewards do not reliably guide anchor optimization. Across $49{,}416$ (prompt--anchor--video) records, the anchor's aesthetic and quality scores are nearly uncorrelated with video rewards, while its structure/texture score is negatively correlated with the video's visual and motion quality (\Cref{fig:reward_correlation}). Fine-tuning the T2I generator on image-based preferences produces sharper anchors but lowers VideoReward-Overall on all four evaluated I2V models (\Cref{tab:reward_models,tab:reward_transfer}), with motion decreasing on some models and increasing on others (\Cref{tab:gap_internal}).

We therefore make the T2I generator itself \emph{video-aware}: an anchor should be optimized for the video it induces, not for its standalone appearance. We formalize this target through the notion of a \emph{golden anchor}: an anchor's value is determined by the quality of the video generated from it by a frozen I2V model. Golden anchor learning therefore optimizes the T2I generator for this downstream objective (\Cref{sec:gff_objective}; cf.\ \citealp{cheng2025vpo}). Our hypothesis is that the motion and quality biases of a video diffusion model can be injected upstream into the image prior of a T2I generator: the anchor is not a static preview but a control surface for downstream dynamics and quality.

\WFclear
\paragraph{Contributions.}
To our knowledge, this is the first work to improve image-to-video generation by learning a better anchor rather than selecting one~\citep{guo2024i4vgen}, and the first to formalize anchor generation as a learning problem (\Cref{eq:gff_objective}).  Concretely:

\begingroup
\setlength{\leftmargini}{1.2em}%
\begin{itemize}
\setlength{\itemsep}{2pt}

\item \textbf{Finding: image quality is misaligned with downstream video quality.}
Across $49{,}416$ (prompt--anchor--video) records, the UniPercept
structure-and-texture axis is negatively correlated with the
visual quality and motion quality axes of VideoReward
(\Cref{fig:reward_correlation}). This mismatch makes image rewards
unreliable objectives for anchor optimization. Anchors chosen by image-side criteria can therefore induce worse videos, so anchor supervision must come from the video itself.

\item \textbf{Method: Golden Anchor Learning (GAL).}
To our knowledge, GAL is the first framework to train a T2I generator using downstream video rewards from a frozen I2V model. It converts video scores into image-level preferences for offline DPO~\citep{rafailov2023direct,wallace2024diffusion}, requiring no additional rendering or scoring at inference.

\item \textbf{Result: GAL improves video quality and motion across I2V models.}
On all four I2V models, GAL raises the VBench average by $2.97\%$--$8.46\%$ over the untuned generator (Table~\ref{tab:unified_results_corr}), and the rendered clips score higher under VideoReward-Overall and their frames under HPSv3 and UniPercept (Table~\ref{tab:reward_main}). Motion gains are consistent across the four I2V models, with dynamic degree rising by $5.62\%$--$38.24\%$, alongside improvements in imaging and aesthetic quality (\Cref{tab:unified_results_corr}). GAL is trained on HunyuanVideo-1.5 renders alone, yet the gains transfer to LTX-Video, CogVideoX-1.5, and the closed-source Kling v3 Pro.

\end{itemize}
\endgroup

\section{Related Work}
\label{sec:related}
\textbf{Image-anchored video generation pipelines. }
Many high-quality video-generation workflows follow a T2I→I2V decomposition~\citep{10.1007/978-3-031-73033-7_12,li2024anim,wang-etal-2026-mavis}: a text-to-image model first synthesizes an anchor image~\citep{esser2024scaling,cai2025z}, which a downstream image-to-video model then animates into a clip~\citep{blattmann2023stable,xing2024dynamicrafter,zhang2023i2vgen}. Golden Noise learns prompt-conditioned initial noise~\citep{zhou2025golden}, while Zigzag sampling refines denoising trajectories~\citep{lichen2025zigzag} and CoRe$^2$ improves T2I quality and speed~\citep{shao2026improved}. Prior work shows how strongly the anchor governs the render: I2V diffusion models over-rely on the conditioning image at large timesteps and lose motion~\citep{zhao2024identifying,jeon2026rebalancing}, early exposure of its high-frequency detail steers sampling toward a static copy of the reference~\citep{choi2025improving}, and the first frame acts as a memory buffer for visual entities~\citep{chen2026first}. IV-Mixed Sampler combines image and video diffusion models during sampling~\citep{shao2025ivmixed}. Agentic pipelines also incorporate inference-time feedback: CoTriSyGen analyzes generated keyframes and videos to refine prompts and memory, and can trigger keyframe regeneration~\citep{yin2026closedloop}. These methods act through image conditioning, rendering dynamics, or inference-time feedback; our method, Golden Anchor Learning (GAL), instead trains the upstream generator to shift the anchor distribution, so the two are complementary. The anchor influences the clip yet is not the final artifact, so its value should be assessed through the video it induces, which we formalize next (\Cref{sec:gff_objective}).

\textbf{Preference learning across image and video.}
DPO and its variants \citep{rafailov2023direct,azar2024general,meng2024simpo} give an offline alternative to PPO-style preference optimization, and Diffusion-DPO carries the idea to image diffusion \citep{wallace2024diffusion,10.1145/3796982}. The reward signal is usually image-domain~\citep{sun2026craft}, however: HPSv2/v3 and UniPercept score images \citep{wu2023human,ma2025hpsv3,cao2026unipercept}. Optimizing a T2I generator with these rewards can improve anchor appearance, but its effect on downstream video motion is unstable across I2V models. Video-domain rewards such as VideoReward instead score rendered clips \citep{NEURIPS2025_76227feb}. Recent work optimizes video generators directly from preferences at the clip, segment, or physical level \citep{liu2025videodpo,wu2026densedpo,chen2026hierarchical}. Closest in structure, VPO tunes an upstream prompt rewriter with preference learning driven by video-level feedback from the rendered clips, and the rewriter generalizes across video generators~\citep{cheng2025vpo}; we instead optimize the anchor image, on which the I2V model over-relies. Our setting is complementary and makes the image-vs-video distinction central: because the anchor is not the deployed artifact, we keep the I2V model frozen and optimize the T2I generator using rendered-video preferences.

\section{Problem Setup: The Golden Anchor of an I2V Render}
\label{sec:prelim}
\label{sec:gff_objective}

We view a T2I$\to$I2V pipeline as a two-stage conditional generator. Given a prompt $p$, the upstream image generator $\pi_\psi$ samples a candidate anchor $x \sim \pi_\psi(\cdot \mid p)$, which then conditions a \emph{frozen} downstream I2V model $G$ to produce a rendered clip $V = G(x,p,z)$, where $z\sim\mathcal{N}(0,1)$ denotes i.i.d. I2V noise, interpreted elementwise. We call $x$ the \emph{anchor} of the render: unlike standard T2I preference learning, $x$ is not the final output, and its downstream value is assessed through the video it induces. For a fixed I2V model $G$ and video reward $R_v$, the downstream value of an anchor is
\begin{equation}
\label{eq:downstream_value}
O(x,p)
=
\E_{z\sim\mathcal{N}(0,1)}
\left[
R_v\bigl(G(x,p,z)\bigr)
\right],
\end{equation}
For fixed, frozen $(G,R_v)$ and prompt $p$, a \emph{golden anchor} is an image with high $O(x,p)$: it tends to induce highly rewarded videos across rendering noise. The \emph{golden anchor objective} shifts the T2I distribution toward these anchors:
\begin{equation}
\label{eq:gff_objective}
\max_{\psi}\;
\E_{p}
\E_{x\sim\pi_\psi(\cdot\mid p)}
\left[O(x,p)\right].
\end{equation}

Estimating $O(x,p)$ requires rendering and scoring videos; HunyuanVideo-1.5 renders one five-second clip in roughly five minutes on a single RTX 4090. This makes online reinforcement learning~\citep{black2024training,shao2024deepseekmath,liu2026flow,zheng2025diffusionnft,wang2026affordancer1,zhang2025hoidr1reinforcementlearningopenworld} expensive in our setting, because collecting fresh rollouts repeatedly invokes the I2V model. We therefore adopt an offline strategy: we pay the video-reward scorer once to build a preference dataset over \emph{sibling anchors}, candidate images sampled for the same prompt, then optimize $\pi_\psi$ against these fixed preferences with DPO, keeping the expensive I2V renderer out of the training loop (\Cref{sec:method}).

\section{GAL: Golden Anchor Learning}
\label{sec:method}

GAL first builds a fixed preference dataset
$\mathcal{D}_{\mathrm{pref}}$ of tuples $(p,x_w,x_l)$
labeled in video space (\Cref{sec:pipeline}), then optimizes the upstream image generator against these preferences while keeping the I2V renderer and video reward frozen (\Cref{sec:pref_distill}). \Cref{alg:gff_dpo} summarizes the procedure.

\subsection{Constructing cross-modal preferences}
\label{sec:pipeline}

\begin{figure*}[t]
\centering
\iflatexml
\includegraphics[width=520bp]{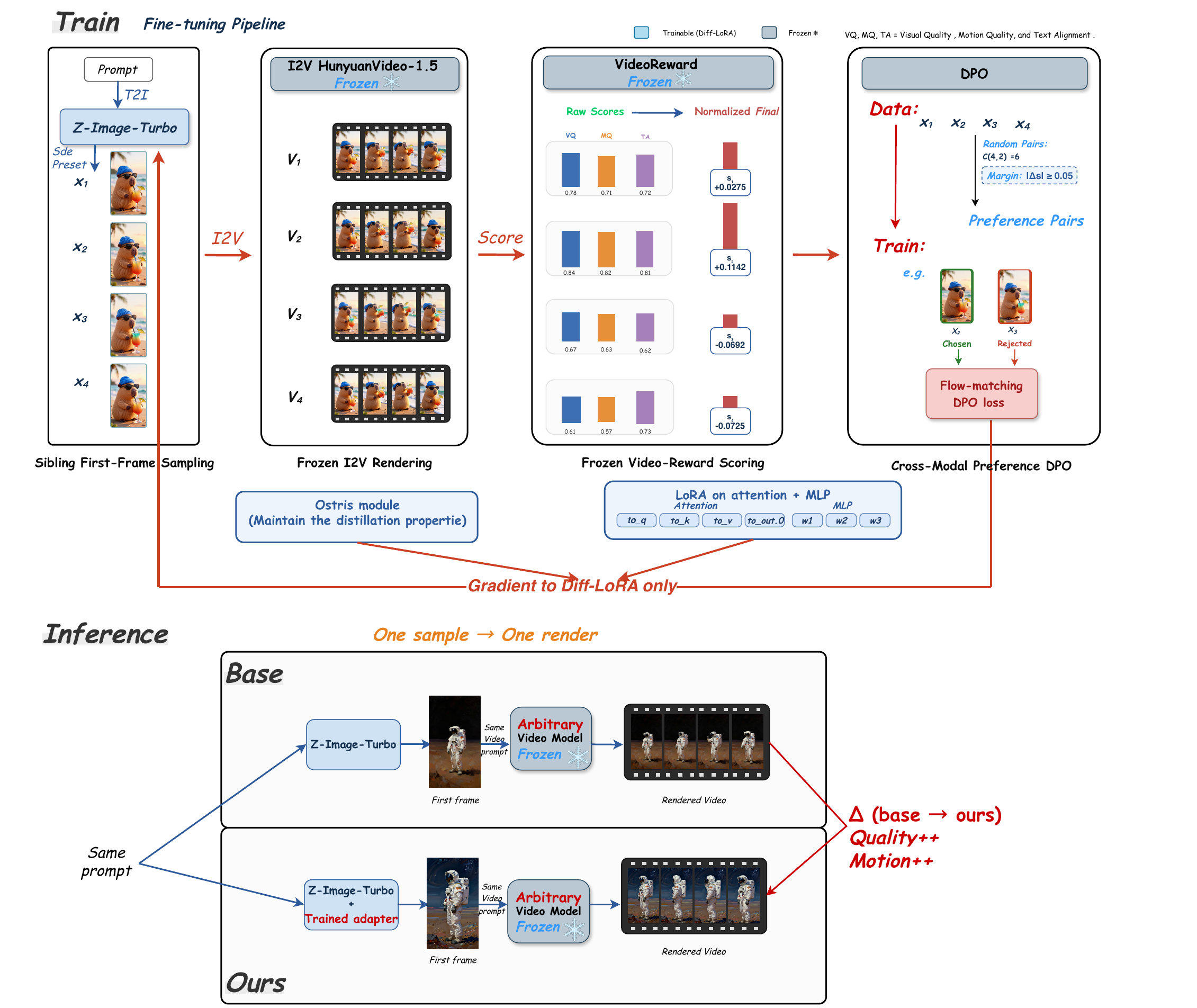}
\else
\makebox[\textwidth][c]{\includegraphics[width=0.9\textwidth]{figures/diff_lora_dpo_pipeline_1200dpi.pdf}}
\fi
\caption{\textbf{Overview of GAL. Training (top):} For each prompt, we sample $K=4$ anchor images from the T2I generator, render each into a video with a frozen I2V model, and score the videos with a frozen video-reward model. We use these scores to construct image-level preference pairs and fine-tune only the upstream T2I generator with cross-modal DPO. Video rendering, scoring, and pair construction are performed offline. \textbf{Inference (bottom):} The fine-tuned T2I generator produces anchor images that improve video quality, while the downstream I2V model remains unchanged.}
\label{fig:pipeline}
\end{figure*}

\Cref{fig:pipeline} shows how GAL converts downstream video scores into image-level preferences. For each prompt $p$, we draw $K=4$ sibling anchors from the initial policy $\pi_{\psi_0}$ using the reverse-time SDE in \Cref{app:sde}. We use the untuned Z-Image-Turbo model as the initial policy. Independent Brownian paths make the sampling trajectories stochastic and provide the within-prompt diversity needed for pairwise comparison~\citep{song2020score}.

A frozen I2V model renders each anchor once,
$V_i=G(x_i,p,z_i)$~\citep{wu2025hunyuanvideo}, and a frozen
video-reward model assigns $s_i=R_v(V_i)$
~\citep{NEURIPS2025_76227feb}. We use the scalar \emph{Overall} score for
training and retain the per-axis scores only for analysis. For a
higher-scoring anchor $x_w$ and a lower-scoring anchor $x_l$, we keep
the pair and write
\begin{equation}
\label{eq:pref_label}
x_w \succ_p x_l
\quad\text{whenever}\quad
s_w-s_l\ge m.
\end{equation}
The margin removes near-ties that are more sensitive to I2V sampling
noise. The compared samples remain images, while their labels are
determined by rendered videos.

\subsection{Preference optimization via DPO}
\label{sec:pref_distill}

For an anchor image $x$, let $y_0=\operatorname{Enc}(x)$ be its clean
latent. Draw a shared noising context
$c=\{(\tau_j,\epsilon_j)\}_{j=1}^{M}\sim q_c$ from the base scheduler,
and let
$y_{\tau_j}
=\operatorname{NoiseScheduler}(y_0,\epsilon_j,\tau_j)$.
Using the flow-matching residual~\citep{lipman2023flow,liu2022flow},
define
\begin{equation}
\label{eq:dpo_fm_loss}
\ell_\psi(x,p\iflatexml\text{;}\else;\fi c)
=\frac{1}{M}\sum_{j=1}^{M}w(\tau_j)
\Bigl\|
\widehat{u}_\psi(y_{\tau_j},\tau_j,p)-u_{\tau_j}(y_0,\epsilon_j)
\Bigr\|_2^2.
\end{equation}
Here $\widehat{u}_\psi$ is the trainable model prediction,
$u_{\tau_j}$ is the target velocity, and $\ell_{\mathrm{ref}}$ is
defined identically for the frozen reference model. We reuse the same
context $c$ for the winner, the loser, the trainable model, and the
reference model.

Adapting DPO and Diffusion-DPO
~\citep{rafailov2023direct,wallace2024diffusion,NEURIPS2025_76227feb} to the flow-matching loss above, we define
$\Delta_\psi(x,p\iflatexml\text{;}\else;\fi c)
=\ell_{\mathrm{ref}}(x,p\iflatexml\text{;}\else;\fi c)-\ell_\psi(x,p\iflatexml\text{;}\else;\fi c)$ and
$d_\psi
=\Delta_\psi(x_w,p\iflatexml\text{;}\else;\fi c)-\Delta_\psi(x_l,p\iflatexml\text{;}\else;\fi c)$.
The GAL objective is
\begin{equation}
\label{eq:dpo_loss}
\mathcal{L}_{\mathrm{DPO}}(\psi)
=
-\mathbb{E}_{(p,x_w,x_l)\sim\mathcal{D}_{\mathrm{pref}},\;c\sim q_c}
\left[\log\sigma\!\left(\beta d_\psi\right)\right].
\end{equation}
The loss decreases when fine-tuning reduces the flow-matching residual
more for the video-preferred anchor than for the rejected anchor.
The renderer and reward model are used only to build
$\mathcal{D}_{\mathrm{pref}}$.

\iflatexml
\begin{algorithm}[t]
\caption{GAL: Golden Anchor Learning}
\label{alg:gff_dpo}
\begingroup
\normalfont\normalsize
\setlength{\tabcolsep}{0pt}
\renewcommand{\arraystretch}{1}

\begin{tabular}{@{}p{2.2em}|@{}l@{}l@{}l@{}l@{}}
 & & \multicolumn{3}{@{}l@{}}{\textbf{Require:} Prompt set $\mathcal{P}$; pretrained T2I policy
$\pi_{\psi_0}$; frozen reference $\pi_{\mathrm{ref}}$;
frozen I2V renderer $G$; frozen video reward $R_v$;
T2I noising-context distribution $q_c$;
sibling count $K$; margin $m$;
DPO scale $\beta$; step size $\eta$}\\[5pt]

 & \makebox[0.65em][l]{} & \multicolumn{3}{@{}l@{}}{\textbf{Phase 1: Offline preference construction}}\\[2pt]
1: &  & \multicolumn{3}{@{}l@{}}{$\mathcal{D}_{\mathrm{pref}} \gets \varnothing$}\\
2: &  & \multicolumn{3}{@{}l@{}}{\textbf{for} $p \in \mathcal{P}$ \textbf{do}}\\
3: &  & \makebox[1.5em][l]{} & \multicolumn{2}{@{}l@{}}{Sample $K$ sibling anchor images $x_1,\ldots,x_K \sim \pi_{\psi_0}(\cdot \mid p)$}\\
 &  & & \multicolumn{2}{@{}l@{}}{via \Cref{eq:reverse_sde} with independent Brownian paths}\\
4: &  & & \multicolumn{2}{@{}l@{}}{\textbf{for} $k = 1,\ldots,K$ \textbf{do}}\\
5: &  & & \makebox[1.5em][l]{} & $z_k \sim \mathcal{N}(0,1)$\\
6: &  & & & $V_k \gets G(x_k,p,z_k)$\\
7: &  & & & $s_k \gets R_v(V_k)$\\
8: &  & & \multicolumn{2}{@{}l@{}}{\textbf{end for}}\\
9: &  & & \multicolumn{2}{@{}l@{}}{\textbf{for all} ordered pairs $(i,j)$ satisfying $s_i-s_j \ge m$ \textbf{do}}\\
10: &  & & & $\mathcal{D}_{\mathrm{pref}} \gets$ $\mathcal{D}_{\mathrm{pref}} \cup \{(p,x_i,x_j)\}$\\
11: &  & & \multicolumn{2}{@{}l@{}}{\textbf{end for}}\\
12: &  & \multicolumn{3}{@{}l@{}}{\textbf{end for}}\\[5pt]
 &  & \multicolumn{3}{@{}l@{}}{\textbf{Phase 2: T2I preference optimization}}\\[2pt]
13: &  & \multicolumn{3}{@{}l@{}}{$\psi \gets \psi_0$}\\
14: &  & \multicolumn{3}{@{}l@{}}{\textbf{while} not converged \textbf{do}}\\
15: &  & & \multicolumn{2}{@{}l@{}}{Sample $(p,x_w,x_l) \sim \mathcal{D}_{\mathrm{pref}}$}\\
16: &  & & \multicolumn{2}{@{}l@{}}{Sample a shared T2I noising context $c \sim q_c$}\\
17: &  & & \multicolumn{2}{@{}l@{}}{Evaluate $\ell_\psi$ and $\ell_{\mathrm{ref}}$ for both anchors under the same $c$ using \Cref{eq:dpo_fm_loss}}\\
18: &  & & \multicolumn{2}{@{}l@{}}{$d_\psi \gets$ $[\ell_{\mathrm{ref}}(x_w,p\text{;}c)$ $-\ell_\psi(x_w,p\text{;}c)]$ $-[\ell_{\mathrm{ref}}(x_l,p\text{;}c)$ $-\ell_\psi(x_l,p\text{;}c)]$}\\
19: &  & & \multicolumn{2}{@{}l@{}}{$\psi \gets$ $\psi-\eta\nabla_\psi[-\log\sigma(\beta d_\psi)]$}\\
20: &  & \multicolumn{3}{@{}l@{}}{\textbf{end while}}\\
21: &  & \multicolumn{3}{@{}l@{}}{\textbf{return} $\pi_\psi$}\\
\end{tabular}
\endgroup
\end{algorithm}
\else
\begin{algorithm}[t]
\caption{GAL: Golden Anchor Learning}
\label{alg:gff_dpo}
\begin{algorithmic}[1]

\Require Prompt set $\mathcal{P}$; pretrained T2I policy
         $\pi_{\psi_0}$; frozen reference $\pi_{\mathrm{ref}}$;
         frozen I2V renderer $G$; frozen video reward $R_v$;
         T2I noising-context distribution $q_c$;
         sibling count $K$; margin $m$;
         DPO scale $\beta$; step size $\eta$

\Statex \textbf{Phase 1: Offline preference construction}
\State $\mathcal{D}_{\mathrm{pref}} \gets \varnothing$

\For{$p \in \mathcal{P}$}
    \State Sample $K$ sibling anchor images $x_1,\ldots,x_K \sim \pi_{\psi_0}(\cdot \mid p)$
    \Statex \hspace{\algorithmicindent}via \Cref{eq:reverse_sde} with independent Brownian paths
    \For{$k = 1,\ldots,K$}
        \State $z_k \sim \mathcal{N}(0,1)$
        \State $V_k \gets G(x_k,p,z_k)$
        \State $s_k \gets R_v(V_k)$
    \EndFor
    \ForAll{ordered pairs $(i,j)$ satisfying $s_i-s_j \ge m$}
        \State $\mathcal{D}_{\mathrm{pref}} \gets
                \mathcal{D}_{\mathrm{pref}} \cup \{(p,x_i,x_j)\}$
    \EndFor
\EndFor

\Statex
\Statex \textbf{Phase 2: T2I preference optimization}
\State $\psi \gets \psi_0$

\While{not converged}
    \State Sample $(p,x_w,x_l) \sim \mathcal{D}_{\mathrm{pref}}$
    \State Sample a shared T2I noising context $c \sim q_c$
    \State Evaluate $\ell_\psi$ and $\ell_{\mathrm{ref}}$
           for both anchors under the same $c$ using \Cref{eq:dpo_fm_loss}
    \State $d_\psi \gets
        [\ell_{\mathrm{ref}}(x_w,p\iflatexml\text{;}\else;\fi c)-\ell_\psi(x_w,p\iflatexml\text{;}\else;\fi c)]
        -
        [\ell_{\mathrm{ref}}(x_l,p\iflatexml\text{;}\else;\fi c)-\ell_\psi(x_l,p\iflatexml\text{;}\else;\fi c)]$
    \State $\psi \gets
        \psi-\eta\nabla_\psi[-\log\sigma(\beta d_\psi)]$
\EndWhile

\State \Return $\pi_\psi$

\end{algorithmic}
\end{algorithm}
\fi
\subsection{Few-step implementation}
\label{sec:fewstep}
\label{sec:cdlora}        
\label{sec:eta-estimator} 
Adapting few-step distilled generators \citep{salimans2022progressive,song2023consistency,sauer2024adversarial,cai2025z,liu2026decoupled,bai2026optimizing,shao2026fastlightgen} can improve image metrics while degrading downstream motion. We train a task LoRA \citep{hu2022lora} within the differential-LoRA scaffold for Z-Image-Turbo \citep{ostris2025zimageadapter,kelseye2025trainingstrategies}. The frozen preset adapter is attached to the trainable and reference models during training and removed at inference (\Cref{app:implementation}); \Cref{eq:gff_objective} targets the deployed generator. Siblings are sampled without adapters; \Cref{app:proof_theory} discusses the collection/reference mismatch. Removing the scaffold under identical video-reward pairs reverses the motion gain despite image-side improvements (\Cref{sec:exp_cdlora_necessity,tab:gap_internal}).

\section{Analysis of Golden Anchor Learning}
\label{sec:analysis}

We motivate video-derived labels and margin filtering at the objective level. \Cref{app:proof_theory} gives supporting derivations, and \Cref{sec:exp_vbench} evaluates the trained system.

\paragraph{Why image and video feedback can disagree.}
The renderer maps an anchor to an entire clip, so an image change that improves an image score need not improve downstream value $O(x,p)$. For intuition, let $R_{\mathrm{img}}(x,p)$ be an image reward. Even when both objectives are differentiable, the image-improving direction $\nabla_x R_{\mathrm{img}}(x,p)$ need not align with $\nabla_x O(x,p)$: the latter also depends on how the frozen renderer responds to the anchor (\Cref{app:proof_theory}). This is an objective-level intuition, not the GAL update rule. GAL never computes either gradient; $O$ enters training only through preference labels estimated by rendering and scoring candidate anchors. The correlations and GAL-Image control in \Cref{sec:exp_vbench} provide empirical evidence of image--video misalignment, motivating video-derived labels.

\paragraph{The margin removes uncertain comparisons.}
A one-render score contains I2V sampling noise, so comparisons with small score gaps are the most vulnerable to order reversal. The margin $m$ in \Cref{eq:pref_label} removes these near-ties before training. Increasing $m$ yields a smaller but more separated preference set (\Cref{tab:margin_pairs}). We treat the margin as a practical noise-control choice, not as a proof that one render exactly estimates $O(x,p)$. The multi-seed evaluation in \Cref{sec:exp_vbench} tests whether the learned effect persists under downstream rendering stochasticity.
\section{Experiments}
\label{sec:experiments}

The experiments test whether rendered-video feedback produces more effective anchors than image-based feedback. \Cref{sec:exp_qualitative,sec:exp_quantitative} present the main comparison across four frozen I2V models, \Cref{sec:exp_reward_signal} selects the reward signal within GAL, and \Cref{sec:exp_ablation} ablates the training scaffold.

\subsection{Settings}
\label{sec:exp_settings}
\label{sec:exp_setup}

\paragraph{Variants.}
All fine-tuned variants share the sibling anchors and the LoRA-DPO recipe of Section~\ref{sec:method} and differ only in the signal that labels a pair. Video-supervised variants are named by their rewards: GAL-VideoReward uses VideoReward-Overall on the HunyuanVideo-1.5 render; GAL-HPSv3 and GAL-Unipercept use the corresponding image rewards on frames of that render; GAL-Image uses UniPercept and HPSv3 on the anchor image itself. Base is the untuned generator, and \mbox{GAL-VideoReward w/o scaffold} removes the preset adapter of Section~\ref{sec:fewstep}. GAL-VideoReward is our default and appears as GAL (ours) in \Cref{tab:unified_results_corr,tab:reward_main}. Hyperparameters are listed in Appendix~\ref{app:implementation}.

\paragraph{I2V models.}
HunyuanVideo-1.5 alone supplies labels. LTX-Video, CogVideoX-1.5, and the closed-source Kling v3 Pro never label a pair and test transfer to unseen renderers. We call these four signal-bearing: their VBench dynamic degree keeps motion headroom and responds to the anchor.

\paragraph{Evaluation.}
Each variant generates one anchor per prompt under a fixed T2I sampling setting. On the open-source renderers we use the official VBench per-dimension prompt sets~\citep{huang2024vbench} with one fixed video-instruction template, render each anchor under three I2V seeds, and average the VBench scores, to reduce the contribution of I2V sampling noise to the estimated anchor effect. Kling v3 Pro renders a held-out set of 50 prompts once. Reward-model scores (Table~\ref{tab:reward_main}) are computed on the rendered clips, with VideoReward-Overall on the clip and HPSv3 and UniPercept on its extracted frames. Training, VBench, and Kling prompts are pairwise disjoint (Appendix~\ref{app:vbench_prompt_protocol}).

\subsection{Qualitative Comparison}
\label{sec:exp_qual}
\label{sec:exp_qualitative}

\Cref{fig:qual_grid} provides a visual check of the motion claim. For each I2V model, it shows clips rendered from Base, GAL-Image, and GAL-VideoReward under the same prompt and renderer. In the shown examples, GAL-VideoReward produces larger frame-to-frame changes, and its per-interval in-scene motion is the highest of the three variants in all 20 one-second intervals across the four I2V models. This pattern matches the dynamic-degree gains in \Cref{tab:unified_results_corr}.

\subsection{Main Comparison across I2V Models}
\label{sec:exp_quantitative}
\label{sec:exp_vbench}



\begin{table}[t]
\centering
\fontsize{6}{7}\selectfont
\setlength{\tabcolsep}{1.5pt}\iflatexml\setlength{\tabcolsep}{4pt}\fi
\renewcommand{\arraystretch}{1.05}\iflatexml\renewcommand{\arraystretch}{1}\fi
\vspace{-3pt}
\caption{Main VBench comparison across four I2V models. GAL (ours), the GAL-VideoReward variant, improves the VBench average, dynamic degree, imaging quality, and aesthetic quality over Base on every model. Parentheses give percentage changes from Base; bold marks the best within each model. Open-source results average three I2V seeds; Kling v3 Pro uses one seed.}
\label{tab:unified_results_corr}
\vspace{2pt}
\begin{tabular*}{\textwidth}{@{\extracolsep{\fill}}l c c c c c c c@{}}
\toprule
{Variant} & \shortstack{Dynamic\\Degree} & \shortstack{Imaging\\Quality} & \shortstack{Aesthetic\\Quality} & \shortstack{Subject\\Consistency} & \shortstack{Motion\\Smoothness} & \shortstack{Overall\\Consistency} & \shortstack{VBench\\Average} \\
\midrule
\multicolumn{8}{l}{\emph{HunyuanVideo-1.5}}\\
{Base} & 0.8241 & 0.6586 & 0.6160 & 0.9577 & \textbf{0.9922} & 0.2247 & 0.7122 \\
\iflatexml
{\textbf{GAL~(ours)}} & \textbf{0.8704} & \textbf{0.7111} & \textbf{0.6426} & \textbf{0.9601} & 0.9909 & \textbf{0.2248} & \textbf{0.7333} \\
 & $(+5.62\%)$ & $(+7.97\%)$ & $(+4.32\%)$ & $(+0.25\%)$ & $(-0.13\%)$ & $(+0.07\%)$ & $(+2.97\%)$ \\[3pt]
\else
{\textbf{GAL (ours)}} & \textbf{0.8704} {($+5.62\%$)} & \textbf{0.7111} {($+7.97\%$)} & \textbf{0.6426} {($+4.32\%$)} & \textbf{0.9601} {($+0.25\%$)} & 0.9909 {($-0.13\%$)} & \textbf{0.2248} {($+0.07\%$)} & \textbf{0.7333} {($+2.97\%$)} \\
\fi
\midrule
\multicolumn{8}{l}{\emph{LTX-Video} (transfer)}\\
{Base} & 0.5509 & 0.6493 & 0.5909 & 0.9557 & \textbf{0.9922} & 0.2537 & 0.6654 \\
\iflatexml
{\textbf{GAL~(ours)}} & \textbf{0.6389} & \textbf{0.7076} & \textbf{0.6064} & \textbf{0.9613} & 0.9903 & \textbf{0.2551} & \textbf{0.6933} \\
 & $(+15.97\%)$ & $(+8.99\%)$ & $(+2.62\%)$ & $(+0.59\%)$ & $(-0.19\%)$ & $(+0.53\%)$ & $(+4.18\%)$ \\[3pt]
\else
{\textbf{GAL (ours)}} & \textbf{0.6389} {($+15.97\%$)} & \textbf{0.7076} {($+8.99\%$)} & \textbf{0.6064} {($+2.62\%$)} & \textbf{0.9613} {($+0.59\%$)} & 0.9903 {($-0.19\%$)} & \textbf{0.2551} {($+0.53\%$)} & \textbf{0.6933} {($+4.18\%$)} \\
\fi
\midrule
\multicolumn{8}{l}{\emph{CogVideoX-1.5} (transfer)}\\
{Base} & 0.7222 & 0.6121 & 0.5868 & 0.9203 & \textbf{0.9716} & 0.2490 & 0.6770 \\
\iflatexml
{\textbf{GAL~(ours)}} & \textbf{0.7731} & \textbf{0.6749} & \textbf{0.6016} & \textbf{0.9219} & 0.9633 & \textbf{0.2527} & \textbf{0.6979} \\
 & $(+7.05\%)$ & $(+10.26\%)$ & $(+2.52\%)$ & $(+0.17\%)$ & $(-0.86\%)$ & $(+1.45\%)$ & $(+3.09\%)$ \\[3pt]
\else
{\textbf{GAL (ours)}} & \textbf{0.7731} {($+7.05\%$)} & \textbf{0.6749} {($+10.26\%$)} & \textbf{0.6016} {($+2.52\%$)} & \textbf{0.9219} {($+0.17\%$)} & 0.9633 {($-0.86\%$)} & \textbf{0.2527} {($+1.45\%$)} & \textbf{0.6979} {($+3.09\%$)} \\
\fi
\midrule
\multicolumn{8}{l}{\emph{Kling v3 Pro} (closed-source transfer)}\\
{Base} & 0.6800 & 0.6721 & 0.6607 & 0.9576 & \textbf{0.9928} & 0.2434 & 0.7011 \\
\iflatexml
{\textbf{GAL~(ours)}} & \textbf{0.9400} & \textbf{0.7263} & \textbf{0.7018} & \textbf{0.9609} & 0.9888 & \textbf{0.2448} & \textbf{0.7604} \\
 & $(+38.24\%)$ & $(+8.07\%)$ & $(+6.22\%)$ & $(+0.35\%)$ & $(-0.41\%)$ & $(+0.57\%)$ & $(+8.46\%)$ \\[3pt]
\else
{\textbf{GAL (ours)}} & \textbf{0.9400} {($+38.24\%$)} & \textbf{0.7263} {($+8.07\%$)} & \textbf{0.7018} {($+6.22\%$)} & \textbf{0.9609} {($+0.35\%$)} & 0.9888 {($-0.41\%$)} & \textbf{0.2448} {($+0.57\%$)} & \textbf{0.7604} {($+8.46\%$)} \\
\fi
\bottomrule
\end{tabular*}
\vspace{-3pt}
\end{table}

\begin{table}[t]
\centering
\fontsize{6}{7}\selectfont
\setlength{\tabcolsep}{1.5pt}\iflatexml\setlength{\tabcolsep}{4pt}\fi
\renewcommand{\arraystretch}{1.05}\iflatexml\renewcommand{\arraystretch}{1}\fi
\vspace{-3pt}
\caption{Main reward-model comparison across four I2V models. GAL (ours) improves all three reward scores over Base on every model. VideoReward-Overall scores rendered clips; HPSv3 and UniPercept score their extracted frames. Parentheses give absolute changes from Base.}
\label{tab:reward_main}
\vspace{2pt}
\begin{tabular*}{\textwidth}{@{\extracolsep{\fill}}l l c c c@{}}
\toprule
I2V model & Variant & \shortstack{VideoReward\\Overall} & \shortstack{HPSv3\\(video frames)} & \shortstack{UniPercept\\(video frames)} \\
\midrule
\emph{HunyuanVideo-1.5} & Base & $-0.014$ & 7.0025 & 51.4375 \\
\iflatexml
 & \textbf{GAL~(ours)} & $\bm{+0.131}$ & \textbf{7.6286} & \textbf{55.5926} \\
 &  & $(+0.145)$ & $(+0.6261)$ & $(+4.1550)$ \\[3pt]
\else
 & \textbf{GAL (ours)} & $\bm{+0.131}$ {($+0.145$)} & \textbf{7.6286} {($+0.6261$)} & \textbf{55.5926} {($+4.1550$)} \\
\fi
\midrule
\emph{LTX-Video} (transfer) & Base & $-0.328$ & 6.4788 & 50.0465 \\
\iflatexml
 & \textbf{GAL~(ours)} & $\bm{-0.224}$ & \textbf{7.3407} & \textbf{54.1946} \\
 &  & $(+0.103)$ & $(+0.8619)$ & $(+4.1481)$ \\[3pt]
\else
 & \textbf{GAL (ours)} & $\bm{-0.224}$ {($+0.103$)} & \textbf{7.3407} {($+0.8619$)} & \textbf{54.1946} {($+4.1481$)} \\
\fi
\midrule
\emph{CogVideoX-1.5} (transfer) & Base & $-0.284$ & 5.1392 & 46.5880 \\
\iflatexml
 & \textbf{GAL~(ours)} & $\bm{-0.129}$ & \textbf{6.1601} & \textbf{51.4183} \\
 &  & $(+0.155)$ & $(+1.0209)$ & $(+4.8303)$ \\[3pt]
\else
 & \textbf{GAL (ours)} & $\bm{-0.129}$ {($+0.155$)} & \textbf{6.1601} {($+1.0209$)} & \textbf{51.4183} {($+4.8303$)} \\
\fi
\midrule
\emph{Kling v3 Pro} (closed-source transfer) & Base & $+0.394$ & 8.2367 & 59.5410 \\
\iflatexml
 & \textbf{GAL~(ours)} & $\bm{+0.909}$ & \textbf{8.9403} & \textbf{62.4870} \\
 &  & $(+0.515)$ & $(+0.7035)$ & $(+2.9460)$ \\[3pt]
\else
 & \textbf{GAL (ours)} & $\bm{+0.909}$ {($+0.515$)} & \textbf{8.9403} {($+0.7035$)} & \textbf{62.4870} {($+2.9460$)} \\
\fi
\bottomrule
\end{tabular*}
\vspace{-3pt}
\end{table}

\begin{table}[t]
\centering
\fontsize{6}{7}\selectfont
\setlength{\tabcolsep}{1pt}\iflatexml\setlength{\tabcolsep}{4pt}\fi
\renewcommand{\arraystretch}{1.05}\iflatexml\renewcommand{\arraystretch}{1}\fi
\vspace{-3pt}
\caption{Reward-signal ablation within GAL, evaluated with VBench across four I2V models. Parentheses give percentage changes from Base; bold marks the best within each model. \mbox{\galhtmlname{GAL-VideoReward}} is the default variant.}
\label{tab:gap_internal}
\vspace{2pt}
\begin{tabular*}{\textwidth}{@{\extracolsep{\fill}}l c c c c c c c@{}}
\toprule
{Variant} & \shortstack{Dynamic\\Degree} & \shortstack{Imaging\\Quality} & \shortstack{Aesthetic\\Quality} & \shortstack{Subject\\Consistency} & \shortstack{Motion\\Smoothness} & \shortstack{Overall\\Consistency} & \shortstack{VBench\\Average} \\
\midrule
\multicolumn{8}{l}{\emph{HunyuanVideo-1.5}}\\
{Base} & 0.8241 & 0.6586 & 0.6160 & 0.9577 & \textbf{0.9922} & 0.2247 & 0.7122 \\
\iflatexml
{\galhtmlname{GAL-Image}} & 0.7917 & \textbf{0.7302} & 0.6453 & 0.9568 & 0.9902 & 0.2230 & 0.7229 \\
 & $(-3.93\%)$ & $(+10.87\%)$ & $(+4.76\%)$ & $(-0.09\%)$ & $(-0.20\%)$ & $(-0.76\%)$ & $(+1.50\%)$ \\[3pt]
\else
{\galhtmlname{GAL-Image}} & 0.7917{($-3.93\%$)} & \textbf{0.7302}{($+10.87\%$)} & 0.6453{($+4.76\%$)} & 0.9568{($-0.09\%$)} & 0.9902{($-0.20\%$)} & 0.2230{($-0.76\%$)} & 0.7229{($+1.50\%$)} \\
\fi
\iflatexml
{\galhtmlname{GAL-HPSv3}} & \textbf{0.8796} & 0.7035 & 0.6310 & 0.9529 & 0.9882 & \textbf{0.2248} & 0.7300 \\
 & $(+6.74\%)$ & $(+6.82\%)$ & $(+2.43\%)$ & $(-0.50\%)$ & $(-0.41\%)$ & $(+0.03\%)$ & $(+2.50\%)$ \\[3pt]
\else
{\galhtmlname{GAL-HPSv3}} & \textbf{0.8796}{($+6.74\%$)} & 0.7035{($+6.82\%$)} & 0.6310{($+2.43\%$)} & 0.9529{($-0.50\%$)} & 0.9882{($-0.41\%$)} & \textbf{0.2248}{($+0.03\%$)} & 0.7300{($+2.50\%$)} \\
\fi
\iflatexml
{\galhtmlname{GAL-Unipercept}} & 0.8380 & 0.7038 & \textbf{0.6638} & 0.9564 & 0.9911 & 0.2223 & 0.7292 \\
 & $(+1.69\%)$ & $(+6.86\%)$ & $(+7.75\%)$ & $(-0.13\%)$ & $(-0.11\%)$ & $(-1.07\%)$ & $(+2.39\%)$ \\[3pt]
\else
{\galhtmlname{GAL-Unipercept}} & 0.8380{($+1.69\%$)} & 0.7038{($+6.86\%$)} & \textbf{0.6638}{($+7.75\%$)} & 0.9564{($-0.13\%$)} & 0.9911{($-0.11\%$)} & 0.2223{($-1.07\%$)} & 0.7292{($+2.39\%$)} \\
\fi
\iflatexml
{\textbf{\galhtmlname{GAL-VideoReward}}} & 0.8704 & 0.7111 & 0.6426 & \textbf{0.9601} & 0.9909 & \textbf{0.2248} & \textbf{0.7333} \\
 & $(+5.62\%)$ & $(+7.97\%)$ & $(+4.32\%)$ & $(+0.25\%)$ & $(-0.13\%)$ & $(+0.07\%)$ & $(+2.97\%)$ \\[3pt]
\else
{\textbf{\galhtmlname{GAL-VideoReward}}} & 0.8704{($+5.62\%$)} & 0.7111{($+7.97\%$)} & 0.6426{($+4.32\%$)} & \textbf{0.9601}{($+0.25\%$)} & 0.9909{($-0.13\%$)} & \textbf{0.2248}{($+0.07\%$)} & \textbf{0.7333}{($+2.97\%$)} \\
\fi
\iflatexml
{\quad$\hookrightarrow$\,w/o scaffold} & 0.8056 & 0.7147 & 0.6413 & 0.9599 & 0.9911 & 0.2241 & 0.7228 \\
 & $(-2.25\%)$ & $(+8.52\%)$ & $(+4.11\%)$ & $(+0.23\%)$ & $(-0.12\%)$ & $(-0.24\%)$ & $(+1.48\%)$ \\[3pt]
\else
{\quad$\hookrightarrow$\,w/o scaffold} & 0.8056{($-2.25\%$)} & 0.7147{($+8.52\%$)} & 0.6413{($+4.11\%$)} & 0.9599{($+0.23\%$)} & 0.9911{($-0.12\%$)} & 0.2241{($-0.24\%$)} & 0.7228{($+1.48\%$)} \\
\fi
\midrule
\multicolumn{8}{l}{\emph{LTX-Video} (transfer)}\\
{Base} & 0.5509 & 0.6493 & 0.5909 & 0.9557 & \textbf{0.9922} & 0.2537 & 0.6654 \\
\iflatexml
{\galhtmlname{GAL-Image}} & 0.5278 & \textbf{0.7287} & 0.6211 & \textbf{0.9636} & 0.9899 & 0.2565 & 0.6813 \\
 & $(-4.20\%)$ & $(+12.23\%)$ & $(+5.13\%)$ & $(+0.83\%)$ & $(-0.23\%)$ & $(+1.11\%)$ & $(+2.38\%)$ \\[3pt]
\else
{\galhtmlname{GAL-Image}} & 0.5278{($-4.20\%$)} & \textbf{0.7287}{($+12.23\%$)} & 0.6211{($+5.13\%$)} & \textbf{0.9636}{($+0.83\%$)} & 0.9899{($-0.23\%$)} & 0.2565{($+1.11\%$)} & 0.6813{($+2.38\%$)} \\
\fi
\iflatexml
{\galhtmlname{GAL-HPSv3}} & 0.5278 & 0.7136 & 0.6148 & 0.9627 & 0.9897 & 0.2518 & 0.6767 \\
 & $(-4.20\%)$ & $(+9.90\%)$ & $(+4.06\%)$ & $(+0.73\%)$ & $(-0.25\%)$ & $(-0.75\%)$ & $(+1.70\%)$ \\[3pt]
\else
{\galhtmlname{GAL-HPSv3}} & 0.5278{($-4.20\%$)} & 0.7136{($+9.90\%$)} & 0.6148{($+4.06\%$)} & 0.9627{($+0.73\%$)} & 0.9897{($-0.25\%$)} & 0.2518{($-0.75\%$)} & 0.6767{($+1.70\%$)} \\
\fi
\iflatexml
{\galhtmlname{GAL-Unipercept}} & 0.5972 & 0.7052 & \textbf{0.6299} & 0.9575 & 0.9903 & \textbf{0.2575} & 0.6896 \\
 & $(+8.40\%)$ & $(+8.62\%)$ & $(+6.60\%)$ & $(+0.19\%)$ & $(-0.19\%)$ & $(+1.51\%)$ & $(+3.63\%)$ \\[3pt]
\else
{\galhtmlname{GAL-Unipercept}} & 0.5972{($+8.40\%$)} & 0.7052{($+8.62\%$)} & \textbf{0.6299}{($+6.60\%$)} & 0.9575{($+0.19\%$)} & 0.9903{($-0.19\%$)} & \textbf{0.2575}{($+1.51\%$)} & 0.6896{($+3.63\%$)} \\
\fi
\iflatexml
{\textbf{\galhtmlname{GAL-VideoReward}}} & \textbf{0.6389} & 0.7076 & 0.6064 & 0.9613 & 0.9903 & 0.2551 & \textbf{0.6933} \\
 & $(+15.97\%)$ & $(+8.99\%)$ & $(+2.62\%)$ & $(+0.59\%)$ & $(-0.19\%)$ & $(+0.53\%)$ & $(+4.18\%)$ \\[3pt]
\else
{\textbf{\galhtmlname{GAL-VideoReward}}} & \textbf{0.6389}{($+15.97\%$)} & 0.7076{($+8.99\%$)} & 0.6064{($+2.62\%$)} & 0.9613{($+0.59\%$)} & 0.9903{($-0.19\%$)} & 0.2551{($+0.53\%$)} & \textbf{0.6933}{($+4.18\%$)} \\
\fi
\midrule
\multicolumn{8}{l}{\emph{CogVideoX-1.5} (transfer)}\\
{Base} & 0.7222 & 0.6121 & 0.5868 & 0.9203 & \textbf{0.9716} & 0.2490 & 0.6770 \\
\iflatexml
{\galhtmlname{GAL-Image}} & 0.7546 & \textbf{0.6942} & 0.6088 & 0.9060 & 0.9529 & 0.2501 & 0.6944 \\
 & $(+4.49\%)$ & $(+13.41\%)$ & $(+3.75\%)$ & $(-1.55\%)$ & $(-1.92\%)$ & $(+0.44\%)$ & $(+2.57\%)$ \\[3pt]
\else
{\galhtmlname{GAL-Image}} & 0.7546{($+4.49\%$)} & \textbf{0.6942}{($+13.41\%$)} & 0.6088{($+3.75\%$)} & 0.9060{($-1.55\%$)} & 0.9529{($-1.92\%$)} & 0.2501{($+0.44\%$)} & 0.6944{($+2.57\%$)} \\
\fi
\iflatexml
{\galhtmlname{GAL-HPSv3}} & 0.7639 & 0.6772 & 0.5967 & 0.9089 & 0.9470 & 0.2493 & 0.6905 \\
 & $(+5.77\%)$ & $(+10.64\%)$ & $(+1.69\%)$ & $(-1.24\%)$ & $(-2.53\%)$ & $(+0.12\%)$ & $(+1.99\%)$ \\[3pt]
\else
{\galhtmlname{GAL-HPSv3}} & 0.7639{($+5.77\%$)} & 0.6772{($+10.64\%$)} & 0.5967{($+1.69\%$)} & 0.9089{($-1.24\%$)} & 0.9470{($-2.53\%$)} & 0.2493{($+0.12\%$)} & 0.6905{($+1.99\%$)} \\
\fi
\iflatexml
{\galhtmlname{GAL-Unipercept}} & 0.7639 & 0.6776 & \textbf{0.6213} & 0.9162 & 0.9601 & \textbf{0.2542} & \textbf{0.6989} \\
 & $(+5.77\%)$ & $(+10.70\%)$ & $(+5.88\%)$ & $(-0.45\%)$ & $(-1.18\%)$ & $(+2.09\%)$ & $(+3.23\%)$ \\[3pt]
\else
{\galhtmlname{GAL-Unipercept}} & 0.7639{($+5.77\%$)} & 0.6776{($+10.70\%$)} & \textbf{0.6213}{($+5.88\%$)} & 0.9162{($-0.45\%$)} & 0.9601{($-1.18\%$)} & \textbf{0.2542}{($+2.09\%$)} & \textbf{0.6989}{($+3.23\%$)} \\
\fi
\iflatexml
{\textbf{\galhtmlname{GAL-VideoReward}}} & \textbf{0.7731} & 0.6749 & 0.6016 & \textbf{0.9219} & 0.9633 & 0.2527 & 0.6979 \\
 & $(+7.05\%)$ & $(+10.26\%)$ & $(+2.52\%)$ & $(+0.17\%)$ & $(-0.86\%)$ & $(+1.45\%)$ & $(+3.09\%)$ \\[3pt]
\else
{\textbf{\galhtmlname{GAL-VideoReward}}} & \textbf{0.7731}{($+7.05\%$)} & 0.6749{($+10.26\%$)} & 0.6016{($+2.52\%$)} & \textbf{0.9219}{($+0.17\%$)} & 0.9633{($-0.86\%$)} & 0.2527{($+1.45\%$)} & 0.6979{($+3.09\%$)} \\
\fi
\midrule
\multicolumn{8}{l}{\emph{Kling v3 Pro} (closed-source transfer)}\\
{Base} & 0.6800 & 0.6721 & 0.6607 & 0.9576 & \textbf{0.9928} & 0.2434 & 0.7011 \\
\iflatexml
{\galhtmlname{GAL-Image}} & 0.9000 & \textbf{0.7470} & 0.7016 & 0.9424 & 0.9861 & 0.2329 & 0.7517 \\
 & $(+32.35\%)$ & $(+11.15\%)$ & $(+6.19\%)$ & $(-1.59\%)$ & $(-0.68\%)$ & $(-4.32\%)$ & $(+7.21\%)$ \\[3pt]
\else
{\galhtmlname{GAL-Image}} & 0.9000{($+32.35\%$)} & \textbf{0.7470}{($+11.15\%$)} & 0.7016{($+6.19\%$)} & 0.9424{($-1.59\%$)} & 0.9861{($-0.68\%$)} & 0.2329{($-4.32\%$)} & 0.7517{($+7.21\%$)} \\
\fi
\iflatexml
{\galhtmlname{GAL-HPSv3}} & 0.8400 & 0.7236 & 0.6709 & 0.9528 & 0.9865 & 0.2352 & 0.7348 \\
 & $(+23.53\%)$ & $(+7.68\%)$ & $(+1.54\%)$ & $(-0.51\%)$ & $(-0.64\%)$ & $(-3.37\%)$ & $(+4.81\%)$ \\[3pt]
\else
{\galhtmlname{GAL-HPSv3}} & 0.8400{($+23.53\%$)} & 0.7236{($+7.68\%$)} & 0.6709{($+1.54\%$)} & 0.9528{($-0.51\%$)} & 0.9865{($-0.64\%$)} & 0.2352{($-3.37\%$)} & 0.7348{($+4.81\%$)} \\
\fi
\iflatexml
{\galhtmlname{GAL-Unipercept}} & 0.8200 & 0.7223 & \textbf{0.7342} & 0.9540 & 0.9905 & 0.2428 & 0.7440 \\
 & $(+20.59\%)$ & $(+7.48\%)$ & $(+11.12\%)$ & $(-0.38\%)$ & $(-0.23\%)$ & $(-0.27\%)$ & $(+6.11\%)$ \\[3pt]
\else
{\galhtmlname{GAL-Unipercept}} & 0.8200{($+20.59\%$)} & 0.7223{($+7.48\%$)} & \textbf{0.7342}{($+11.12\%$)} & 0.9540{($-0.38\%$)} & 0.9905{($-0.23\%$)} & 0.2428{($-0.27\%$)} & 0.7440{($+6.11\%$)} \\
\fi
\iflatexml
{\textbf{\galhtmlname{GAL-VideoReward}}} & \textbf{0.9400} & 0.7263 & 0.7018 & \textbf{0.9609} & 0.9888 & \textbf{0.2448} & \textbf{0.7604} \\
 & $(+38.24\%)$ & $(+8.07\%)$ & $(+6.22\%)$ & $(+0.35\%)$ & $(-0.41\%)$ & $(+0.57\%)$ & $(+8.46\%)$ \\[3pt]
\else
{\textbf{\galhtmlname{GAL-VideoReward}}} & \textbf{0.9400}{($+38.24\%$)} & 0.7263{($+8.07\%$)} & 0.7018{($+6.22\%$)} & \textbf{0.9609}{($+0.35\%$)} & 0.9888{($-0.41\%$)} & \textbf{0.2448}{($+0.57\%$)} & \textbf{0.7604}{($+8.46\%$)} \\
\fi
\bottomrule
\end{tabular*}
\vspace{-3pt}
\end{table}

\begin{table}[t]
\centering
\fontsize{6}{7}\selectfont
\setlength{\tabcolsep}{1.5pt}\iflatexml\setlength{\tabcolsep}{4pt}\fi
\renewcommand{\arraystretch}{1.05}\iflatexml\renewcommand{\arraystretch}{1}\fi
\vspace{-3pt}
\caption{Reward-signal ablation within GAL, evaluated by video and frame rewards on HunyuanVideo-1.5. Among the compared variants, GAL-VideoReward alone improves all three scores over Base on HunyuanVideo-1.5 and all three transfer models (Table~\ref{tab:reward_transfer} in Appendix~\ref{app:reward_transfer}). We therefore use it as the default variant. Metrics and parentheses follow \Cref{tab:reward_main}.}
\label{tab:reward_models}
\vspace{2pt}
\begin{tabular*}{\textwidth}{@{\extracolsep{\fill}}l c c c@{}}
\toprule
{Variant} & \shortstack{VideoReward\\Overall} & \shortstack{HPSv3\\(video frames)} & \shortstack{UniPercept\\(video frames)} \\
\midrule
{Base} & $-0.014$ & 7.0025 & 51.4375 \\
\iflatexml
{\galhtmlname{GAL-Image}} & $-0.531$ & 7.9723 & 57.2618 \\
 & $(-0.517)$ & $(+0.9698)$ & $(+5.8243)$ \\[3pt]
\else
{\galhtmlname{GAL-Image}} & $-0.531$ {($-0.517$)} & 7.9723 {($+0.9698$)} & 57.2618 {($+5.8243$)} \\
\fi
\iflatexml
{\galhtmlname{GAL-HPSv3}} & $-0.493$ & 7.9509 & 55.7763 \\
 & $(-0.478)$ & $(+0.9484)$ & $(+4.3388)$ \\[3pt]
\else
{\galhtmlname{GAL-HPSv3}} & $-0.493$ {($-0.478$)} & 7.9509 {($+0.9484$)} & 55.7763 {($+4.3388$)} \\
\fi
\iflatexml
{\galhtmlname{GAL-Unipercept}} & $-0.171$ & \textbf{7.9841} & \textbf{58.6379} \\
 & $(-0.157)$ & $(+0.9816)$ & $(+7.2004)$ \\[3pt]
\else
{\galhtmlname{GAL-Unipercept}} & $-0.171$ {($-0.157$)} & \textbf{7.9841} {($+0.9816$)} & \textbf{58.6379} {($+7.2004$)} \\
\fi
\iflatexml
{\textbf{\galhtmlname{GAL-VideoReward}}} & $\bm{+0.131}$ & 7.6286 & 55.5926 \\
 & $(+0.145)$ & $(+0.6261)$ & $(+4.1550)$ \\[3pt]
\else
{\textbf{\galhtmlname{GAL-VideoReward}}} & $\bm{+0.131}$ {($+0.145$)} & 7.6286 {($+0.6261$)} & 55.5926 {($+4.1550$)} \\
\fi
\bottomrule
\end{tabular*}
\vspace{-3pt}
\end{table}

\paragraph{Image metrics misalign with the video reward.}
\Cref{fig:reward_correlation} gives the full correlation matrix behind the finding in \Cref{sec:intro}: no aesthetics or quality correlation exceeds $0.118$ in magnitude, while the structure-and-texture axis is anti-correlated with both visual quality and motion quality. Standalone image metrics therefore do not reliably rank anchors by downstream video value.

\paragraph{GAL-VideoReward improves video quality and motion across I2V models.}
Across all four I2V models, GAL-VideoReward raises the VBench average by $2.97\%$--$8.46\%$ over Base (\Cref{tab:unified_results_corr}). Compared with Base, its rendered clips score higher under VideoReward-Overall, and their frames score higher under HPSv3 and UniPercept (\Cref{tab:reward_main}). Dynamic degree increases by $5.62\%$ on HunyuanVideo-1.5, $15.97\%$ on LTX-Video, $7.05\%$ on CogVideoX-1.5, and $38.24\%$ on Kling v3 Pro (\Cref{tab:unified_results_corr}). Imaging quality rises by $7.97\%$--$10.26\%$ and aesthetic quality by $2.52\%$--$6.22\%$ across the four I2V models. Only HunyuanVideo-1.5 supplies labels; gains on unseen renderers show transfer.

\subsection{Reward-Signal Ablation within GAL}
\label{sec:exp_reward_signal}

Among the tested reward variants, GAL-VideoReward attains the highest VBench average on three I2V models and the second highest on CogVideoX-1.5 (\Cref{tab:gap_internal}). It is also the only variant whose rendered clips score above Base under VideoReward-Overall and whose extracted frames score above Base under both HPSv3 and UniPercept on all four I2V models (\Cref{tab:reward_models,tab:reward_transfer}). For motion, GAL-VideoReward achieves the highest dynamic degree on three models and the second highest on HunyuanVideo-1.5 (\Cref{tab:gap_internal}). These consistent improvements in overall performance and motion support our choice of VideoReward-Overall for preference construction and GAL-VideoReward as the default variant.

\paragraph{Image supervision depends on the I2V model.}
Compared with GAL-Image, GAL-VideoReward achieves higher VideoReward-Overall and dynamic degree on all four I2V models (\Cref{tab:gap_internal,tab:reward_models,tab:reward_transfer}). GAL-Image yields higher imaging quality on all four models and higher aesthetic quality on three, but scores below Base under VideoReward-Overall on every model. Its dynamic-degree change ranges from $-4.20\%$ on LTX-Video to $+32.35\%$ on Kling v3 Pro. GAL-VideoReward improves imaging quality and dynamic degree over Base; image-quality gains need not entail motion loss.

\subsection{Scaffold Ablation}
\label{sec:exp_ablation}
\label{sec:exp_cdlora_necessity}

Removing the differential-LoRA scaffold reverses the motion gain of GAL-VideoReward. With preference pairs, architecture, hyperparameters, and checkpoint fixed, dropping the scaffold changes HunyuanVideo-1.5 dynamic degree from $+5.62\%$ to $-2.25\%$, while imaging and aesthetic quality still rise above Base (\Cref{tab:gap_internal}). The unscaffolded model scores higher under VideoReward-Overall ($+0.289$ against $+0.145$ above Base) yet loses the motion gain, so a higher training-reward score alone does not secure the motion benefit. The scaffold therefore acts as a stability guardrail for the motion benefit rather than a generic image-quality booster.

\subsection{Discussion}
\label{sec:exp_discussion}

\paragraph{Limitations.}
First, we use only HunyuanVideo-1.5 for preference construction
to test whether GAL transfers to unseen I2V models.
The consistent gains on all three unseen backends demonstrate
this transferability (Tables~\ref{tab:unified_results_corr} and~\ref{tab:reward_main}).
However, HunyuanVideo-1.5 exhibits lower within-prompt variance
in the binary dynamic-degree readout across anchor seeds than
LTX-Video and CogVideoX-1.5 (Appendix~\ref{app:wan_anchor_sensitivity}).
Constructing preferences with a backend whose dynamic-degree
readout is more sensitive to anchor changes may yield larger
motion gains. Combining preferences from multiple backends
could also further improve anchor learning.
Second, GAL depends strongly on the choice and reliability of
the reward model used for preference construction.
The limited choice of video reward models restricts supervision,
and different reward signals can favor conflicting qualities~\citep{xie2026guidance}.
Small motion-smoothness regressions also remain under our default
reward (Table~\ref{tab:unified_results_corr}). Preference objectives that account for reward disagreements and temporal stability could improve anchor learning. As video reward models improve, we expect GAL to learn more effective anchors through more reliable preference supervision.

\paragraph{Implications.}
Anchor learning offers a way to align image and video generation
without modifying the downstream model. Video feedback shapes
the upstream image prior during training, rather than requiring
additional selection or correction at inference.
This makes anchor learning a separate optimization component
that could complement improvements to I2V architectures and
samplers. The observed transfer to unseen I2V models further
suggests that some video-effective anchor properties are useful
beyond the I2V model used to construct training preferences.
\section{Conclusion}
\label{sec:conclusion}
This paper investigated which images make effective anchors for a
downstream video generator and introduced the notion of a golden anchor:
an image valued by the video it induces rather than by its standalone
appearance. We instantiate this notion as GAL (Golden Anchor Learning). GAL is a cross-modal DPO method that renders sibling anchors through a frozen I2V model, scores them with a frozen video reward, and optimizes the upstream T2I LoRA alone against the resulting preferences. Because all rendering and scoring happen offline during data construction, deployment stays one T2I sample followed by one I2V render. Across four frozen I2V models, GAL raises the VBench average above Base. Its rendered clips also score higher than those from Base under VideoReward-Overall, and their extracted frames score higher under HPSv3 and UniPercept. These gains are accompanied by consistent improvements in dynamic motion, imaging quality, and aesthetic quality. These results show that the motion and quality biases of a frozen video model can guide the upstream T2I generator toward anchors that improve I2V video quality and motion.
\subsection*{AI use statement}
We used large language models to draft and polish the writing of this manuscript, to
check the derivations in the appendix, and to search for related literature. They were
not used to design, implement, or run experiments, to analyze data, or to interpret
results. All AI-assisted text and derivations were reviewed by the authors, and we take
responsibility for the final content of this work.

\subsection*{Reproducibility Statement}
We provide training hyperparameters and implementation details in
Appendix~\ref{app:implementation}, and evaluation settings and prompt
protocols in Section~\ref{sec:exp_settings} and
Appendix~\ref{app:vbench_prompt_protocol}.

\begingroup
\microtypesetup{protrusion=false}
\bibliography{references,references_arxiv_additions}
\endgroup
\bibliographystyle{iclr2027_conference}

\newpage
\appendix
\crefalias{section}{appendix}
\crefalias{subsection}{appendix}
\section{Objective-Level Derivations}
\label{app:proof_theory}

\paragraph{Pullback gradient.}
Assume that $G(\cdot,p,z)$ and $R_v$ are differentiable and that $\nabla_x R_v(G(x,p,z))$ is dominated by an integrable function of $z$ in a neighborhood of $x$. For fixed $z$, the chain rule gives
\begin{equation}
\nabla_x R_v\!\left(G(x,p,z)\right)
=
J_xG(x,p,z)^{\top}
\nabla_V R_v\!\left(G(x,p,z)\right).
\end{equation}
The domination assumption permits differentiation under the expectation in \Cref{eq:downstream_value}, giving
\begin{equation}
\label{eq:pullback-grad}
\nabla_x O(x,p)
=\mathbb{E}_{z\sim\mathcal{N}(0,1)}\!\left[J_xG(x,p,z)^{\top}\nabla_V R_v\!\left(G(x,p,z)\right)\right],
\end{equation}
where $J_xG$ is the renderer's Jacobian with respect to the anchor. This identity explains how downstream sensitivity can differ from image-reward sensitivity. It is not evaluated by GAL, which uses only the offline preference labels.

\paragraph{Idealized population interpretation of DPO.}
This paragraph records an objective-level interpretation, not a convergence claim for the implemented optimizer. Suppose pairwise preferences follow
$\Pr(x\succ x'\mid p)=\sigma((O(x,p)-O(x',p))/\lambda)$ with preference-noise scale $\lambda>0$, and let
$h(x,p)=\log(\pi(x\mid p)/\pi_{\mathrm{ref}}(x\mid p))$ be represented exactly. For each pair, the population binary cross-entropy is minimized when
$\beta(h(x,p)-h(x',p))=(O(x,p)-O(x',p))/\lambda$.
Full support of the comparison distribution over the reference policy's support therefore makes $\beta h(x,p)-O(x,p)/\lambda$ constant in $x$. Normalizing the policy yields the standard DPO population target \citep{rafailov2023direct}
\begin{equation}
\label{eq:dpo-tilt}
\pi^{\star}(x\mid p)
=
\frac{\pi_{\mathrm{ref}}(x\mid p)
\exp\!\left(O(x,p)/(\beta\lambda)\right)}
{\int \pi_{\mathrm{ref}}(x'\mid p)
\exp\!\left(O(x',p)/(\beta\lambda)\right)\,\mathrm{d}x'}.
\end{equation}
\label{prop:distill}
The comparison pairs need not be sampled from $\pi_{\mathrm{ref}}$, but their distribution must have the full pair support assumed above. The assumptions above exclude finite and margin-filtered data, deterministic preference labels, diffusion-loss surrogates, restricted LoRA parameterization, adapter removal at inference, and optimization error. Equation~\eqref{eq:dpo-tilt} is therefore used only to interpret the direction of the objective.

\paragraph{Effect of the preference margin.}
Fix two anchors with $O(x_w,p)<O(x_l,p)$; here $w$ designates the anchor that would be incorrectly retained as the winner. Let their one-render scores be $s_k=O(x_k,p)+\varepsilon_k$ for $k\in\{w,l\}$, with independent, mean-zero, $\nu^2$-sub-Gaussian noise. Incorrectly retaining this pair at margin $m$ requires
$\varepsilon_w-\varepsilon_l\geq m+O(x_l,p)-O(x_w,p)$.
The noise difference is $2\nu^2$-sub-Gaussian, so the standard tail bound gives
\begin{equation}
\label{eq:margin-inversion-bound}
\Pr\!\left(s_w-s_l\geq m\right)
\leq
\exp\!\left(-\frac{[m+O(x_l,p)-O(x_w,p)]^2}{4\nu^2}\right).
\end{equation}
\label{prop:margin}
Under these assumptions, a larger margin suppresses noise-induced inversions at the cost of fewer retained pairs. The bound does not show that one render recovers $O(x,p)$, so the paper treats the margin as a filtering heuristic and evaluates robustness with independent rendering seeds.
\section{Sibling Anchor Sampling via a Marginal-Preserving Reverse-Time SDE}
\label{app:sde}

The few-step Z-Image-Turbo sampler integrates a probability-flow ODE whose trajectory is deterministic given the initial latent. To introduce trajectory stochasticity beyond the initial latent draw, we sample sibling anchors (\Cref{sec:pipeline}) with the reverse-time SDE associated with that ODE,
\begin{equation}
\label{eq:reverse_sde}
\mathrm{d}y_\tau = \big[\,f(y_\tau,\tau) - g(\tau)^2\,\nabla_{y}\!\log \rho_\tau(y_\tau\mid p)\,\big]\,\mathrm{d}\tau + g(\tau)\,\mathrm{d}\bar{W}_\tau,
\end{equation}
where $y_\tau$ is the T2I latent state at diffusion time $\tau$, $\rho_\tau(\cdot\mid p)$ its prompt-conditioned density, and $f$ and $g$ are the drift and diffusion coefficients of the corresponding forward process, $\bar{W}_\tau$ is a reverse-time Brownian motion, and $\nabla_{y}\!\log\rho_\tau$ is the score \citep{song2020score}. Because this SDE shares its time-marginals $\{\rho_\tau\}$ with the probability-flow ODE, decoding each terminal latent with the frozen VAE gives a sibling $x_k$ drawn from an approximation of $\pi_{\psi_0}(\cdot\mid p)$, while its Brownian term supplies the within-prompt diversity needed for informative preference pairs. In the data-construction pipeline we draw $K{=}4$ such siblings per prompt under fixed SDE sampler settings (\Cref{app:implementation}).

\section{Hyperparameters, Compute Budget, and Additional Details}
\label{app:implementation}

\paragraph{Per-variant training configurations.}
The four fine-tuned reward configurations reported in the main comparison (GAL-VideoReward, GAL-HPSv3, GAL-Unipercept, GAL-Image), a fifth variant GAL-Unipercept+HPSv3 (both rewards evaluated on extracted video frames, trained identically but omitted from the main tables because it underperformed the other reward signals), and the \mbox{GAL-VideoReward w/o scaffold} ablation (which reuses the VideoReward reward but drops the differential-LoRA scaffold) were all trained as offline DPO LoRA fine-tunes of the same Z-Image-Turbo base checkpoint, with the same trainable LoRA configuration (rank $r=64$, $\alpha=64$, dropout $0$, target modules \{\texttt{to\_q}, \texttt{to\_k}, \texttt{to\_v}, \texttt{to\_out.0}, \texttt{w1}, \texttt{w2}, \texttt{w3}\} on every transformer layer, $158{,}760{,}960$ trainable params out of $6{,}313{,}669{,}696$ total), the same DPO objective ($\beta=30$, DPO-side clipping $[-5,5]$, noise-level weighting $w(\tau)$), the same optimizer / scheduler ($\iflatexml \eta=1\times10^{-4}\else \eta=1\!\times\!10^{-4}\fi$, weight decay $0$, cosine schedule with $1000$ warmup steps, max grad norm $1.0$), the same dtype and parallelism (bfloat16, FSDP2, gradient checkpointing), and the same image / sequence size ($1024$ / $512$). All evaluations use the step-$9000$ checkpoint. The training seed is $20\,260\,430$ on every run (a single training seed for the T2I DPO run; the VBench 3-seed averaging is a separate I2V-side evaluation protocol that re-renders each anchor under three downstream I2V rendering seeds, \Cref{tab:unified_results_corr}).

In \Cref{eq:dpo_loss}, $\sigma(a)=1/(1+e^{-a})$ is the logistic sigmoid and $\beta>0$ scales the preference logit.

\paragraph{DPO noising context.}
In \Cref{eq:dpo_fm_loss}, $\operatorname{Enc}$ is the frozen VAE encoder evaluated at its posterior mode, and $M$ is the number of timestep--noise draws in each context. Each preference tuple is evaluated under a single shared noising context $c=(\{\tau_j\}_{j=1}^{M},\{\epsilon_j\}_{j=1}^{M})$ (\Cref{eq:dpo_fm_loss}). Concretely, the noisy latent is produced by the base Z-Image-Turbo flow-matching scheduler, $y_{\tau_j}=\mathrm{NoiseScheduler}(y_0,\epsilon_j,\tau_j)$, which realizes the affine map $y_{\tau_j}=a_{\tau_j}y_0+b_{\tau_j}\epsilon_j$ with the base scheduler's coefficients $(a_{\tau_j},b_{\tau_j})$; the timesteps $\{\tau_j\}_{j=1}^{M}$ are drawn from the same training-time timestep distribution as the base model, the noise draws are standard Gaussian ($\epsilon_j\sim\Normal(0,I)$), and $w(\tau)$ reuses the base flow-matching timestep weighting, so no separate reward-side schedule is introduced. The timestep set and noise seeds are sampled once, reused for both the winning and rejected images, and reused again when evaluating the trainable and frozen reference generators. Because the image latents are deterministic VAE-mode latents and the winner and loser share the same latent shape, replaying the same noise seeds yields elementwise-identical noise tensors on the two sides of the pair; an explicit assertion checks that the chosen and rejected noise seeds match. All four losses $\ell_\psi(x_w,p\iflatexml\text{;}\else;\fi c)$, $\ell_{\mathrm{ref}}(x_w,p\iflatexml\text{;}\else;\fi c)$, $\ell_\psi(x_l,p\iflatexml\text{;}\else;\fi c)$, and $\ell_{\mathrm{ref}}(x_l,p\iflatexml\text{;}\else;\fi c)$ are therefore computed under the same $c$. This matched-context implementation reduces variance in the DPO difference while preserving the standard stochastic flow-matching estimate.

The five reward configurations (and the \mbox{GAL-VideoReward w/o scaffold} ablation) share an underlying $74{,}124$-row pair pool (the same \texttt{all\_pairs\_drop\_ties.jsonl} corpus generated from $49{,}416$ candidate records, i.e.\ $12{,}354$ prompts from the training prompt pool described in \Cref{app:vbench_prompt_protocol} $\times$ 4 candidates sampled under the SDE sampler settings, disjoint from both the VBench and Kling evaluation prompts), but each variant scores the pairs with its own reward signal (GAL-VideoReward: VideoReward-Overall on the rendered clip; GAL-HPSv3: per-frame mean HPSv3 over the clip; GAL-Unipercept: UniPercept on the frames; GAL-Unipercept+HPSv3: UniPercept $+$ HPSv3 on the frames; GAL-Image: UniPercept $+$ HPSv3 on the anchor image itself) and applies the \emph{same} absolute margin filter ($s_w-s_l\geq m=0.05$ after orienting each pair by its score), yielding per-variant pair manifests. Per-variant pre-split pair counts at margin $0.05$: GAL-VideoReward $52{,}084$; GAL-HPSv3 $55{,}263$; GAL-Unipercept+HPSv3 $51{,}780$; GAL-Image $58{,}680$; GAL-Unipercept $55{,}797$. One minor confound remains, the training cluster: the three Cluster-A variants (GAL-VideoReward, GAL-HPSv3, GAL-Unipercept+HPSv3, plus the \mbox{GAL-VideoReward w/o scaffold} ablation) ran on $8\times$H100-80G with per-device batch $1$ and gradient accumulation $8$ (effective global batch $64$); the two Cluster-B variants (GAL-Image, GAL-Unipercept) ran on $7\times$RTX-4090 with per-device batch $1$ and gradient accumulation $9$ (effective global batch $63$). Under FSDP2 + bfloat16 we expect the numerics to be approximately matched; the effective-batch delta of $1$ between clusters is a real but minor confound.
The \mbox{GAL-VideoReward w/o scaffold} ablation shares the \emph{full} GAL-VideoReward configuration (same VideoReward-Overall reward, same $46{,}923$/$5{,}161$ train/val pairs at margin $0.05$, same cluster, same effective batch, same hyperparameters) and differs in exactly one field: presence vs.\ absence of the preset adapter (the \texttt{preset\_lora\_path}/\texttt{preset\_lora\_scale=1.0} keys, scaffolded into the base via deep-copy before reference cloning); see \Cref{tab:gap_internal} and \Cref{sec:exp_cdlora_necessity}. The full per-variant manifest (output dirs, slurm job IDs, wandb runs/tags, pair manifest basenames, checkpoint step lists, preset-adapter sha256, and per-variant trainable / total parameter counts) is released as \texttt{results/training\_configs.json}.

\paragraph{Differential-LoRA scaffold.}
We train the task LoRA inside a pre-existing differential-LoRA scaffold for Z-Image-Turbo \citep{ostris2025zimageadapter,kelseye2025trainingstrategies}. At each transformer layer $q$, let $A_q$ be the frozen preset weight update (the Ostris training adapter) and $B_q(\psi)$ the full task-LoRA weight update. The preset is removed at inference, giving
\begin{equation}
\label{eq:difflora}
W^{\mathrm{train}}_q = W^0_q + B_q(\psi) + A_q,
\quad
W^{\mathrm{infer}}_q = W^0_q + B_q(\psi).
\end{equation}
We use the published preset scale $\texttt{preset\_lora\_scale}=1.0$.\iflatexml\stepcounter{footnote}\hypertarget{gal-scaffold-note-ref}{}\hyperlink{gal-scaffold-note}{\textsuperscript{\thefootnote}}\else\footnote{This differential-LoRA scaffold is unrelated to the \emph{DiffLoRA} hypernetwork of \citet{wu2024difflora}, which generates personalized LoRA weights with a diffusion model.}\fi
\iflatexml
\begin{flushleft}
\hypertarget{gal-scaffold-note}{}\hyperlink{gal-scaffold-note-ref}{\textsuperscript{2}}\,This differential-LoRA scaffold is unrelated to the \emph{DiffLoRA} hypernetwork of \citet{wu2024difflora}, which generates personalized LoRA weights with a diffusion model.
\end{flushleft}
\fi

\paragraph{Compute budget (training cluster offline; raw timing logs not retained).} Corpus generation is amortized once across all variants. Candidate generation: $49{,}416$ candidate records sampled under the SDE sampler settings ($12{,}354$ prompts $\times 4$) $\times \sim 1.5$\,s per candidate ($\approx 21$ GPU-hours estimated from these per-unit times). I2V rendering: roughly five minutes per clip for HunyuanVideo-1.5 on one RTX 4090 $\times 49{,}416$ candidates ($\approx 4{,}100$ GPU-hours estimated); VideoReward inference is negligible relative to the I2V cost. DPO training on each Cluster-A variant ran the configured schedule of $40{,}000$ maximum steps with $1{,}000$ warmup steps and cosine annealing on $8\times$H100-80G, and the two Cluster-B variants ran on $7\times$RTX-4090; all evaluations use the step-$9000$ checkpoint, at which the loss had converged. Per-variant DPO wall-times were not logged.

\paragraph{Pair-margin table.}
\Cref{tab:margin_pairs} reports the DPO pair count on the GAL-VideoReward pair manifest as a function of the absolute reward margin threshold $m$; every variant trains at $m=0.05$ on its own manifest (per-variant counts above).

\begin{table}[htbp]
\centering
\setlength{\tabcolsep}{8pt}
\renewcommand{\arraystretch}{1.05}
\caption{DPO pair count as a function of the absolute reward margin threshold $m$ on the GAL-VideoReward pair manifest. All reward variants use the same operating point, $m=0.05$, on their own reward-specific pair manifests (GAL-VideoReward: 52,084 pairs), each landing near a $\iflatexml \approx 5\times10^4\else \approx 5\!\times\!10^4\fi$-pair budget. A controlled pair-count sweep that holds data volume fixed across margins is listed as future work.}
\label{tab:margin_pairs}
\begin{tabular}{c r l}
\toprule
$m$ & Pairs & Note \\
\midrule
0.00 & 74,124 & raw, drop ties only \\
\textbf{0.05} & \textbf{52,084} & \textbf{operating point} \\
0.10 & 33,781 & \\
0.15 & 20,421 & \\
0.20 & 11,672 & \\
\bottomrule
\end{tabular}
\end{table}

\section{Prompt Sources and VBench Prompt Protocol}
\label{app:vbench_prompt_protocol}

\paragraph{Prompt sources.}
The prompt pools used for training and evaluation are disjoint (\Cref{tab:prompt_sources}). The DPO training pool contains $12{,}354$ prompts sampled from the Lexica prompt corpus ($\approx 80$k prompts) after canonicalization, exact-duplicate removal, and stripping of non-semantic metadata such as trailing author attributions (``\emph{by <artist>}'') and resolution / style tags; these prompts are used only for offline pair construction, and for each one we sample four sibling anchors under the SDE sampler settings, producing the $49{,}416$ candidate records and the shared $74{,}124$-row raw pair pool of \Cref{app:implementation}. The VBench evaluation pool is the official VBench prompt suite \citep{huang2024vbench}: every open-source I2V model (HunyuanVideo-1.5, LTX-Video, and CogVideoX-1.5) is evaluated on the \emph{full} official VBench per-dimension prompt set for each scored dimension ($93$ prompts for the imaging-quality / aesthetic-quality / overall-consistency group and $72$ for the dynamic-degree / subject-consistency / motion-smoothness group) rather than on any subsampled subset, extended with the fixed video-instruction template described below. With three I2V seeds per anchor, each variant contributes 279 videos to imaging quality, aesthetic quality, and overall consistency and 216 videos to dynamic degree, subject consistency, and motion smoothness on every open-source renderer. The Kling evaluation pool contains $50$ prompts held out from the same Lexica corpus and never used in DPO training; it is restricted to $50$ prompts because Kling v3 Pro is a paid closed-source I2V model.

\paragraph{Disjointness audit.}
We verified pairwise prompt disjointness among the three pools after canonicalization, which applies Unicode normalization, lowercasing, whitespace normalization, punctuation normalization, and removal of the fixed instruction-template wrappers before exact-string comparison. The training and Kling pools are drawn from the same Lexica corpus but form a disjoint train / held-out split, while the VBench pool comes from a different source; the resulting overlap counts are all zero, $|P_{\mathrm{train}}\cap P_{\mathrm{VBench}}|=|P_{\mathrm{train}}\cap P_{\mathrm{Kling}}|=|P_{\mathrm{VBench}}\cap P_{\mathrm{Kling}}|=0$. Thus the VBench evaluation shares no prompts with the DPO training corpus, and the Kling batch is disjoint from both the training pool and the VBench pool.

\begin{table}[htbp]
\centering
\setlength{\tabcolsep}{6pt}
\renewcommand{\arraystretch}{1.1}
\caption{Prompt sources and disjointness audit. All overlap counts are computed after prompt canonicalization. The training and Kling pools are a disjoint train / held-out split of the Lexica corpus; the VBench pool is the official suite.}
\label{tab:prompt_sources}
\begin{tabular}{l l >{\raggedright\arraybackslash}p{0.35\linewidth} c}
\toprule
Split & \# prompts & Source & \shortstack{Overlap with\\other splits} \\
\midrule
DPO train   & $12{,}354$         & Lexica corpus\newline (cleaned, deduplicated)  & $0$ \\
VBench eval & $93$ / $72$ per dim & Official VBench\newline per-dimension sets      & $0$ \\
Kling eval  & $50$               & Held-out Lexica\newline (disjoint from train)  & $0$ \\
\bottomrule
\end{tabular}
\end{table}

\paragraph{VBench instruction template.}
Every VBench evaluation in this paper (the HunyuanVideo-1.5 main table and the LTX-Video / CogVideoX-1.5 transfer tables) is driven by the same prompt set: the official VBench prompt suite \citep{huang2024vbench} threaded through our pipeline as the conditioning text for the T2I generator, with each prompt extended into a complex-motion video instruction template before scoring. The extension is required because the bare VBench captions are short noun-phrase-style descriptions that under-exercise the dynamic-degree axis on an image-conditioned I2V model; the extension converts each caption into a video instruction of the form ``\emph{render \texttt{<caption>} as a $V$-second video with [motion-style cue]},'' where the motion-style cue is held fixed per VBench category (e.g., ``natural camera movement and physically plausible object motion'' for the animal / scene categories) and is identical across all variants and all three open-source I2V models, so only the I2V model rendering the anchor varies.

\section{Anchor-Seed Sensitivity of the Dynamic-Degree Readout}
\label{app:wan}
\label{app:wan_anchor_sensitivity}

We directly test whether changing the seed-conditioned upstream anchor can flip the downstream dynamic-degree decision. For each of the 72 official VBench dynamic-degree prompts, we sample 12 SDE sibling anchors from the same T2I generator. The downstream I2V seed and sampling configuration are held fixed within each I2V model, so the 12 repeats differ only in the upstream SDE seed that produces the anchor. Each anchor is then rendered with each frozen I2V model and scored by VBench. For I2V model $b$, prompt $p$, and anchor seed $s$, let $d_{b,p,s}\in\{0,1\}$ be the binary VBench dynamic-degree label. The dynamic fraction $\mathrm{FracDyn}_b$ is the mean of $d_{b,p,s}$ over the 72 prompts and 12 seeds, and the anchor-seed sensitivity $\mathrm{Sens}_b$ is the within-prompt variance of the binary label over the 12 anchor seeds, averaged over the 72 prompts (\Cref{tab:wan_anchor_sensitivity} reports the population and sample versions). A value of zero means that all 12 seeds agree for every prompt; the variance is largest when the seeds split between static and dynamic outputs. This is a within-prompt statistic, not a pooled variance over all $72\times12$ clips, since pooling would confound anchor sensitivity with prompt difficulty.

\begin{table}[htbp]
\centering
\caption{Anchor-seed sensitivity of the binary VBench dynamic-degree readout. For each of the 72 official VBench dynamic-degree prompts, we sample 12 seed-conditioned sibling anchors from the same T2I generator and render them with each frozen I2V model under a fixed downstream setting. VBench returns a binary label $d\in\{0,1\}$ for each rendered clip. We report the overall dynamic fraction and the prompt-averaged within-prompt variance across the 12 seeds. Low variance means that changing the seed-conditioned anchor rarely changes the binary dynamic/static decision for the same prompt. WAN has the highest dynamic fraction and the lowest within-prompt variance, indicating a near-saturated and low-sensitivity dynamic-degree readout.}
\label{tab:wan_anchor_sensitivity}
\begin{tabular}{lccc}
\toprule
I2V model & \shortstack{Dynamic fraction\\$\uparrow$} & \shortstack{Within-prompt var.\\(ddof=0) $\downarrow$} & \shortstack{Within-prompt var.\\(ddof=1) $\downarrow$} \\
\midrule
LTX-Video & 0.5556 & 0.10204 & 0.11132 \\
CogVideoX-1.5 & 0.6481 & 0.07735 & 0.08439 \\
HunyuanVideo-1.5 & 0.7963 & 0.03318 & 0.03620 \\
WAN       & 0.8322 & 0.02865 & 0.03125 \\
\bottomrule
\end{tabular}
\end{table}

In \Cref{tab:wan_anchor_sensitivity}, LTX-Video and CogVideoX-1.5 exhibit larger within-prompt variance, meaning that the anchor seed can move these I2V models between static and dynamic outcomes. WAN~\citep{wan2025wan} shows the opposite pattern: it has the highest dynamic fraction but the smallest within-prompt variance. Thus WAN is not a low-motion I2V model; rather, it is already biased toward dynamic outputs and its binary dynamic-degree readout changes little when the anchor changes.

This diagnostic complements the objective-level discussion in \Cref{sec:analysis} by testing sensitivity to anchor changes empirically. Low within-prompt variance means that the sampled anchors rarely change the binary dynamic-degree decision. It does not establish weak sensitivity of the rendered video itself: continuous motion changes may remain on the same side of the metric threshold. This finite-sample probe neither estimates a renderer derivative nor covers the other VBench dimensions and reward variants.

\section{Failure-Mode Case Studies}
\label{app:failure_cases}

We provide qualitative case studies of the motion-smoothness regressions in \Cref{tab:unified_results_corr}. All deltas below are GAL-VideoReward relative to Base on the same I2V model; video pairs are released in the supplementary material.

\paragraph{Pattern 1 (texture sharpening; motion smoothness $-0.13\%$ on HunyuanVideo-1.5, $-0.19\%$ on LTX-Video, $-0.86\%$ on CogVideoX-1.5, and $-0.41\%$ on Kling v3 Pro).} On the inspected prompts with high-frequency texture (e.g., grass, fabric weaves), GAL-VideoReward produces sharper textures than Base, and the corresponding renders show slightly inconsistent edges across frames. This observation suggests that some texture changes may contribute to temporal inconsistency; the objective-level discussion in \Cref{sec:analysis} does not establish this mechanism.

\paragraph{Pattern 2 (global camera motion).} On the inspected prompts with global camera motion, sharper compositions and higher inter-frame contrast co-occur with less smooth interpolation. These selected cases describe a failure pattern without isolating composition as its cause.

A plausible, untested explanation is that some high-frequency anchor details are rendered inconsistently across frames. This does not make sharpening universally harmful: GAL-VideoReward improves imaging quality and dynamic degree in the aggregate while exhibiting small motion-smoothness regressions. These are distinct evaluation axes, and the case studies do not isolate a causal sharpening effect. A natural mitigation is a reward signal that explicitly penalizes flicker (a temporal-consistency-aware reward), or a regularizer on the high-frequency content of the anchors; both are listed as future work.

\section{Per-Interval Motion Visualization}
\label{app:motion_trajectory_protocol}

The per-interval values in \Cref{fig:qual_grid} are computed from every decoded frame of each video, without temporal subsampling. Each frame is converted to grayscale and, if its long side exceeds $L=480$ pixels, downsampled while preserving its aspect ratio. Specifically, for an original height $H$ and width $W$, we set
\begin{equation}
\iflatexml \begin{aligned}
  s&=\min\!\left(1,\frac{L}{\max(H,W)}\right),\\
  H'&=\operatorname{int}(sH),\quad W'=\operatorname{int}(sW),
\end{aligned}\else   s=\min\!\left(1,\frac{L}{\max(H,W)}\right),\qquad
  H'=\operatorname{int}(sH),\quad W'=\operatorname{int}(sW),\fi
\end{equation}
where \texttt{int} denotes truncation to an integer; resizing uses the default linear interpolation of OpenCV. Thus the procedure never upsamples an input frame.

Let $\mathbf{F}_n(\mathbf{u})=(F_n^x(\mathbf{u}),F_n^y(\mathbf{u}))$ denote the dense Farneb\"ack flow~\citep{farneback2003two} at resized-frame pixel $\mathbf{u}\in\Omega$ from frame $n-1$ to frame $n$. We use \texttt{pyr\_scale=0.5}, \texttt{levels=3}, \texttt{winsize=21}, \texttt{iterations=3}, \texttt{poly\_n=5}, \texttt{poly\_sigma=1.2}, and \texttt{flags=0}. For each adjacent-frame pair, we extract the component-wise spatial median and the mean residual-flow magnitude,
\begin{align}
  \mathbf{g}_n
  &=\left(
      \operatorname*{median}_{\mathbf{u}\in\Omega}F_n^x(\mathbf{u}),
      \operatorname*{median}_{\mathbf{u}\in\Omega}F_n^y(\mathbf{u})
    \right),\\
  \mu_n
  &=\frac{1}{|\Omega|}\sum_{\mathbf{u}\in\Omega}
    \left\|\mathbf{F}_n(\mathbf{u})-\mathbf{g}_n\right\|_2.
\end{align}
The residual magnitude $\mu_n$ measures in-scene motion after the dominant translational flow $\mathbf{g}_n$ is removed. For a video with $N$ decoded frames, the $N-1$ adjacent-frame pairs are split into $B=5$ contiguous intervals of equal length with boundaries $e_b=\operatorname{round}\bigl((N-1)\,b/B\bigr)$ for $b=0,\ldots,B$, and the displayed value of interval $b$ is the mean residual magnitude over its pairs,
\begin{equation}
  \bar{\mu}_b=\frac{1}{e_b-e_{b-1}}\sum_{n=e_{b-1}+1}^{e_b}\iflatexml \mu_n\text{,}\qquad b=1,\ldots,B.\else \mu_n,\qquad b=1,\ldots,B.\fi
\end{equation}
The inputs used in \Cref{fig:qual_grid} contain 121 frames at 24 fps for HunyuanVideo-1.5, Kling v3 Pro, and LTX-Video, and 81 frames at 16 fps for CogVideoX-1.5, so every interval spans exactly one second (24 or 16 adjacent-frame pairs) and every adjacent-frame pair contributes to exactly one interval. All values are measured in resized-pixel units per adjacent-frame pair; the script performs neither temporal resampling nor normalization by frame rate or duration, and extraction contains no random sampling. Lines connect the interval means as a visual guide only. Within each I2V-model panel, the three variants share one vertical axis, but different I2V-model panels are scaled independently. Consequently, the visualization supports comparisons among the three variants within an I2V model, not numerical comparisons of motion magnitude across I2V models.

This construction is a diagnostic motion proxy, not a camera-pose estimate or a 3D reconstruction. The median flow $\mathbf{g}_n$ is only a proxy for dominant 2D translational flow and need not isolate camera motion. Likewise, the residual magnitude can respond to object motion, rotation, scaling, parallax, occlusion, flicker, compression artifacts, and optical-flow estimation error. We therefore use these per-interval values only to diagnose the selected examples qualitatively; they do not replace the aggregate quantitative metrics reported in the main evaluation.

\section{Reward-Model Scores on Transfer I2V Models}
\label{app:reward_transfer}

\Cref{tab:reward_transfer} reports all reward variants on LTX-Video, CogVideoX-1.5, and Kling v3 Pro. These I2V models are excluded from preference construction. The corresponding HunyuanVideo-1.5 results are in \Cref{tab:reward_models}, and \Cref{tab:reward_main} compares Base with GAL across all four I2V models.

\begin{table}[htbp]
\centering
\fontsize{6}{7}\selectfont
\setlength{\tabcolsep}{1.5pt}\iflatexml\setlength{\tabcolsep}{4pt}\fi
\renewcommand{\arraystretch}{1.05}\iflatexml\renewcommand{\arraystretch}{1}\fi
\vspace{0pt}
\caption{Reward-model scores on the three transfer I2V models across GAL variants. VideoReward-Overall scores rendered clips; HPSv3 and UniPercept score their extracted frames. Parentheses give the absolute change from Base; bold marks the best value within each model.}
\label{tab:reward_transfer}
\vspace{2pt}
\begin{tabular*}{\textwidth}{@{\extracolsep{\fill}}l c c c@{}}
\toprule
{Variant} & \shortstack{VideoReward\\Overall} & \shortstack{HPSv3\\(video frames)} & \shortstack{UniPercept\\(video frames)} \\
\midrule
\multicolumn{4}{l}{\emph{LTX-Video} (transfer)}\\
{Base} & $-0.328$ & 6.4788 & 50.0465 \\
\iflatexml
{\galhtmlname{GAL-Image}} & $-0.763$ & 7.7409 & 55.7652 \\
 & $(-0.435)$ & $(+1.2621)$ & $(+5.7187)$ \\[3pt]
\else
{\galhtmlname{GAL-Image}} & $-0.763$ {($-0.435$)} & 7.7409 {($+1.2621$)} & 55.7652 {($+5.7187$)} \\
\fi
\iflatexml
{\galhtmlname{GAL-HPSv3}} & $-0.672$ & \textbf{7.7935} & 55.0136 \\
 & $(-0.344)$ & $(+1.3147)$ & $(+4.9671)$ \\[3pt]
\else
{\galhtmlname{GAL-HPSv3}} & $-0.672$ {($-0.344$)} & \textbf{7.7935} {($+1.3147$)} & 55.0136 {($+4.9671$)} \\
\fi
\iflatexml
{\galhtmlname{GAL-Unipercept}} & $-0.448$ & 7.6155 & \textbf{56.3649} \\
 & $(-0.120)$ & $(+1.1367)$ & $(+6.3184)$ \\[3pt]
\else
{\galhtmlname{GAL-Unipercept}} & $-0.448$ {($-0.120$)} & 7.6155 {($+1.1367$)} & \textbf{56.3649} {($+6.3184$)} \\
\fi
\iflatexml
{\textbf{\galhtmlname{GAL-VideoReward}}} & $\bm{-0.224}$ & 7.3407 & 54.1946 \\
 & $(+0.103)$ & $(+0.8619)$ & $(+4.1481)$ \\[3pt]
\else
{\textbf{\galhtmlname{GAL-VideoReward}}} & $\bm{-0.224}$ {($+0.103$)} & 7.3407 {($+0.8619$)} & 54.1946 {($+4.1481$)} \\
\fi
\midrule
\multicolumn{4}{l}{\emph{CogVideoX-1.5} (transfer)}\\
{Base} & $-0.284$ & 5.1392 & 46.5880 \\
\iflatexml
{\galhtmlname{GAL-Image}} & $-0.727$ & 6.2742 & 51.7527 \\
 & $(-0.443)$ & $(+1.1350)$ & $(+5.1647)$ \\[3pt]
\else
{\galhtmlname{GAL-Image}} & $-0.727$ {($-0.443$)} & 6.2742 {($+1.1350$)} & 51.7527 {($+5.1647$)} \\
\fi
\iflatexml
{\galhtmlname{GAL-HPSv3}} & $-0.716$ & 6.2377 & 50.3643 \\
 & $(-0.432)$ & $(+1.0985)$ & $(+3.7763)$ \\[3pt]
\else
{\galhtmlname{GAL-HPSv3}} & $-0.716$ {($-0.432$)} & 6.2377 {($+1.0985$)} & 50.3643 {($+3.7763$)} \\
\fi
\iflatexml
{\galhtmlname{GAL-Unipercept}} & $-0.380$ & \textbf{6.4148} & \textbf{53.3527} \\
 & $(-0.096)$ & $(+1.2756)$ & $(+6.7647)$ \\[3pt]
\else
{\galhtmlname{GAL-Unipercept}} & $-0.380$ {($-0.096$)} & \textbf{6.4148} {($+1.2756$)} & \textbf{53.3527} {($+6.7647$)} \\
\fi
\iflatexml
{\textbf{\galhtmlname{GAL-VideoReward}}} & $\bm{-0.129}$ & 6.1601 & 51.4183 \\
 & $(+0.155)$ & $(+1.0209)$ & $(+4.8303)$ \\[3pt]
\else
{\textbf{\galhtmlname{GAL-VideoReward}}} & $\bm{-0.129}$ {($+0.155$)} & 6.1601 {($+1.0209$)} & 51.4183 {($+4.8303$)} \\
\fi
\midrule
\multicolumn{4}{l}{\emph{Kling v3 Pro} (closed-source transfer)}\\
{Base} & $+0.394$ & 8.2367 & 59.5410 \\
\iflatexml
{\galhtmlname{GAL-Image}} & $-0.862$ & 9.5553 & 63.8397 \\
 & $(-1.256)$ & $(+1.3185)$ & $(+4.2987)$ \\[3pt]
\else
{\galhtmlname{GAL-Image}} & $-0.862$ {($-1.256$)} & 9.5553 {($+1.3185$)} & 63.8397 {($+4.2987$)} \\
\fi
\iflatexml
{\galhtmlname{GAL-HPSv3}} & $-0.160$ & \textbf{10.5556} & 62.4798 \\
 & $(-0.554)$ & $(+2.3189)$ & $(+2.9389)$ \\[3pt]
\else
{\galhtmlname{GAL-HPSv3}} & $-0.160$ {($-0.554$)} & \textbf{10.5556} {($+2.3189$)} & 62.4798 {($+2.9389$)} \\
\fi
\iflatexml
{\galhtmlname{GAL-Unipercept}} & $-0.025$ & 9.4017 & \textbf{69.0857} \\
 & $(-0.419)$ & $(+1.1650)$ & $(+9.5448)$ \\[3pt]
\else
{\galhtmlname{GAL-Unipercept}} & $-0.025$ {($-0.419$)} & 9.4017 {($+1.1650$)} & \textbf{69.0857} {($+9.5448$)} \\
\fi
\iflatexml
{\textbf{\galhtmlname{GAL-VideoReward}}} & $\bm{+0.909}$ & 8.9403 & 62.4870 \\
 & $(+0.515)$ & $(+0.7035)$ & $(+2.9460)$ \\[3pt]
\else
{\textbf{\galhtmlname{GAL-VideoReward}}} & $\bm{+0.909}$ {($+0.515$)} & 8.9403 {($+0.7035$)} & 62.4870 {($+2.9460$)} \\
\fi
\bottomrule
\end{tabular*}
\vspace{0pt}
\end{table}

\end{document}

%% file: math_commands.tex
\newcommand{\E}{\mathbb{E}}

\newcommand{\Normal}{\mathcal{N}}



%% file: references.bib
@inproceedings{hu2022lora,
  title={{LoRA}: Low-rank adaptation of large language models},
  author={Hu, Edward J and Shen, Yelong and Wallis, Phillip and Allen-Zhu, Zeyuan and Li, Yuanzhi and Wang, Shean and Wang, Lu and Chen, Weizhu},
  booktitle={International Conference on Learning Representations (ICLR)},
  year={2022}
}

@inproceedings{salimans2022progressive,
  title={Progressive distillation for fast sampling of diffusion models},
  author={Salimans, Tim and Ho, Jonathan},
  year={2022},
  booktitle={International Conference on Learning Representations (ICLR)}
}

@inproceedings{song2023consistency,
  title={Consistency models},
  author={Song, Yang and Dhariwal, Prafulla and Chen, Mark and Sutskever, Ilya},
  booktitle={International Conference on Machine Learning (ICML)},
  year={2023}
}

@inproceedings{sauer2024adversarial,
  title={Adversarial diffusion distillation},
  author={Sauer, Axel and Lorenz, Dominik and Blattmann, Andreas and Rombach, Robin},
  booktitle={European Conference on Computer Vision},
  pages={87--103},
  year={2024},
  organization={Springer}
}

@article{cai2025z,
  title={{Z-Image}: An efficient image generation foundation model with single-stream diffusion transformer},
  author={{Z-Image Team} and Cai, Huanqia and Cao, Sihan and Du, Ruoyi and Gao, Peng and Hao, Aiming and Hoi, Steven and Hou, Zhaohui and Huang, Shijie and Jiang, Dengyang and Jiang, Yuming and Jin, Xin and Li, Liangchen and Li, Zhen and Li, Zhong-Yu and Liu, David and Liu, Dongyang and Wu, Qilong and Yu, Feng and Zhan, Zechao and Zhang, Chi and Zhang, Shifeng and Zhou, Ruikai and Zhou, Shilin},
  journal={arXiv preprint arXiv:2511.22699},
  year={2025}
}

@inproceedings{liu2026decoupled,
  title={Decoupled {DMD}: {CFG} Augmentation as the Spear, Distribution Matching as the Shield},
  author={Dongyang Liu and Peng Gao and David Liu and Ruoyi Du and Zhen Li and Qilong Wu and Xin Jin and Sihan Cao and Shifeng Zhang and Steven HOI and Hongsheng Li},
  booktitle={The Fourteenth International Conference on Learning Representations},
  year={2026},
  url={https://openreview.net/forum?id=jBztvOiCKE}
}

@misc{ostris2025zimageadapter,
  title        = {{Z-Image-Turbo} Training Adapter},
  author       = {{Ostris}},
  year         = {2025},
  note         = {Hugging Face model card},
  url          = {https://huggingface.co/ostris/zimage_turbo_training_adapter}
}

@misc{kelseye2025trainingstrategies,
  title        = {Training Strategies of {Z-Image-Turbo}},
  author       = {{kelseye.xh} and Duan, Zhongjie},
  year         = {2025},
  note         = {Hugging Face Blog. Accessed: 2026-07-05},
  url          = {https://huggingface.co/blog/kelseye/training-strategies-of-z-image-turbo}
}

@article{wu2024difflora,
  title={{DiffLoRA}: Generating personalized low-rank adaptation weights with diffusion},
  author={Wu, Yujia and Shi, Yiming and Wei, Jiwei and Sun, Chengwei and Yang, Yang and Shen, Heng Tao},
  journal={arXiv preprint arXiv:2408.06740},
  year={2024}
}

@article{rafailov2023direct,
  title={Direct preference optimization: Your language model is secretly a reward model},
  author={Rafailov, Rafael and Sharma, Archit and Mitchell, Eric and Manning, Christopher D and Ermon, Stefano and Finn, Chelsea},
  journal={Advances in neural information processing systems},
  volume={36},
  pages={53728--53741},
  year={2023}
}

@inproceedings{azar2024general,
  title={A general theoretical paradigm to understand learning from human preferences},
  author={Azar, Mohammad Gheshlaghi and Guo, Zhaohan Daniel and Piot, Bilal and Munos, Remi and Rowland, Mark and Valko, Michal and Calandriello, Daniele},
  booktitle={International Conference on Artificial Intelligence and Statistics},
  pages={4447--4455},
  year={2024},
  organization={PMLR}
}

@article{meng2024simpo,
  title={{SimPO}: Simple preference optimization with a reference-free reward},
  author={Meng, Yu and Xia, Mengzhou and Chen, Danqi},
  journal={Advances in Neural Information Processing Systems},
  volume={37},
  pages={124198--124235},
  year={2024}
}

@inproceedings{wallace2024diffusion,
  title={Diffusion model alignment using direct preference optimization},
  author={Wallace, Bram and Dang, Meihua and Rafailov, Rafael and Zhou, Linqi and Lou, Aaron and Purushwalkam, Senthil and Ermon, Stefano and Xiong, Caiming and Joty, Shafiq and Naik, Nikhil},
  booktitle={Proceedings of the IEEE/CVF Conference on Computer Vision and Pattern Recognition},
  pages={8228--8238},
  year={2024}
}

@article{wu2023human,
  title={Human preference score v2: A solid benchmark for evaluating human preferences of text-to-image synthesis},
  author={Wu, Xiaoshi and Hao, Yiming and Sun, Keqiang and Chen, Yixiong and Zhu, Feng and Zhao, Rui and Li, Hongsheng},
  journal={arXiv preprint arXiv:2306.09341},
  year={2023}
}

@inproceedings{ma2025hpsv3,
  title={{HPSv3}: Towards wide-spectrum human preference score},
  author={Ma, Yuhang and Wu, Xiaoshi and Sun, Keqiang and Li, Hongsheng},
  booktitle={Proceedings of the IEEE/CVF International Conference on Computer Vision},
  pages={15086--15095},
  year={2025}
}

@inproceedings{cao2026unipercept,
  title={{UniPercept}: Towards Unified Perceptual-Level Image Understanding across Aesthetics, Quality, Structure, and Texture},
  author={Shuo Cao and Jiayang Li and Xiaohui Li and Yuandong Pu and Kaiwen Zhu and Yuanting Gao and Siqi Luo and Yi Xin and Qi Qin and Yu Zhou and Xiangyu Chen and Wenlong Zhang and Bin Fu and Yu Qiao and Yihao Liu},
  booktitle={Forty-third International Conference on Machine Learning},
  year={2026},
  url={https://openreview.net/forum?id=8VVbElV06v}
}

@inproceedings{NEURIPS2025_76227feb,
  author={Liu, Jie and Liu, Gongye and Liang, Jiajun and Yuan, Ziyang and Liu, Xiaokun and Zheng, Mingwu and Wu, Xiele and Wang, Qiulin and Xia, Menghan and Wang, Xintao and Liu, Xiaohong and Yang, Fei and Wan, Pengfei and ZHANG, Di and Gai, Kun and Yang, Yujiu and Ouyang, Wanli},
  booktitle={Advances in Neural Information Processing Systems},
  doi={10.52202/085713-2750},
  editor={D. Belgrave and C. Zhang and H. Lin and R. Pascanu and P. Koniusz and M. Ghassemi and N. Chen},
  pages={82155--82192},
  publisher={Curran Associates, Inc.},
  title={Improving Video Generation with Human Feedback},
  url={https://proceedings.neurips.cc/paper_files/paper/2025/file/76227feb18ea0ee40bd15cf02c33e18e-Paper-Conference.pdf},
  volume={38, Main Conference},
  year={2025}
}

@inproceedings{huang2024vbench,
  title={{VBench}: Comprehensive benchmark suite for video generative models},
  author={Huang, Ziqi and He, Yinan and Yu, Jiashuo and Zhang, Fan and Si, Chenyang and Jiang, Yuming and Zhang, Yuanhan and Wu, Tianxing and Jin, Qingyang and Chanpaisit, Nattapol and others},
  booktitle={Proceedings of the IEEE/CVF Conference on Computer Vision and Pattern Recognition},
  pages={21807--21818},
  year={2024}
}

@article{wu2025hunyuanvideo,
  title={{HunyuanVideo} 1.5 technical report},
  author={Wu, Bing and Zou, Chang and Li, Changlin and Huang, Duojun and Yang, Fang and Tan, Hao and Peng, Jack and Wu, Jianbing and Xiong, Jiangfeng and Jiang, Jie and others},
  journal={arXiv preprint arXiv:2511.18870},
  year={2025}
}

@inproceedings{song2020score,
  title={Score-based generative modeling through stochastic differential equations},
  author={Song, Yang and Sohl-Dickstein, Jascha and Kingma, Diederik P and Kumar, Abhishek and Ermon, Stefano and Poole, Ben},
  year={2021},
  booktitle={International Conference on Learning Representations (ICLR)}
}

@article{shao2024deepseekmath,
  title={{DeepSeekMath}: Pushing the limits of mathematical reasoning in open language models},
  author={Shao, Zhihong and Wang, Peiyi and Zhu, Qihao and Xu, Runxin and Song, Junxiao and Bi, Xiao and Zhang, Haowei and Zhang, Mingchuan and Li, Y. K. and Wu, Y. and Guo, Daya},
  journal={arXiv preprint arXiv:2402.03300},
  year={2024}
}

@article{liu2026flow,
  title={{Flow-GRPO}: Training flow matching models via online {RL}},
  author={Liu, Jie and Liu, Gongye and Liang, Jiajun and Li, Yangguang and Liu, Jiaheng and Wang, Xintao and Wan, Pengfei and Zhang, Di and Ouyang, Wanli},
  journal={Advances in neural information processing systems},
  volume={38},
  pages={40783--40818},
  year={2025}
}

@inproceedings{black2024training,
  title={Training diffusion models with reinforcement learning},
  author={Black, Kevin and Janner, Michael and Du, Yilun and Kostrikov, Ilya and Levine, Sergey},
  booktitle={International Conference on Learning Representations},
  volume={2024},
  pages={4965--4987},
  year={2024}
}

@inproceedings{zheng2025diffusionnft,
  title={{DiffusionNFT}: Online diffusion reinforcement with forward process},
  author={Zheng, Kaiwen and Chen, Huayu and Ye, Haotian and Wang, Haoxiang and Zhang, Qinsheng and Jiang, Kai and Su, Hang and Ermon, Stefano and Zhu, Jun and Liu, Ming-Yu},
  year={2026},
  booktitle={International Conference on Learning Representations (ICLR)}
}

@inproceedings{liu2025videodpo,
  title={{VideoDPO}: Omni-preference alignment for video diffusion generation},
  author={Liu, Runtao and Wu, Haoyu and Zheng, Ziqiang and Wei, Chen and He, Yingqing and Pi, Renjie and Chen, Qifeng},
  booktitle={Proceedings of the Computer Vision and Pattern Recognition Conference},
  pages={8009--8019},
  year={2025}
}

@article{wu2026densedpo,
  title={{DenseDPO}: Fine-grained temporal preference optimization for video diffusion models},
  author={Wu, Ziyi and Kag, Anil and Skorokhodov, Ivan and Menapace, Willi and Mirzaei, Ashkan and Gilitschenski, Igor and Tulyakov, Sergey and Siarohin, Aliaksandr},
  journal={Advances in Neural Information Processing Systems},
  volume={38},
  pages={171632--171668},
  year={2025}
}

@article{chen2026hierarchical,
  title={Hierarchical fine-grained preference optimization for physically plausible video generation},
  author={Chen, Harold Haodong and Huang, Haojian and Chen, Qifeng and Yang, Harry and Lim, Ser Nam},
  journal={Advances in Neural Information Processing Systems},
  volume={38},
  pages={133919--133951},
  year={2025}
}

@article{polyak2024movie,
  title={{Movie Gen}: A cast of media foundation models},
  author={Polyak, Adam and Zohar, Amit and Brown, Andrew and Tjandra, Andros and Sinha, Animesh and Lee, Ann and Vyas, Apoorv and Shi, Bowen and Ma, Chih-Yao and Chuang, Ching-Yao and others},
  journal={arXiv preprint arXiv:2410.13720},
  year={2024}
}

@article{kong2024hunyuanvideo,
  title={{HunyuanVideo}: A systematic framework for large video generative models},
  author={Kong, Weijie and Tian, Qi and Zhang, Zijian and Min, Rox and Dai, Zuozhuo and Zhou, Jin and Xiong, Jiangfeng and Li, Xin and Wu, Bo and Zhang, Jianwei and others},
  journal={arXiv preprint arXiv:2412.03603},
  year={2024}
}

@inproceedings{yang2025cogvideox,
  title={{CogVideoX}: Text-to-video diffusion models with an expert transformer},
  author={Yang, Zhuoyi and Teng, Jiayan and Zheng, Wendi and Ding, Ming and Huang, Shiyu and Xu, Jiazheng and Yang, Yuanming and Hong, Wenyi and Zhang, Xiaohan and Feng, Guanyu and others},
  booktitle={International Conference on Learning Representations},
  volume={2025},
  pages={83048--83077},
  year={2025}
}

@article{wan2025wan,
  title={Wan: Open and advanced large-scale video generative models},
  author={{Wan Team} and Wang, Ang and Ai, Baole and Wen, Bin and Mao, Chaojie and Xie, Chen-Wei and Chen, Di and Yu, Feiwu and Zhao, Haiming and Yang, Jianxiao and others},
  journal={arXiv preprint arXiv:2503.20314},
  year={2025}
}

@article{hacohen2024ltx,
  title={{LTX-Video}: Realtime video latent diffusion},
  author={HaCohen, Yoav and Chiprut, Nisan and Brazowski, Benny and Shalem, Daniel and Moshe, Dudu and Richardson, Eitan and Levin, Eran and Shiran, Guy and Zabari, Nir and Gordon, Ori and others},
  journal={arXiv preprint arXiv:2501.00103},
  year={2024}
}

@inproceedings{valevski2025diffusion,
  title={Diffusion models are real-time game engines},
  author={Valevski, Dani and Leviathan, Yaniv and Arar, Moab and Fruchter, Shlomi},
  booktitle={International Conference on Learning Representations},
  volume={2025},
  pages={73754--73776},
  year={2025}
}

@inproceedings{podell2024sdxl,
  title={{SDXL}: Improving Latent Diffusion Models for High-Resolution Image Synthesis},
  author={Podell, Dustin and English, Zion and Lacey, Kyle and Blattmann, Andreas and Dockhorn, Tim and M{\"u}ller, Jonas and Penna, Joe and Rombach, Robin},
  booktitle={International Conference on Learning Representations},
  volume={2024},
  pages={1862--1874},
  year={2024}
}

@article{blattmann2023stable,
  title={Stable video diffusion: Scaling latent video diffusion models to large datasets},
  author={Blattmann, Andreas and Dockhorn, Tim and Kulal, Sumith and Mendelevitch, Daniel and Kilian, Maciej and Lorenz, Dominik and Levi, Yam and English, Zion and Voleti, Vikram and Letts, Adam and others},
  journal={arXiv preprint arXiv:2311.15127},
  year={2023}
}

@inproceedings{xing2024dynamicrafter,
  title={{DynamiCrafter}: Animating Open-Domain Images with Video Diffusion Priors},
  author={Xing, Jinbo and Xia, Menghan and Zhang, Yong and Chen, Haoxin and Yu, Wangbo and Liu, Hanyuan and Liu, Gongye and Wang, Xintao and Shan, Ying and Wong, Tien-Tsin},
  booktitle={European Conference on Computer Vision},
  pages={399--417},
  year={2024},
  organization={Springer}
}

@article{zhang2023i2vgen,
  title={{I2VGen-XL}: High-Quality Image-to-Video Synthesis via Cascaded Diffusion Models},
  author={Zhang, Shiwei and Wang, Jiayu and Zhang, Yingya and Zhao, Kang and Yuan, Hangjie and Qin, Zhiwu and Wang, Xiang and Zhao, Deli and Zhou, Jingren},
  journal={arXiv preprint arXiv:2311.04145},
  year={2023}
}

@inproceedings{li2024anim,
  title={{Anim-Director}: A large multimodal model powered agent for controllable animation video generation},
  author={Li, Yunxin and Shi, Haoyuan and Hu, Baotian and Wang, Longyue and Zhu, Jiashun and Xu, Jinyi and Zhao, Zhen and Zhang, Min},
  booktitle={SIGGRAPH Asia 2024 Conference Papers},
  pages={1--11},
  year={2024}
}

@inproceedings{lipman2023flow,
  title={Flow Matching for Generative Modeling},
  author={Yaron Lipman and Ricky T. Q. Chen and Heli Ben-Hamu and Maximilian Nickel and Matthew Le},
  booktitle={The Eleventh International Conference on Learning Representations},
  year={2023},
  url={https://openreview.net/forum?id=PqvMRDCJT9t}
}

@inproceedings{liu2022flow,
  title={Flow straight and fast: Learning to generate and transfer data with rectified flow},
  author={Liu, Xingchao and Gong, Chengyue and Liu, Qiang},
  year={2023},
  booktitle={International Conference on Learning Representations (ICLR)}
}

@article{kling2025omni,
  title={{Kling-Omni} technical report},
  author={{Kling Team}},
  journal={arXiv preprint arXiv:2512.16776},
  year={2025}
}

@inproceedings{esser2024scaling,
  title={Scaling rectified flow transformers for high-resolution image synthesis},
  author={Esser, Patrick and Kulal, Sumith and Blattmann, Andreas and Entezari, Rahim and M{\"u}ller, Jonas and Saini, Harry and Levi, Yam and Lorenz, Dominik and Sauer, Axel and Boesel, Frederic and others},
  booktitle={International Conference on Machine Learning},
  year={2024}
}

@inproceedings{Lin_2025_ICCV,
  author={Lin, Gaojie and Jiang, Jianwen and Yang, Jiaqi and Zheng, Zerong and Liang, Chao and Zhang, Yuan and Liu, Jingtuo},
  title={{OmniHuman-1}: Rethinking the Scaling-Up of One-Stage Conditioned Human Animation Models},
  booktitle={Proceedings of the IEEE/CVF International Conference on Computer Vision (ICCV)},
  month={October},
  year={2025},
  pages={13847--13858}
}

@inproceedings{tian2024emo,
  title={{EMO}: Emote portrait alive -- generating expressive portrait videos with {Audio2Video} diffusion model under weak conditions},
  author={Tian, Linrui and Wang, Qi and Zhang, Bang and Bo, Liefeng},
  booktitle={European Conference on Computer Vision},
  year={2024}
}

@article{wu2025movieagent,
  title={Automated movie generation via multi-agent {CoT} planning},
  author={Wu, Weijia and Zhu, Zeyu and Shou, Mike Zheng},
  journal={arXiv preprint arXiv:2503.07314},
  year={2025}
}

@inproceedings{wang-etal-2026-mavis,
  title={{MAV}i{S}: A Multi-Agent Framework for Long-Sequence Video Storytelling},
  author={Wang, Qian and Huang, Ziqi and Jia, Ruoxi and Debevec, Paul and Yu, Ning},
  editor={Demberg, Vera and Inui, Kentaro and Marquez, Llu{\'i}s},
  booktitle={Proceedings of the 19th Conference of the {E}uropean Chapter of the {A}ssociation for {C}omputational {L}inguistics (Volume 1: Long Papers)},
  month=mar,
  year={2026},
  address={Rabat, Morocco},
  publisher={Association for Computational Linguistics},
  url={https://aclanthology.org/2026.eacl-long.101/},
  doi={10.18653/v1/2026.eacl-long.101},
  pages={2273--2295},
  ISBN={979-8-89176-380-7}
}

@inproceedings{shi2025animaker,
  title={{AniMaker}: Multi-agent animated storytelling with {MCTS}-driven clip generation},
  author={Shi, Haoyuan and Li, Yunxin and Chen, Xinyu and Wang, Longyue and Hu, Baotian and Zhang, Min},
  year={2025},
  booktitle={Proceedings of the SIGGRAPH Asia 2025 Conference Papers},
  pages={1--11}
}

@inproceedings{bruce2024genie,
  title={Genie: Generative interactive environments},
  author={Bruce, Jake and Dennis, Michael D. and Edwards, Ashley and Parker-Holder, Jack and Shi, Yuge and Hughes, Edward and Lai, Matthew and Mavalankar, Aditi and Steigerwald, Richie and Apps, Chris and others},
  booktitle={International Conference on Machine Learning},
  year={2024}
}

@article{nvidia2025cosmos,
  title={Cosmos world foundation model platform for physical {AI}},
  author={{NVIDIA}},
  journal={arXiv preprint arXiv:2501.03575},
  year={2025}
}

@inproceedings{10.1007/978-3-031-73033-7_12,
  author={Girdhar, Rohit and Singh, Mannat and Brown, Andrew and Duval, Quentin and Azadi, Samaneh and Rambhatla, Sai Saketh and Shah, Akbar and Yin, Xi and Parikh, Devi and Misra, Ishan},
  editor={Leonardis, Ale{\v{s}} and Ricci, Elisa and Roth, Stefan and Russakovsky, Olga and Sattler, Torsten and Varol, G{\"u}l},
  title={Factorizing Text-to-Video Generation by Explicit Image Conditioning},
  booktitle={Computer Vision -- ECCV 2024},
  year={2025},
  publisher={Springer Nature Switzerland},
  address={Cham},
  pages={205--224},
  isbn={978-3-031-73033-7}
}

@inproceedings{zhao2024identifying,
  title={Identifying and solving conditional image leakage in image-to-video diffusion model},
  author={Zhao, Min and Zhu, Hongzhou and Xiang, Chendong and Zheng, Kaiwen and Li, Chongxuan and Zhu, Jun},
  booktitle={Advances in Neural Information Processing Systems},
  year={2024}
}

@inproceedings{choi2025improving,
  title={Improving motion in image-to-video models via adaptive low-pass guidance},
  author={Choi, June Suk and Lee, Kyungmin and Yu, Sihyun and Choi, Yisol and Shin, Jinwoo and Lee, Kimin},
  booktitle={IEEE/CVF Conference on Computer Vision and Pattern Recognition},
  year={2026}
}

@article{jeon2026rebalancing,
  title={Rebalancing reference frame dominance to improve motion in image-to-video models},
  author={Jeon, Wooseok and Park, Seungho and Shin, Seunghyun and Lee, Sangeyl and Jeong, Hyeonho and Jeon, Hae-Gon},
  journal={arXiv preprint arXiv:2605.19398},
  year={2026}
}

@inproceedings{chen2026first,
  title={First frame is the place to go for video content customization},
  author={Chen, Jingxi and Li, Zongxia and Liu, Zhichao and Shi, Guangyao and Wu, Xiyang and Liu, Fuxiao and Ferm{\"u}ller, Cornelia and Feng, Brandon Y. and Aloimonos, Yiannis},
  booktitle={IEEE/CVF Conference on Computer Vision and Pattern Recognition},
  year={2026}
}

@inproceedings{farneback2003two,
  title={Two-frame motion estimation based on polynomial expansion},
  author={Farneb{\"a}ck, Gunnar},
  booktitle={Scandinavian Conference on Image Analysis (SCIA)},
  pages={363--370},
  year={2003}
}

@article{guo2024i4vgen,
  title={{I4VGen}: Image as free stepping stone for text-to-video generation},
  author={Guo, Xiefan and Liu, Jinlin and Cui, Miaomiao and Bo, Liefeng and Huang, Di},
  journal={arXiv preprint arXiv:2406.02230},
  year={2024}
}

@article{ge2025flashi2v,
  title={{FlashI2V}: Fourier-guided latent shifting prevents conditional image leakage in image-to-video generation},
  author={Ge, Yunyang and Cheng, Xinhua and Zhao, Chengshu and He, Xianyi and Yuan, Shenghai and Lin, Bin and Zhu, Bin and Yuan, Li},
  journal={arXiv preprint arXiv:2509.25187},
  year={2025}
}

@inproceedings{cheng2025vpo,
  title={{VPO}: Aligning text-to-video generation models with prompt optimization},
  author={Cheng, Jiale and Lyu, Ruiliang and Gu, Xiaotao and Liu, Xiao and Xu, Jiazheng and Lu, Yida and Teng, Jiayan and Yang, Zhuoyi and Dong, Yuxiao and Tang, Jie and Wang, Hongning and Huang, Minlie},
  booktitle={Proceedings of the IEEE/CVF International Conference on Computer Vision},
  year={2025}
}

@article{yin2026closedloop,
  title={Closed-loop triplet synergistic generation for long-form video},
  author={Yin, Xinlei and Peng, Xiulian and Li, Xiao and Xiong, Zhiwei and Lu, Yan},
  journal={arXiv preprint arXiv:2606.16184},
  year={2026}
}

@inproceedings{zhou2025golden,
  title={Golden Noise for Diffusion Models: A Learning Framework},
  author={Zhou, Zikai and Shao, Shitong and Bai, Lichen and Zhang, Shufei and Xu, Zhiqiang and Han, Bo and Xie, Zeke},
  booktitle={Proceedings of the IEEE/CVF International Conference on Computer Vision (ICCV)},
  pages={17688--17697},
  year={2025},
  url={https://openaccess.thecvf.com/content/ICCV2025/html/Zhou_Golden_Noise_for_Diffusion_Models_A_Learning_Framework_ICCV_2025_paper.html}
}

@article{shao2026improved,
  title={Improved and accelerated text-to-image generation with collect, reflect, and refine},
  author={Shao, Shitong and Zhou, Zikai and Xie, Dian and Fang, Yuetong and Ye, Tian and Bai, Lichen and Han, Bo and Xie, Zeke},
  journal={IEEE Transactions on Pattern Analysis and Machine Intelligence},
  year={2026},
  publisher={IEEE}
}

@inproceedings{shao2025ivmixed,
  title={{IV}-mixed Sampler: Leveraging Image Diffusion Models for Enhanced Video Synthesis},
  author={Shao, Shitong and Zhou, Zikai and Bai, Lichen and Xiong, Haoyi and Xie, Zeke},
  booktitle={The Thirteenth International Conference on Learning Representations},
  year={2025},
  url={https://openreview.net/forum?id=ImpeMDJfVL}
}

@inproceedings{lichen2025zigzag,
  title={Zigzag Diffusion Sampling: Diffusion Models Can Self-Improve via Self-Reflection},
  author={Bai, Lichen and Shao, Shitong and Zhou, Zikai and Qi, Zipeng and Xu, Zhiqiang and Xiong, Haoyi and Xie, Zeke},
  booktitle={The Thirteenth International Conference on Learning Representations},
  year={2025},
  url={https://openreview.net/forum?id=MKvQH1ekeY}
}

@article{10.1145/3796982,
  author={Liu, Buhua and Shao, Shitong and Li, Bao and Bai, Lichen and Xu, Zhiqiang and Xiong, Haoyi and Kwok, James T. and Helal, Sumi and Xie, Zeke},
  title={Alignment of Diffusion Models: Fundamentals, Challenges, and Future},
  year={2026},
  issue_date={July 2026},
  publisher={Association for Computing Machinery},
  address={New York, NY, USA},
  volume={58},
  number={9},
  issn={0360-0300},
  url={https://doi.org/10.1145/3796982},
  doi={10.1145/3796982},
  journal={ACM Comput. Surv.},
  month=mar,
  articleno={244},
  numpages={37},
  pages={244:1--244:37}
}


%% file: references_arxiv_additions.bib
@misc{zhang2025hoidr1reinforcementlearningopenworld,
  title = {{HOID-R1}: Reinforcement Learning for Open-World Human-Object Interaction Detection Reasoning with Multimodal Large Language Model},
  author = {Zhenhao Zhang and Hanqing Wang and Xiangyu Zeng and Ziyu Cheng and Jiaxin Liu and Haoyu Yan and Zhirui Liu and Kaiyang Ji and Tianxiang Gui and Ke Hu and Kangyi Chen and Yahao Fan and Mokai Pan},
  year = {2025},
  eprint = {2508.11350},
  archivePrefix = {arXiv},
  primaryClass = {cs.CV},
  url = {https://arxiv.org/abs/2508.11350}
}

@article{wang2026affordancer1,
  title = {{Affordance-R1}: Reinforcement Learning for Generalizable Affordance Reasoning in Multimodal Large Language Models},
  author = {Wang, Hanqing and Wang, Shaoyang and Zhong, Yiming and Yang, Zemin and Wang, Jiamin and Cui, Zhiqing and Yuan, Jiahao and Han, Yifan and Liu, Mingyu and Ma, Yuexin},
  journal = {Proceedings of the AAAI Conference on Artificial Intelligence},
  volume = {40},
  number = {12},
  pages = {9738--9746},
  year = {2026},
  month = mar,
  doi = {10.1609/aaai.v40i12.37937},
  url = {https://ojs.aaai.org/index.php/AAAI/article/view/37937}
}

@inproceedings{sun2026craft,
  title = {{CRAFT}: Aligning Diffusion Models with Fine-Tuning Is Easier Than You Think},
  author = {Zening Sun and Zhengpeng Xie and Lichen Bai and Shitong Shao and Shuo Yang and Zeke Xie},
  booktitle = {Conference on Computer Vision and Pattern Recognition 2026},
  year = {2026},
  url = {https://openreview.net/forum?id=iNaUPXFbbb}
}

@inproceedings{xie2026guidance,
  title = {Guidance Matters: Rethinking the Evaluation Pitfall for Text-to-Image Generation},
  author = {Dian Xie and Shitong Shao and Lichen Bai and Zikai Zhou and Bojun Cheng and Shuo Yang and Jun Wu and Zeke Xie},
  booktitle = {The Fourteenth International Conference on Learning Representations},
  year = {2026},
  url = {https://openreview.net/forum?id=T9xcbgFD3k}
}

@inproceedings{bai2026optimizing,
  title = {Optimizing Few-Step Generation with Adaptive Matching Distillation},
  author = {Lichen Bai and Zikai Zhou and Shitong Shao and Wenliang Zhong and Shuo Yang and Shuo Chen and Bojun Cheng and Zeke Xie},
  booktitle = {Forty-third International Conference on Machine Learning},
  year = {2026},
  url = {https://openreview.net/forum?id=QGA9cWGb12}
}

@inproceedings{shao2026fastlightgen,
  title = {{FastLightGen}: Fast and Light Video Generation with Fewer Steps and Parameters},
  author = {Shitong Shao and Yufei Gu and Zeke Xie},
  booktitle = {Conference on Computer Vision and Pattern Recognition 2026},
  year = {2026},
  url = {https://openreview.net/forum?id=XNU4lRHhaU}
}

@article{DBLP:journals/corr/abs-2606-17800,
  publtype = {informal},
  author = {Lichen Bai and Tianhao Zhang and Shitong Shao and Dingwei Tan and Qiyu Zhong and Zhengpeng Xie and Haopeng Li and Qinghao Huang and Dandan Shen and Tengjiao Ji and Wei Wang and Peicheng Wu and Yuxuan Zhao and Xiangyu Zhu and Welly Luo and Shurui Yang and Zeke Xie},
  title = {{MaineCoon}: Pursuing A Real-Time Audio-Visual Social World Model},
  year = {2026},
  month = {June},
  cdate = {1780272000000},
  journal = {CoRR},
  volume = {abs/2606.17800},
  url = {https://doi.org/10.48550/arXiv.2606.17800}
}

@misc{shao2025magicdistillationweaktostrongvideodistillation,
  title = {{MagicDistillation}: Weak-to-Strong Video Distillation for Large-Scale Few-Step Synthesis},
  author = {Shitong Shao and Hongwei Yi and Hanzhong Guo and Tian Ye and Daquan Zhou and Michael Lingelbach and Zhiqiang Xu and Zeke Xie},
  year = {2025},
  eprint = {2503.13319},
  archivePrefix = {arXiv},
  primaryClass = {cs.CV},
  url = {https://arxiv.org/abs/2503.13319}
}
